\documentclass{article}
\usepackage{datetime}

\usepackage[final, nonatbib]{neurips_2021}
\usepackage{defs}
\usepackage{todonotes}
\usepackage{wrapfig}
\usepackage{graphicx}
\usepackage{caption}
\usepackage{subcaption}
\usepackage{xcolor}
\usepackage{amssymb}
\usepackage[toc,page,header]{appendix}
\usepackage{minitoc}

\usepackage[utf8]{inputenc} 
\usepackage[T1]{fontenc}    
\usepackage{hyperref}       
\usepackage{url}            
\usepackage{booktabs}       
\usepackage{amsfonts}       
\usepackage{nicefrac}       
\usepackage{microtype}      
\usepackage{xcolor}         
\usepackage{enumitem}
\usepackage[ruled, vlined, linesnumbered]{algorithm2e}%
\usepackage{algpseudocode}
\SetAlFnt{\scriptsize}
\usepackage{rotating}
\usepackage{tabularx}
\usepackage{multirow}
\usepackage{adjustbox}
\title{\model{}: Coreset Selection for Efficient and Robust Semi-Supervised Learning}
\newcommand{\model}{\textsc{Retrieve}}
\newcommand{\figref}[1]{Figure~\ref{#1}}
\newcommand{\tabref}[1]{Table~\ref{#1}}
\newcommand{\secref}[1]{Section~\ref{#1}}
\newcommand{\AlgRef}[1]{Algorithm~\ref{#1}}
\renewcommand{\eqref}[1]{Equation~(\ref{#1})}

\author{Krishnateja Killamsetty \quad  Xujiang Zhao \quad  Feng Chen \quad Rishabh Iyer \\
	Department of Computer Science\\
	The University of Texas at Dallas\\
	Richardson, Texas, USA\\
	\{\texttt{krishnateja.killamsetty,xujiang.zhao,feng.chen,rishabh.iyer\}@utdallas.edu} \\
}
  
\begin{document}
\doparttoc 
\faketableofcontents 

\maketitle
\begin{abstract} 
Semi-supervised learning (SSL) algorithms have had great success in recent years in limited labeled data regimes. However, the current state-of-the-art SSL algorithms are computationally expensive and entail significant compute time and energy requirements. This can prove to be a huge limitation for many smaller companies and academic groups. Our main insight is that training on a subset of unlabeled data instead of entire unlabeled data enables the current SSL algorithms to converge faster, significantly reducing computational costs. In this work, we propose \model{}\footnote{co\textbf{R}esets for \textbf{E}fficien\textbf{T} and \textbf{R}obust  sem\textbf{I}-sup\textbf{E}r\textbf{V}ised l\textbf{E}arning}, a coreset selection framework for efficient and robust semi-supervised learning. \model{} selects the coreset by solving a mixed discrete-continuous bi-level optimization problem such that the selected coreset minimizes the labeled set loss. We use a one-step gradient approximation and show that the discrete optimization problem is approximately submodular, enabling simple greedy algorithms to obtain the coreset. We empirically demonstrate on several real-world datasets that existing SSL algorithms like VAT, Mean-Teacher, FixMatch, when used with \model, achieve a) faster training times,  b) better performance when unlabeled data consists of Out-of-Distribution (OOD) data and imbalance. More specifically, we show that with minimal accuracy degradation, \model{} achieves a speedup of around $3\times$ in the traditional SSL setting and achieves a speedup of $5\times$ compared to state-of-the-art (SOTA) robust SSL algorithms in the case of imbalance and OOD data. \model{} is available as a part of the CORDS toolkit: \url{https://github.com/decile-team/cords}.
\end{abstract}

\section{Introduction} 
Deep learning algorithms have had great success over the past few years, often achieving human or superhuman performance in various tasks like computer vision \cite{ciresan2012multicolumn}, speech recognition \cite{hersheymultitalker}, natural language processing \cite{wmt-2019-1}, and video games \cite{openai2019dota}. One of the significant factors attributing to the recent success of deep learning is the availability of large amounts of labeled data \cite{sun2017revisiting}. However, creating large labeled datasets is often time-consuming and expensive in terms of costs. Moreover, some domains like medical imaging require a domain expert for labeling, making it nearly impossible to create a large labeled set. In order to reduce the dependency on the availability of labeled data, semi-supervised learning (SSL) algorithms \cite{chapellessl} were proposed to train models using large amounts of unlabeled data along with the available labeled data. Recent works \cite{miyato2018virtual, tarvainen2018mean, berthelot2019mixmatch, sohn2020fixmatch} show that semi-supervised learning algorithms can achieve similar performance to standard supervised learning using significantly fewer labeled data instances. 

\begin{wrapfigure}{R}{0.6\textwidth}
\centering
\includegraphics[width=0.55\textwidth, height=2.8cm]{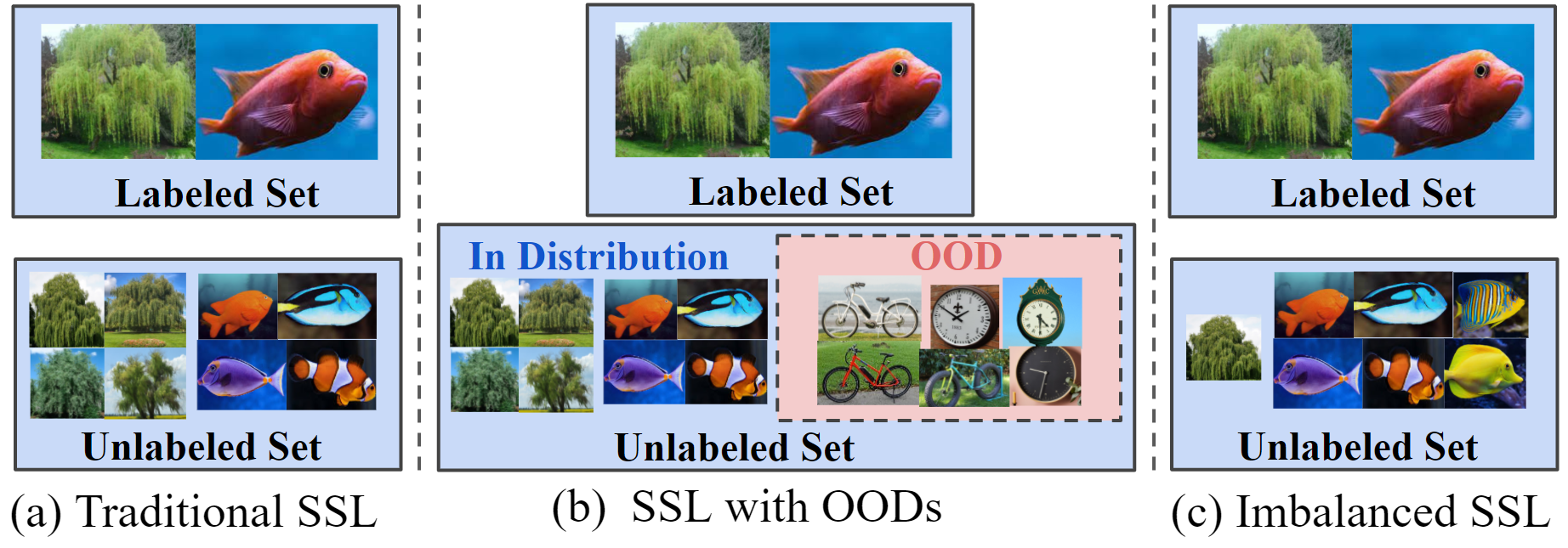}
\caption{{(a )Unlabeled set with the same distribution as the labeled set, (b) Unlabeled set containing OOD instances, (c) Unlabeled set where the class distribution is imbalanced}}
\label{fig:OOD_explanation}
\end{wrapfigure}

However, the current SOTA SSL algorithms are compute-intensive with large training times. For example, from our personal experience, training a WideResNet model \cite{zagoruyko2017wide} on a CIFAR10 \cite{Krizhevsky09learningmultiple} dataset with 4000 labels using the SOTA FixMatch algorithm \cite{sohn2020fixmatch} for 500000 iterations takes around four days on a single RTX2080Ti GPU. This also implies increased energy consumption and an associated carbon footprint \cite{strubell2019energy}.  Furthermore, it is common to tune these SSL algorithms over a large set of hyper-parameters, which means that the training needs to be done hundreds and sometimes thousands of times. For example, \cite{oliver2019realistic} performed hyperparameter tuning by running 1000 trails of Gaussian Process-based Blackbox optimization\cite{vizier} for each SSL algorithm (which runs for 500000 iterations). This process implies significantly higher experimental turnaround times, energy consumption, and CO2 emissions. Furthermore, this is not something that can be done at most universities and smaller companies. The first problem we try to address in this work is: \emph{Can we efficiently train a semi-supervised learning model on coresets of unlabeled data to achieve faster convergence and reduction in training time?}

Despite demonstrating encouraging results on standard and clean datasets, current SSL algorithms perform poorly when OOD data or class imbalance is present in the unlabeled set~\cite{oliver2019realistic, chen2019distributionally}. This performance degradation can be attributed to the fact that the current SSL algorithms assume that both the labeled set and unlabeled set are sampled from the same distribution. A visualization of OOD data and class imbalance in the unlabeled set is shown in \figref{fig:OOD_explanation}. Several recent works \cite{yan2016robust, chen2019distributionally, pmlr-v119-guo20i} were proposed to mitigate the effect of OOD in unlabeled data, in turn improving the performance of SSL algorithms. However, the current SOTA robust SSL method \cite{pmlr-v119-guo20i} is 3X slower than the standard SSL algorithms, further increasing the training times, energy costs, and CO2 emissions. The second problem we try to address in this work is: \emph{In the case where OOD data or class imbalance exists in the unlabeled set, can we robustly train an SSL model on coresets of unlabeled data to achieve similar performance to existing robust SSL methods while being significantly faster?} 

\begin{figure}[h]
\centering
\begin{subfigure}[b]{0.48 \textwidth}
\centering
\includegraphics[width=6.2cm, height=3.5cm]{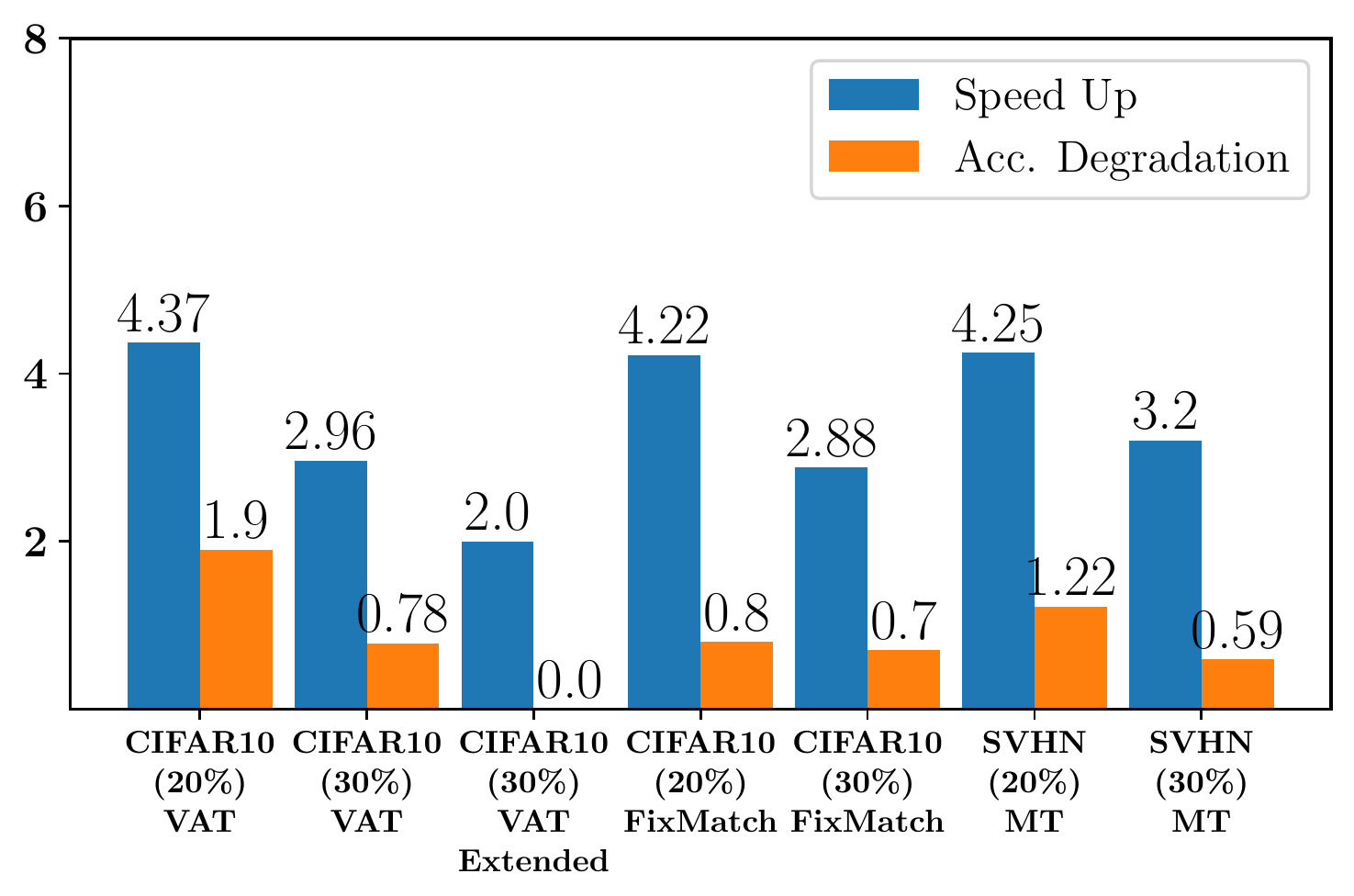}
\caption*{$\underbracket[1pt][1.0mm]{\hspace{6.2cm}}_{\substack{\vspace{-4.0mm}\\
\colorbox{white}{(a) \scriptsize Traditional SSL}}}$}
\phantomcaption
\label{fig:main-eff}
\end{subfigure}\quad
\begin{subfigure}[b]{0.48 \textwidth}
\centering
\includegraphics[width=6.2cm, height=3.5cm]{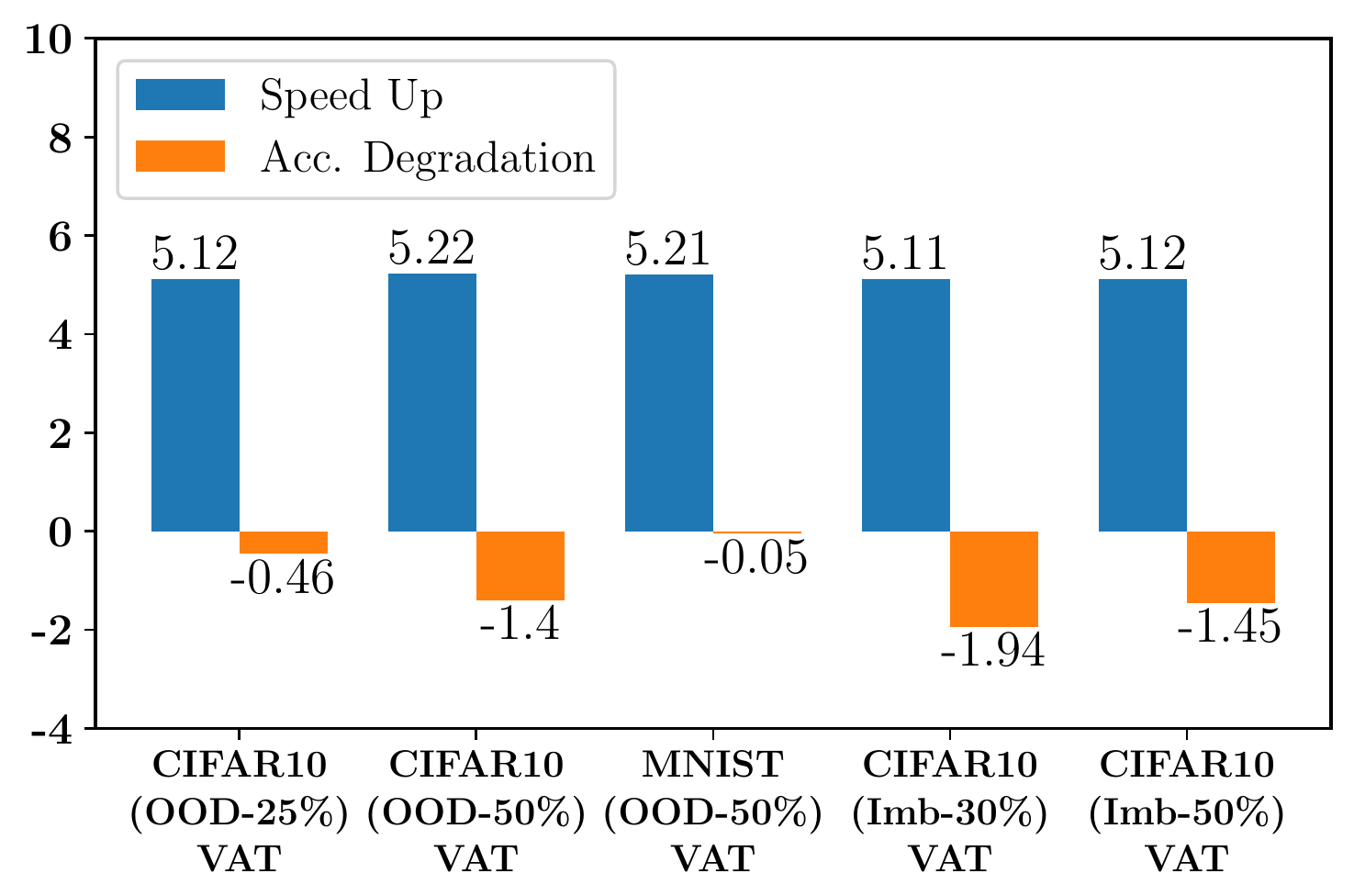} 
\caption*{$\underbracket[1pt][1.0mm]{\hspace{6.2cm}}_{\substack{\vspace{-4.0mm}\\
\colorbox{white}{\scriptsize (b) Robust SSL}}}$}
\phantomcaption
\label{fig:main-rob}
\end{subfigure}\quad
\caption{\small{Comparison of \model{} with VAT, FixMatch, and MT on CIFAR-10 and SVHN: We contrast the accuracy degradation with speedup compared to the base SSL or robust SSL (DS3L) approach. We observe speedups of $3\times$ in standard SSL case with $0.7\%$ accuracy drop and $2\times$ speedup with no accuracy drop. In the robust SSL case, we observe $5\times$ speedup compared to DS3L~\cite{pmlr-v119-guo20i} while outperforming it in terms of accuracy.}}
\label{fig:mainresults}
\end{figure}

To this end, we propose \model{}, a coreset selection framework that enables faster convergence and robust training of SSL algorithms. \model{} selects coreset of the unlabeled data resulting in minimum labeled set loss when trained upon in a semi-supervised manner. Intuitively, \model{} tries to achieve faster convergence by selecting data instances from the unlabeled set whose gradients are aligned with the labeled set gradients. Furthermore,  \model{} also achieves distribution matching by selecting a coreset from the unlabeled set with similar gradients to the labeled set.

\subsection{Our Contributions}
The contributions are our work can be summarized as follows:
\begin{enumerate}[leftmargin=*, label={\large\textbullet}]
\item \textbf{\model{} Framework: } We propose a coreset selection algorithm \model{} for efficient and robust semi-supervised learning. \model{} poses the coreset selection as a discrete-continuous bi-level optimization problem and solves it efficiently using an online approximation of single-step gradient updates. Essentially, \model{} selects a coreset of the unlabeled set, which, when trained using the combination of the labeled set and the specific unlabeled data coreset, minimizes the model loss on the labeled dataset. We also discuss several implementation tricks to speed up the coreset selection step significantly ({\em c.f.}, \secref{taylor-series}, \secref{implementation})
\item \textbf{\model{} in Traditional SSL: } We empirically demonstrate the effectiveness of \model{} in conjunction with several SOTA SSL algorithms like VAT, Mean-Teacher, and FixMatch. The speedups obtained by \model{} are shown in \figref{fig:main-eff}. Specifically, we see that \model{} consistently achieves close to $3\times$ speedup with accuracy degradation of around 0.7\%. \model{} also achieves more than $4.2\times$ speedup with a slightly higher accuracy degradation. Furthermore, when \model{} is trained for more iterations, \model\ can match the performance of VAT while having a $2\times$ speedup (see VAT Extended bar plot in \figref{fig:main-eff}). \model{} also consistently outperforms simple baselines like early stopping and random sampling.  
\item \textbf{\model{} in Robust SSL: } We further demonstrate the utility of \model{} for robust SSL in the presence of OOD data and imbalance in the unlabeled set. We observe that with the VAT SSL algorithm, \model{} outperforms SOTA robust SSL method DS3L~\cite{pmlr-v119-guo20i} (with VAT) while being around $5\times$ faster. \model{} also significantly outperforms just VAT and random sampling. 
\end{enumerate}

\subsection{Related Work}
\looseness-1
\noindent \textbf{Semi-supervised learning: } Several papers have been proposed for semi-supervised learning over the past few years. Due to space constraints, we do not talk about generative~\cite{NIPS2016_8a3363ab, NIPS2016_eb86d510, NIPS2014_d523773c, 10.5555/3104482.3104598, NIPS2007_4b6538a4, NIPS2017_9ef2ed4b, 10.5555/2980539.2980616} and graph-based~\cite{gaussianSSL, liu2019deep} methods for SSL in this work. We instead focus on the main components of the existing SOTA SSL algorithms, viz., a) consistency regularization and b) entropy minimization. The consistency regularization component forces the model to have consistent prediction given an unlabeled data point and its perturbed (or augmented) version. The Entropy-minimization component forces the model instances to have low-entropy predictions on unlabeled data instances to ensure that the classes are well separated. One can achieve entropy minimization by directly adding the entropy loss component on the unlabeled class prediction or using methods like Pseudo-Labeling to enforce it implicitly. Mean-Teacher~\cite{tarvainen2018mean} approach uses a consistency regularization component that forces the predictions of the exponential moving average of the model to be the same as the model prediction of the augmented unlabeled images. VAT~\cite{miyato2018virtual} instead computes the perturbation of the unlabeled data point that changes the prediction distribution the most and enforces the model to have the same prediction on both unlabeled data instance and unlabeled data instance with computed perturbation as a form of consistency regularization. MixMatch~\cite{berthelot2019mixmatch} uses $K$ standard image augmentations for consistency regularization and enforces entropy minimization by using a sharpening function on the average predicted distribution of $K$ augmentations of unlabeled data instances. FixMatch~\cite{sohn2020fixmatch} induces consistency regularization by forcing the model to have the same prediction on a weakly augmented and strongly augmented image instance. Furthermore, FixMatch~\cite{sohn2020fixmatch} also employs confidence thresholding to mask unlabeled data instances on which the model's prediction confidence is below a threshold from being used in consistency loss. 

\looseness-1
\noindent \textbf{Robust Semi-supervised learning: } Several methods have been proposed to make the existing semi-supervised learning algorithms robust to label noise in labeled data and robust to OOD data in the unlabeled set. A popular approach~\cite{ren2018learning,shu2019metaweightnet} to deal with label noises and class imbalance in a supervised learning setting is by reweighing each data instance and jointly learning these weights along with the model parameters. Safe-SSL (DS3L)~\cite{pmlr-v119-guo20i} is a recently proposed SOTA method for robust SSL learning. DS3L is similar to the reweighting in the supervised case and adopts a reweighting approach to deal with OOD data in the unlabeled set. Safe-SSL uses a neural network to predict the weight parameters of unlabeled instances that result in maximum labeled set performance, making it a bi-level optimization problem. In this regard, both \model{} and Safe-SSL approach solves a bi-level optimization problem, except that \model{} solves a discrete optimization problem at the outer level, thereby enabling significant speedup compared to SSL algorithms and an even more considerable speedup compared to safe-SSL (which itself is $3\times$ slower than SSL algorithms). In contrast to safe-SSL and other robust SSL approaches, \model{} achieves both efficiency and robustness. Other approaches for robust SSL include UASD~\cite{Chen_Zhu_Li_Gong_2020} uses an Uncertainty aware self-distillation with OOD filtering to achieve robust performance and a distributionally robust model to deal with OOD~\cite{chen2019distributionally}. 

\looseness-1
\noindent \textbf{Coreset and subset selection methods: } Coresets \cite{feldman2020core} are small and informative weighted data subsets that approximate original data.  Several works~\cite{wei2015submodularity, mirzasoleiman2020coresets, killamsetty2021glister, killamsetty2021gradmatch} have studied coresets for efficient training of deep learning models in the supervised learning scenarios. \textsc{Craig}~\cite{mirzasoleiman2020coresets} selects representative coresets of the training data that closely estimates the full training gradient. Another approach, GLISTER~\cite{killamsetty2021glister} posed the coreset selection as optimizing the validation set loss for efficient learning focused on generalization. Another approach, \textsc{Grad-Match}~\cite{killamsetty2021gradmatch} select subsets that approximately match the full training loss or validation loss gradient using orthogonal matching pursuit. Similarly, coreset selection methods \cite{wei2015submodularity, sener2018active, Ash2020Deep, killamsetty2021glister} were also used for active learning scenario, where a subset of data instances from the unlabeled set is selected to be labeled. Finally, several recent works have used submodular functions for finding diverse and representative subsets for data subset selection~\cite{liu2015svitchboard,kaushal2019learning,wei2015submodularity,wei2014submodular}.

\section{Preliminaries}
\noindent \textbf{Notation: }Denote $\Dcal = \{x_i, y_i\}_{i=1}^n$  to be the labeled set with $n$ labeled data points, and $\Ucal = \{x_j\}_{j=1}^m$ to be the unlabeled set with $m$ data points. Let $\theta$ be the classifier model parameters, $l_s$ be the labeled set loss function (such as cross-entropy loss) and $l_u$ be the unlabeled set loss, e.g. consistency-regularization loss, entropy loss, etc.. Denote $L_S(\Dcal, \theta) = \underset{i \in \Dcal}{\sum}l_{s}(\theta, x_i, y_i)$ and $L_U(\Ucal, \theta, \mb) = \underset{j \in \Ucal}{\sum} \mb_i l_u(x_j, \theta)$ where $\mb \in \{0, 1\}^m$ is the binary mask vector for unlabeled set. For notational convenience, we denote $l_{si}(\theta) = l_s(x_i, y_i, \theta)$ and denote $l_{uj}(\theta) = l_u(x_j, \theta)$.

\begin{figure}[t]
\begin{center}
\includegraphics[width = 0.95\textwidth, height=1.75cm]{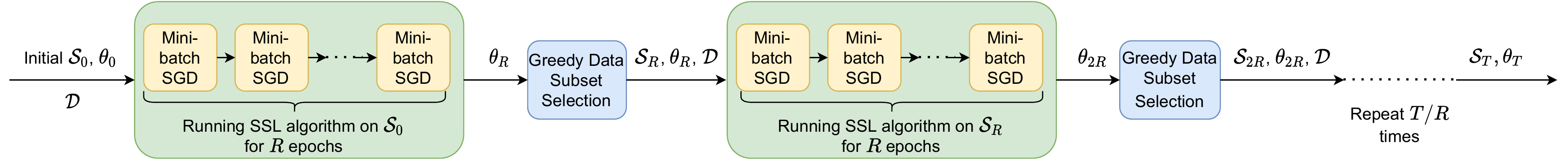}
\caption{Flowchart of \model{} framework, where coreset selection is performed every $R$ epochs and the model is trained on the selected coreset.}
\label{retrieve}
\end{center}
\end{figure}

\noindent \textbf{Semi-supervised loss: }
Following the above notations, the loss function for many existing SSL algorithms can be written as $L_S(\Dcal, \theta) + \lambda L_U(\Ucal, \theta, \mb)$, where $\lambda$ is the regularization coefficient for the unlabeled set loss. For Mean Teacher~\cite{tarvainen2018mean}, VAT\cite{miyato2018virtual}, MixMatch~\cite{berthelot2019mixmatch}, the mask vector $\mb$ is made up entirely of ones, whereas for FixMatch~\cite{sohn2020fixmatch}, $\mb$ is confidence-thresholded binary vector, indicating whether to include an unlabeled data instance or not. Usually, $L_S$ is a cross-entropy loss for classification experiments and squared loss for regression experiments. A detailed formulation of the loss function $L_U$ used in different SSL algorithms is given in Appendix~\ref{app:loss_form}

\noindent \textbf{Robust Semi-supervised loss: }
However, for robust semi-supervised loss, the mask vector $\mb$ is replaced with a weight vector $\wb \in \mathbb{R}^m$ denoting the contribution of data instances in the unlabeled set. The weight vector $\wb$ is unknown and needs to be learned. The weighted SSL loss is: $L_S(\Dcal, \theta) + \lambda L_U(\Ucal, \theta, \wb)$, where $\lambda$ is the regularization coefficient for the unlabeled set loss. 

The state-of-the-art robust SSL method, Safe-SSL~\cite{pmlr-v119-guo20i} poses the learning problem as:
\begin{equation}
\begin{aligned}
    \label{safe-ssl-obj}
    \overbrace{\wb^* = \underset{\wb}{\argmin}L_S(\Dcal, \underbrace{\underset{\theta}{\argmin}\left(L_S(\Dcal, \theta) + \lambda L_U(\Ucal, \theta, \wb)\right)}_{inner-level})}^{outer-level}\text{\hspace{1.7cm}}
\end{aligned}
\end{equation}

In order to solve the problem at the inner level efficiently, Safe-SSL~\cite{pmlr-v119-guo20i} method uses a single-gradient step approximation to estimate the inner problem solution. The weight vector learning problem after the one-step approximation is: 
$\wb^* = \underset{\wb}{\argmin}L_S(\Dcal, \theta - \alpha \nabla_{\theta}L_S(\Dcal, \theta) - \alpha \lambda \nabla_{\theta}L_U(\Ucal, \theta, \wb)$

Safe-SSL also uses a single-step gradient approximation to solve the outer level problem as well. As discussed before, the optimization problem of the Safe-SSL~\cite{pmlr-v119-guo20i} algorithm involves continuous optimization at both inner and outer levels, whereas for \model{}, the outer level involves a discrete optimization problem which makes it significantly faster than Safe-SSL. 
\looseness -1
\section{\model{} framework}
In \model{}, the coreset selection and classifier model learning on the selected coreset is performed in conjunction. As shown in the \figref{retrieve}, \model{} trains the classifier model on the previously selected coreset for $R$ epochs in a semi-supervised manner, and every $R^{th}$ epoch, a new coreset is selected, and the process is repeated until the classifier model reaches convergence, or the required number of epochs is reached. \emph{The vital feature of \model{} is that the coresets selected are adapted with the training.} Let $\theta_t$ be the classifier model parameters and the $\Scal_t$ be the coreset at time step $t$. Since coreset selection is done every $R$ epochs, we have $\Scal_{t} = \Scal_{\floor{t/R}}$, or in other words, the subsets change only after $R$ epochs. The SSL loss function on the selected coreset at iteration $t$ is as follows:
\begin{equation}
    \label{coreset-ssl}
        L_S(\Dcal, \theta_t) + \lambda_t \underset{j \in \Scal_t}{\sum} \mb_{jt} l_u(x_j, \theta_t)
\end{equation}
where $\mb_{jt}$ is the mask binary value associated with the $j^{th}$ point based on model parameters $\theta_t$ and $\lambda_t$ is the unlabeled loss coefficient at iteration $t$. Note that objective function given in \eqref{coreset-ssl} is dependent on the SSL algorithm used in \model{} framework. If gradient descent is used for learning, the parameter update step from time step $t$ to $t+1$ is as follows:
\begin{equation}
    \label{coreset-param-update}
        \theta_{t+1} = \theta_{t} - \alpha_t \nabla_{\theta} L_S(\Dcal, \theta_t) - \alpha_t \lambda_t  \underset{j \in \Scal_t}{\sum} \mb_{jt}  \nabla_{\theta} l_u(x_j, \theta_t)  
\end{equation}
where $\alpha_t$ is the learning rate at iteration $t$. The update step for mini-batch SGD is similar, just that it does the above on minibatches of the dataset.

\subsection{Problem Formulation}
The coreset selection problem of \model{} at timestep $t$ is as follows:

\begin{equation}
\begin{aligned}
    \label{retrieve-obj}
    \overbrace{\Scal_{t} = \underset{\Scal \subseteq \Ucal:|\Scal| \leq k}{\argmin}L_S\Big(\Dcal, \underbrace{\underset{\theta}{\argmin}\big(L_S(\Dcal, \theta_t) + \lambda_t \underset{j \in \Scal}{\sum} \mb_{jt} l_u(x_j, \theta_t) \big)}_{inner-level}\Big)}^{outer-level}\text{\hspace{1.7cm}}
\end{aligned}
\end{equation}
where $k$ is the size of the coreset and $\mb_{jt}$ is the binary value associated with the $j^{th}$ instance based on model parameters $\theta_t$. $k$ is a fraction of the entire dataset (e.g. 20\% or 30\%), and the goal is to select the best subset of the unlabeled set, which maximizes the labeled loss based. The outer level of the above optimization problem is a discrete subset selection problem. However, solving the inner-optimization problem naively is computationally intractable, and, we need to make some approximations.

\subsection{One-Step Gradient Approximation} 
To solve the inner optimization problem efficiently, \model{} adopts a one-step gradient approximation based optimization method similar to~\cite{finn2017model,ren2018learning}. More specifically, \model{} approximates the solution to the inner level problem by taking a single gradient step towards the descent direction of the loss function. The idea here is to jointly optimize the model parameters and the subset as the learning proceeds. After this approximation, the coreset selection optimization problem becomes:

\begin{equation}
\begin{aligned}
    \label{retrieve-meta-approx}
    \Scal_{t} = \underset{\Scal \subseteq \Ucal:|\Scal| \leq k}{\argmin}L_S(\Dcal, \theta_t - \alpha_t \nabla_{\theta}L_S(\Dcal, \theta_t) - \alpha_t \lambda_t \underset{j \in \Scal}{\sum} \mb_{jt} \nabla_{\theta}l_u(x_j, \theta_t))\text{\hspace{1.7cm}}
\end{aligned}
\end{equation}

However, even after this approximation, the above optimization problem (\eqref{retrieve-meta-approx}) is NP-hard.
\begin{theorem}
\label{thm:submod-proof}
Optimization problem (\eqref{retrieve-meta-approx}) is NP hard, even if $l_s$ is a convex loss function. If the labeled set loss function $l_s$ is cross-entropy loss, then the optimization problem give in the \eqref{retrieve-meta-approx} can be converted into an instance of cardinality constrained weakly submodular maximization. 
\end{theorem}

The proof is given in Appendix~\ref{app:submod}. The given Theorem~\ref{thm:submod-proof} holds as long as $l_s$ is a cross-entropy loss irrespective of the form of $l_u$. Further, Theorem~\ref{thm:submod-proof} implies that the optimization problem given in \eqref{retrieve-meta-approx} can be solved efficiently using greedy algorithms \cite{minoux1978accelerated, mirzasoleimanstochastic} with approximation guarantees. \model{} uses stochastic-greedy algorithm~\cite{khanna2017scalable,mirzasoleimanstochastic} to solve the optimization problem \eqref{retrieve-meta-approx} with an approximation guarantee of $1-1/e^{\beta} - \epsilon$ in $\Ocal(m \log(1/\epsilon))$ iterations where $m$ is the unlabeled set size and $\beta$ is the weak submodularity coefficient (see Appendix~\ref{app:submod}). And the set function used in stochastic greedy algorithm is as follows:
\begin{equation}
\begin{aligned}
    \label{retrieve-setf}
    f(\theta_t, \Scal) = -L_S(\Dcal, \theta_t - \alpha_t \nabla_{\theta}L_S(\Dcal, \theta_t) - \alpha_t \lambda_t \underset{j \in \Scal}{\sum} \mb_{jt} \nabla_{\theta}l_u(x_j, \theta_t))\text{\hspace{1.7cm}}
\end{aligned}
\end{equation}

Notice that during each greedy iteration, we need to compute the set function value $f(\theta_t, \Scal \cup e)$ to find the maximal gain element $e$ that can be added to the set $\Scal$. This implies that the loss over the entire labeled set needs to be computed multiple times for each greedy iteration, making the entire greedy selection algorithm computationally expensive.

\subsection{\model\ Algorithm} \label{taylor-series}
To make the greedy selection algorithm efficient, we approximate the set function value $f(\theta_t, \Scal \cup e)$ with the first two terms of it's Taylor-series expansion 
Let, $\theta^{S} = \theta_t - \alpha_t \nabla_{\theta}L_S(\Dcal, \theta_t) - \alpha_t \lambda_t \underset{j \in \Scal}{\sum} \mb_{jt} \nabla_{\theta}l_u(x_j, \theta_t)$. The modified set function value with Taylor-series approximation is as follows:
\begin{equation}
\begin{aligned}
    \label{retrieve-taylor-setf}
    \hat{f}(\theta_t, \Scal \cup e) = -L_S(\Dcal, \theta^{S}) + \alpha_t \lambda_t  {\nabla_{\theta}L_S(\Dcal, \theta^S)}^T  \mb_{et} \nabla_{\theta}l_u(x_e, \theta_t)\text{\hspace{1.7cm}}
\end{aligned}
\end{equation}

where $\mb_{et}$ is the binary mask value associated with element $e$. Note that the term $\mb_{et} \nabla_{\theta} l_u(x_e, \theta_t)$ can be precomputed at the start of the greedy selection algorithm, and the term $\nabla_{\theta}L_S(\Dcal, \theta^S)$ needs to be computed only once every greedy iteration, thereby reducing the computational complexity of the greedy algorithm.
\begin{wrapfigure}[19]{R}{0.5\textwidth}
\vspace{-9mm}
\begin{algorithm}[H]
\SetCustomAlgoRuledWidth{0.45\textwidth}
\LinesNotNumbered
 \DontPrintSemicolon
 \KwIn{Labeled Set: $\Dcal$, Unlabeled Set: $\Ucal$, Reg. Coefficients: $\{\lambda_t\}_{t=0}^{t=T-1}$, Learning rates: $\{\alpha_t\}_{t=0}^{t=T-1}$}
 \KwIn{Total no of epochs: $T$, Epoch interval for subset selection: $R$, Size of the coreset: $k$}
 \SetKwBlock{Begin}{function}{end function}{
    Set $t=0$; Randomly initialize model parameters $\theta_0$ and coreset $\Scal_{0} \subseteq \Ucal: |\Scal_{0}| = k$; \;
    \Repeat{$t \ge T$}
    {
        \If {$(t\%R == 0) \land (t > 0)$}{
        $\Scal_{t}$ = GreedySelection($\Dcal, \Ucal, \theta_{t}, \lambda_{t}, \alpha_{t}, k$)\;} 
        \Else { $\Scal_{t} = \Scal_{t-1}$\;}
        Compute batches $\Dcal_b = ((x_b, y_b); b \in (1 \cdots B))$ from $\Dcal$ \;
        Compute batches $\Scal_{tb} = ((x_b); b \in (1 \cdots B))$ from $\Scal$ \;
        \textcolor{gray}{*** Mini-batch SGD *** } \;
        Set $\theta_{t0} = \theta_{t}$ \;
        \For{$b = 1$ to $B$}
        {   
        Compute mask $\mb_{t}$ on $\Scal_{tb}$ from current model parameters $\theta_{t(b-1)}$\;
        $\theta_{tb} = \theta_{t(b-1)} - \alpha_t \nabla_{\theta}L_S(\Dcal_b, \theta_t) - \alpha_t \lambda_t \underset{j \in \Scal_{tb}}{\sum} \mb_{jt} \nabla_{\theta}l_u(x_j, \theta_t(b-1))$ \;
        }
        Set $\theta_{t+1} = \theta_{tB}$ \;
        $t = t+1$\;
    }
 }
 \Return{$\theta_{T}, \Scal_{T}$}
\caption{\model{} Algorithm}\label{ret_alg}
\end{algorithm}
\end{wrapfigure}

A detailed pseudo-code of the \model{} algorithm is given in \AlgRef{ret_alg}. \model{} uses a greedy selection algorithm for coreset selection, and the detailed pseudo-code of the greedy algorithm is given in \AlgRef{alg:coreset}. \model{} can be easily implemented with popular deep learning frameworks \cite{pytorch, tensorflow} that provide auto differentiation functionalities. In all our experiments, we set $R=20$, i.e., we update the coreset every 20 epochs.

\vspace{-2mm}
\subsection{Additional Implementation Details: }\label{implementation}
In this subsection, we discuss additional implementational and practical tricks to make \model{} scalable and efficient.

\noindent \textbf{Last-layer gradients. } 
Computing the gradients over deep models is time-consuming due to an enormous number of parameters in the model. To address this issue,  we adopt a last-layer gradient approximation similar to~\cite{Ash2020Deep,mirzasoleiman2020coresets,killamsetty2021glister, killamsetty2021gradmatch} by only considering the last classification layer gradients of the  classifier model in \model{}. By simply using the last-layer gradients, we achieve significant speedups in \model{}.
\begin{wrapfigure}[16]{R}{0.5\textwidth}
\vspace{-14mm}
\begin{algorithm}[H]
 \SetCustomAlgoRuledWidth{0.45\textwidth}
 \LinesNotNumbered
 \DontPrintSemicolon
 \KwIn{Labeled Set: $\Dcal$, Unlabeled Set: $\Ucal$, Model Parameters: $\theta_t$, Learning rate: $\alpha_t$}
 \KwIn{Regularization Coefficient: $\lambda_t$, Budget: $k$}
 \SetKwBlock{Begin}{function}{end function}{
    Initialize $\Scal_t = \emptyset$ \;
    Set $m = |\Ucal|$\;
    Compute mask values $\mb$ based on current model parameters $\theta_t$\;
    \For{$e \in \Ucal$}
    {
        Compute $\theta_e = \mb_e \nabla_{\theta}l_u(x_e, \theta_t) $
    }
    \For{$i=1$ to $k$}
    {
        Compute $\theta^S = \theta_t - \alpha_t \nabla_{\theta}L_S(\Dcal, \theta_t) - \alpha_t \lambda_t \underset{j \in \Scal}{\sum} \mb_{jt} \nabla_{\theta}l_u(x_j, \theta_t)$ \;
        Compute $\nabla_{\theta}L_S(\Dcal, \theta^S)$ \;
        $V\sim$ Sample $\ceil{mlog(1/\epsilon)}$ instances from $\Ucal$ \;
        $best-gain = -\infty$
        \For{$e \in V$}
        {
            Compute gain $\hat{g}(e) = \alpha_t \lambda_t  {\nabla_{\theta}L_S(\Dcal, \theta^S)}^T  \mb_{e} \nabla_{\theta}l_u(x_e, \theta_t)$ \;
            \If{$\hat{g}(e) > best-gain$}
            {
                Set $s = e$ \;
                Set $best-gain = \hat{g}(e)$
            }
        }
        $\Scal_t = \Scal_t \cup s$\;
        $\Ucal = \Ucal \setminus s$
    }
 }
 \Return{$\Scal_{t}$}
\caption{GreedySelection}\label{alg:coreset}
\end{algorithm}
\end{wrapfigure}

\vspace{-1ex}
\noindent \textbf{Warm-starting data selection: } We warm start the classifier model by training it on the entire unlabeled dataset for a few epochs similar to \cite{killamsetty2021gradmatch}. Warm starting allows the classifier model to have a good starting point to provide informative loss gradients used for coreset selection. More specifically, we train the classifier model on the entire unlabeled set for $T_w = \frac{\kappa T k}{m}$ epochs where $k$ is coreset size, $T$  is the total number of epochs, $\kappa$ is the fraction of warm start, and $m$ is the size of the unlabeled set. To be fair, we consider all baselines in the standard SSL setting with the same warm start. 

\begin{figure*}
\centering
\includegraphics[width = 12cm, height=0.5cm]{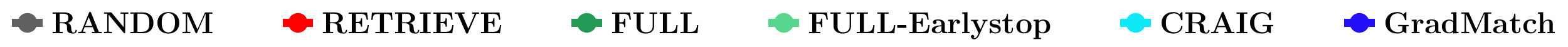}
\captionsetup{labelformat=empty}
\hspace{-0.6cm}
\begin{subfigure}[b]{0.24\textwidth}
\centering
\includegraphics[width=2.8cm, height=2.3cm]{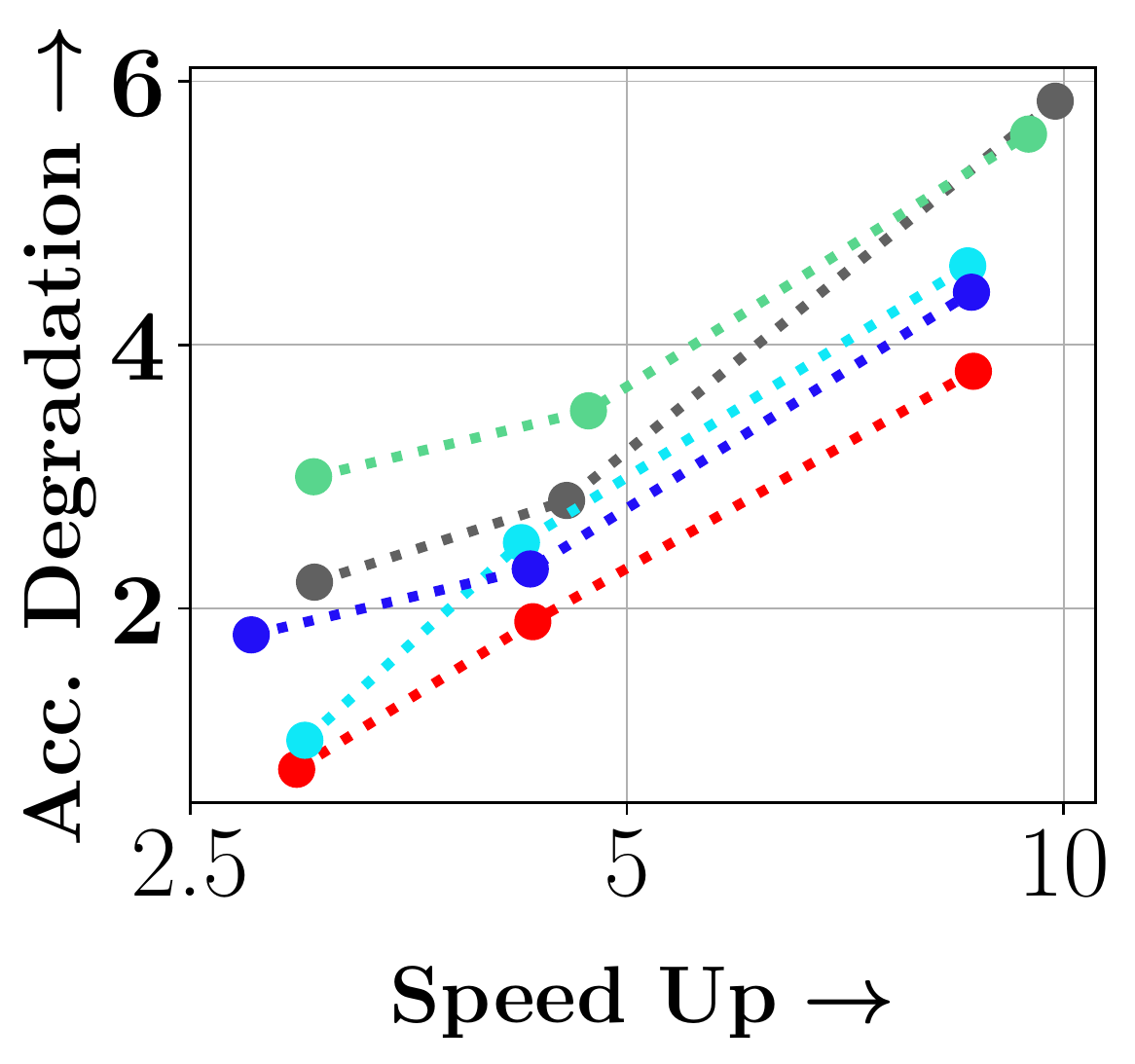}
\caption*{$\underbracket[1pt][1.0mm]{\hspace{2.8cm}}_{\substack{\vspace{-4.0mm}\\
\colorbox{white}{(a) \scriptsize CIFAR10-VAT}}}$}
\phantomcaption
\label{fig:cifar-vat}
\end{subfigure}
\begin{subfigure}[b]{0.24\textwidth}
\centering
\includegraphics[width=2.8cm, height=2.3cm]{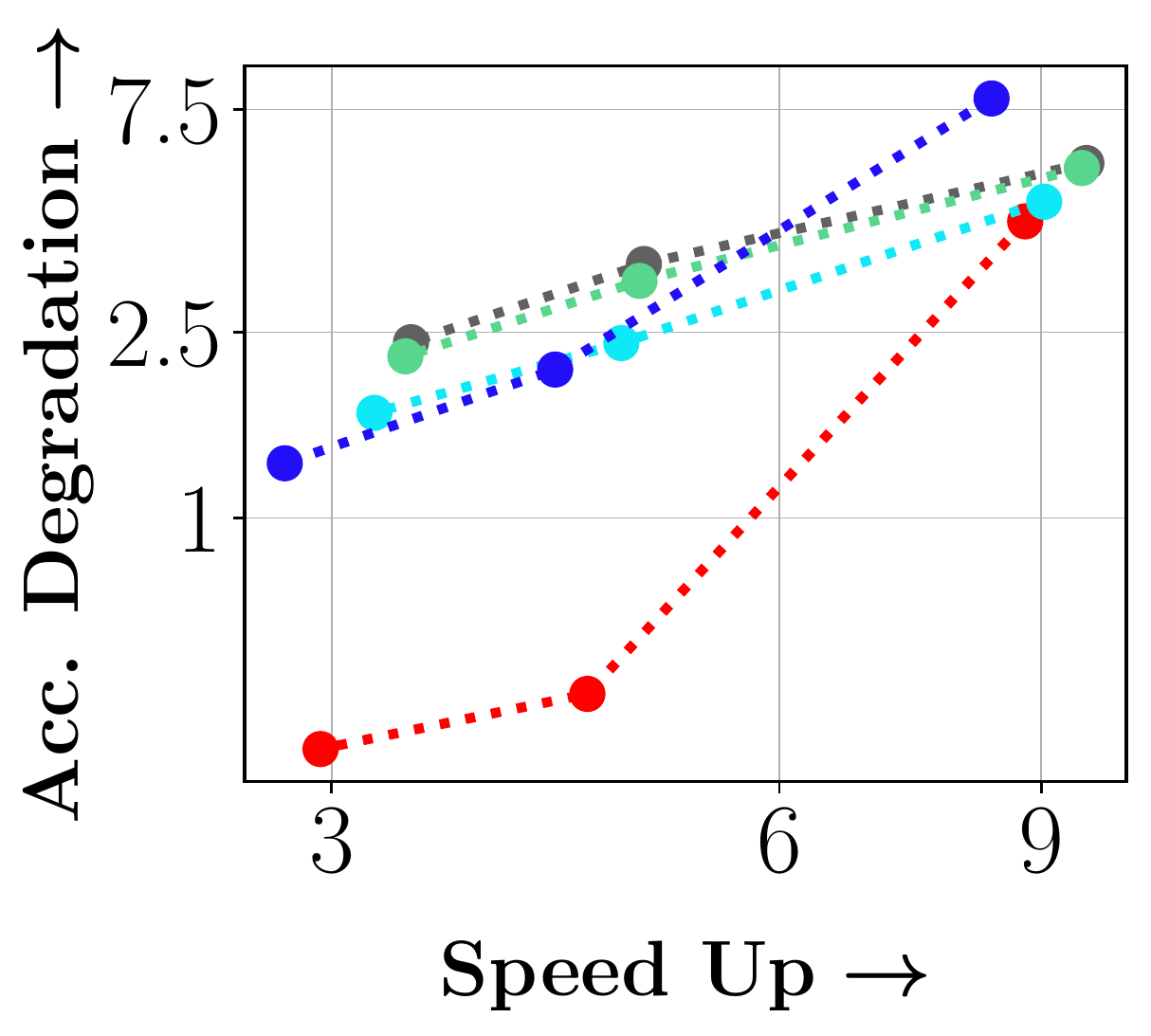}
\caption*{$\underbracket[1pt][1.0mm]{\hspace{2.8cm}}_{\substack{\vspace{-4.0mm}\\
\colorbox{white}{(b) \scriptsize SVHN-VAT}}}$}
\phantomcaption
\label{fig:svhn-vat}
\end{subfigure}
\begin{subfigure}[b]{0.24\textwidth}
\centering
\includegraphics[width=2.8cm, height=2.3cm]{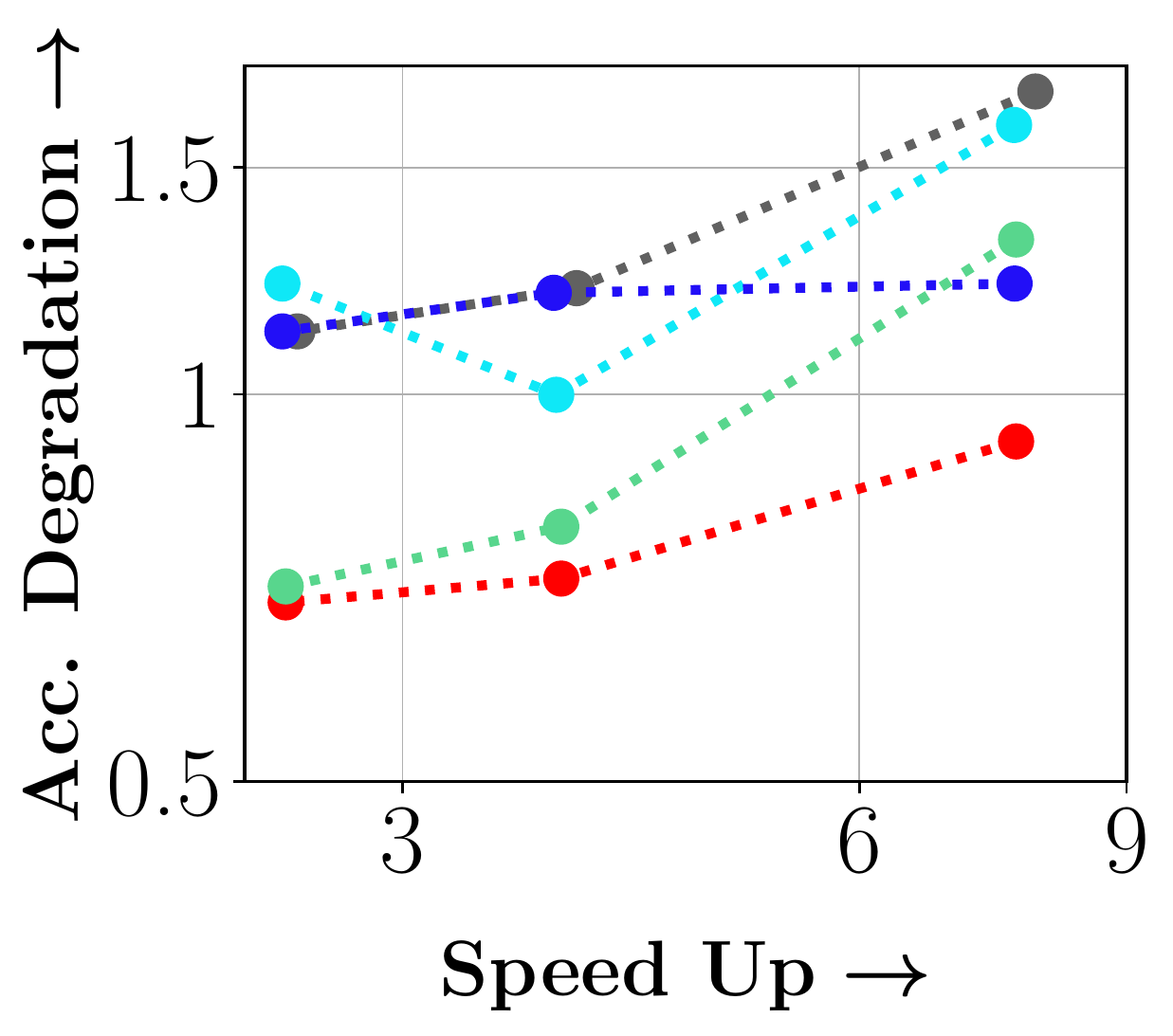}
\caption*{$\underbracket[1pt][1.0mm]{\hspace{3.2cm}}_{\substack{\vspace{-4.0mm}\\
\colorbox{white}{(c) \scriptsize CIFAR10-FixMatch}}}$}
\phantomcaption
\label{fig:cifar-fm}
\end{subfigure}
\begin{subfigure}[b]{0.24\textwidth}
\centering
\includegraphics[width=2.8cm, height=2.3cm]{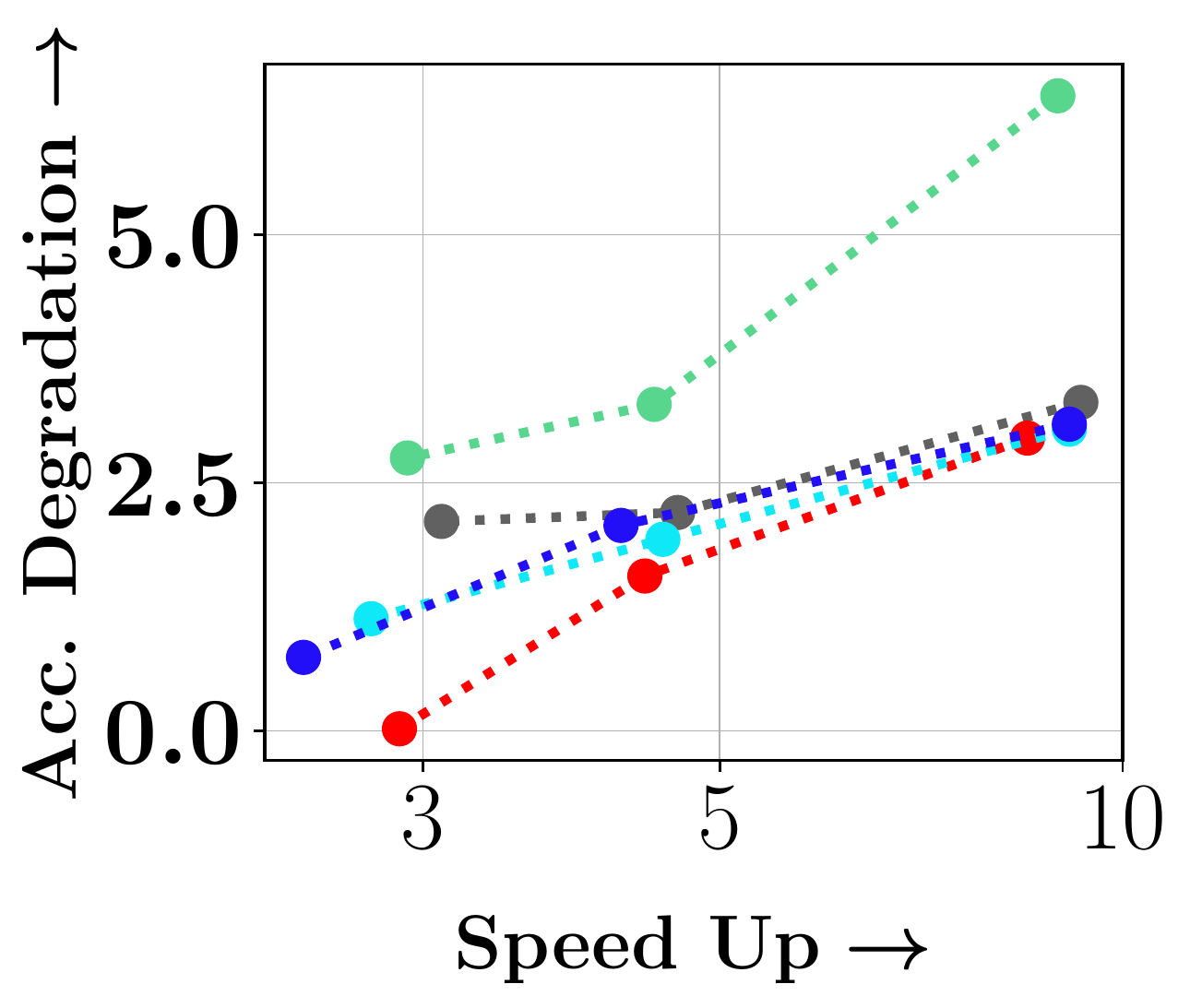}
\caption*{$\underbracket[1pt][1.0mm]{\hspace{2.8cm}}_{\substack{\vspace{-4.0mm}\\
\colorbox{white}{(d) \scriptsize CIFAR10-MT}}}$}
\phantomcaption
\label{fig:cifar-mt}
\end{subfigure}
\begin{subfigure}[b]{0.24\textwidth}
\centering
\includegraphics[width=2.8cm, height=2.3cm]{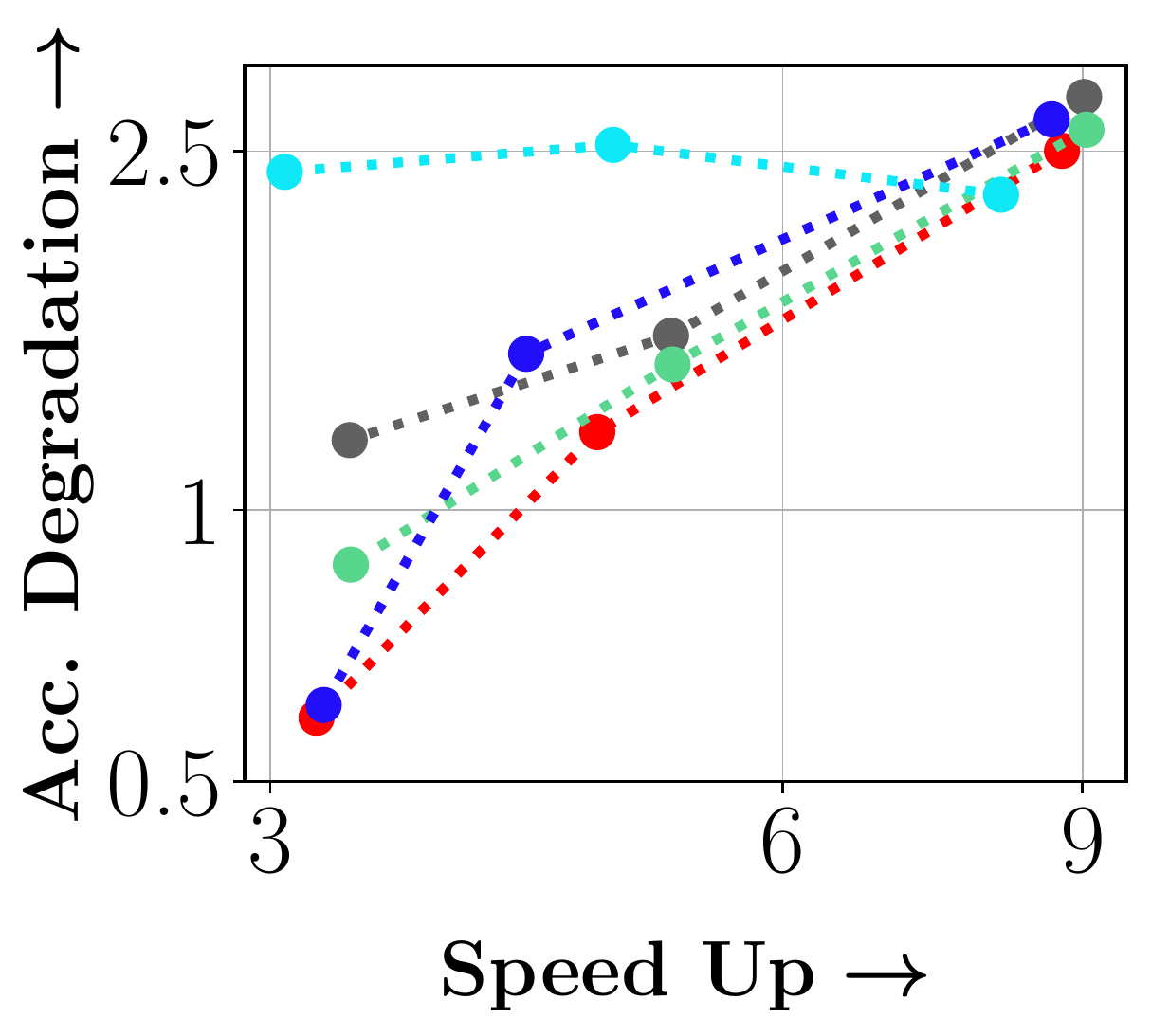}
\caption*{$\underbracket[1pt][1.0mm]{\hspace{2.8cm}}_{\substack{\vspace{-4.0mm}\\
\colorbox{white}{(e) \scriptsize SVHN-MT}}}$}
\phantomcaption
\label{fig:svhn-mt}
\end{subfigure}
\begin{subfigure}[b]{0.24\textwidth}
\centering
\includegraphics[width=2.8cm, height=2.3cm]{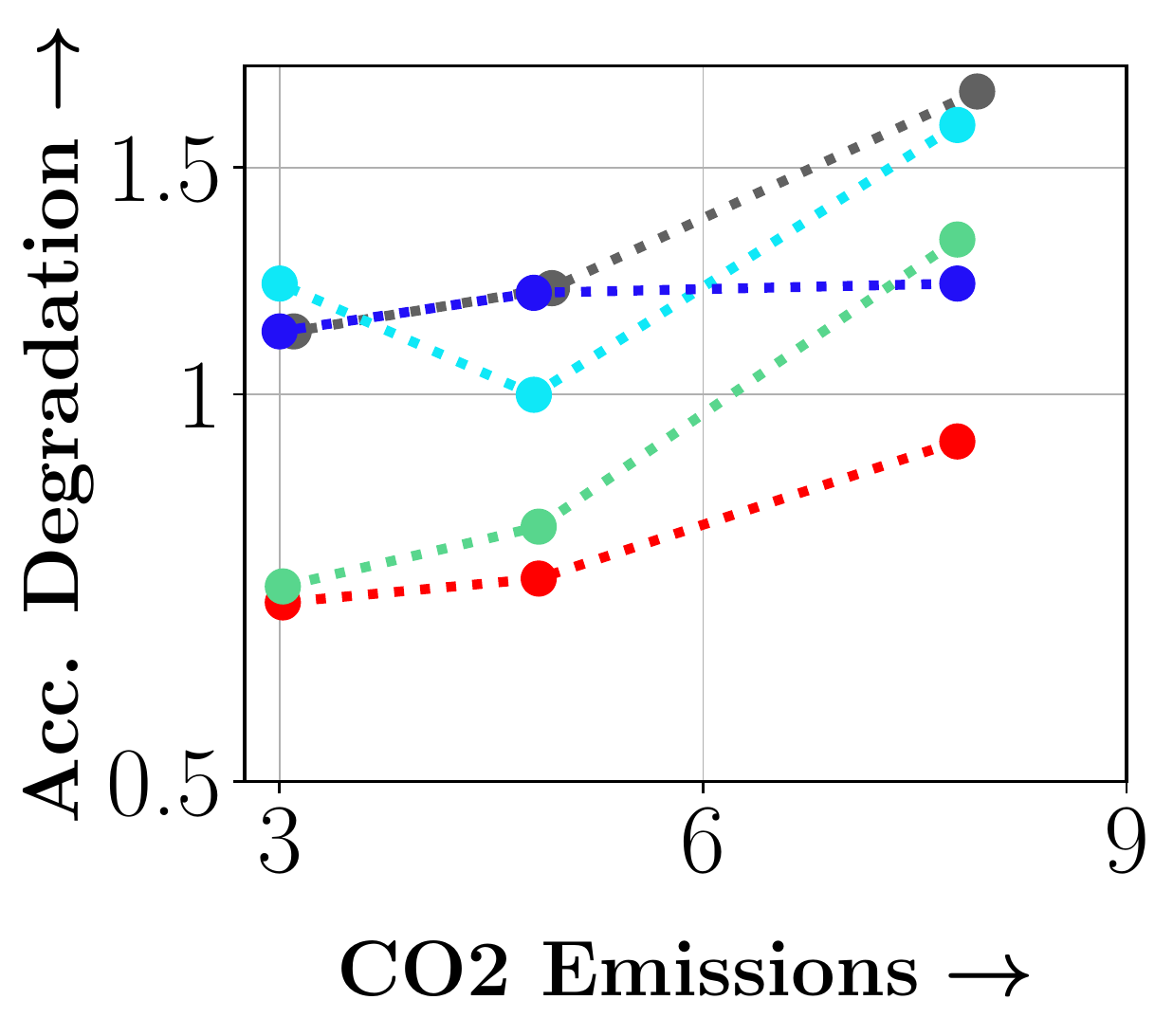}
\caption*{$\underbracket[1pt][1.0mm]{\hspace{2.8cm}}_{\substack{\vspace{-4.0mm}\\
\colorbox{white}{(f) \scriptsize FixMatch CO2 Emissions}}}$}
\phantomcaption
\label{fig:fm-co2}
\end{subfigure}
\begin{subfigure}[b]{0.24\textwidth}
\centering
\includegraphics[width=2.8cm, height=2.3cm]{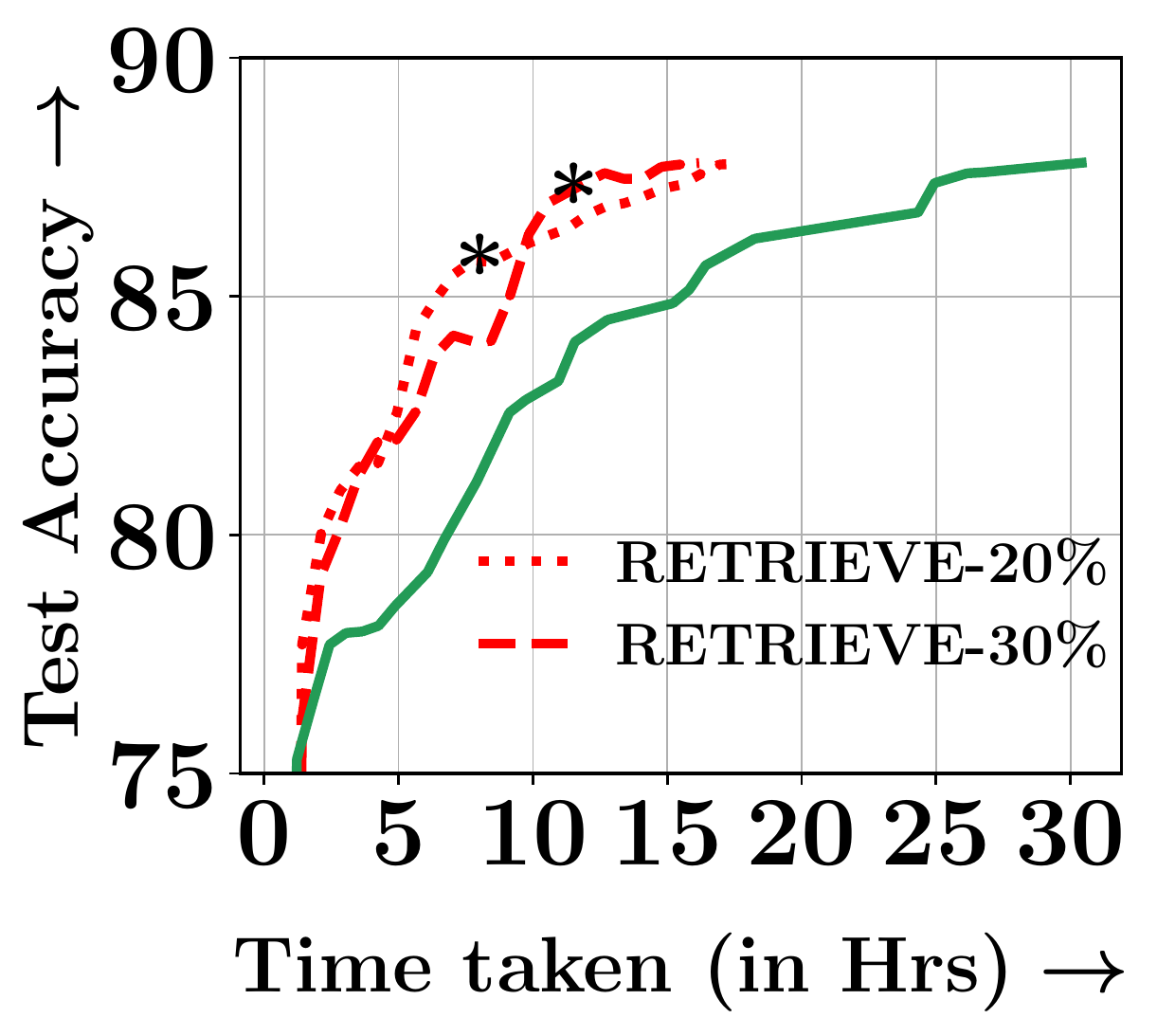}
\caption*{$\underbracket[1pt][1.0mm]{\hspace{2.8cm}}_{\substack{\vspace{-4.0mm}\\
\colorbox{white}{(g) \scriptsize VAT Convergence}}}$}
\phantomcaption
\label{fig:vat-convg}
\end{subfigure}
\begin{subfigure}[b]{0.24\textwidth}
\centering
\includegraphics[width=2.8cm, height=2.3cm]{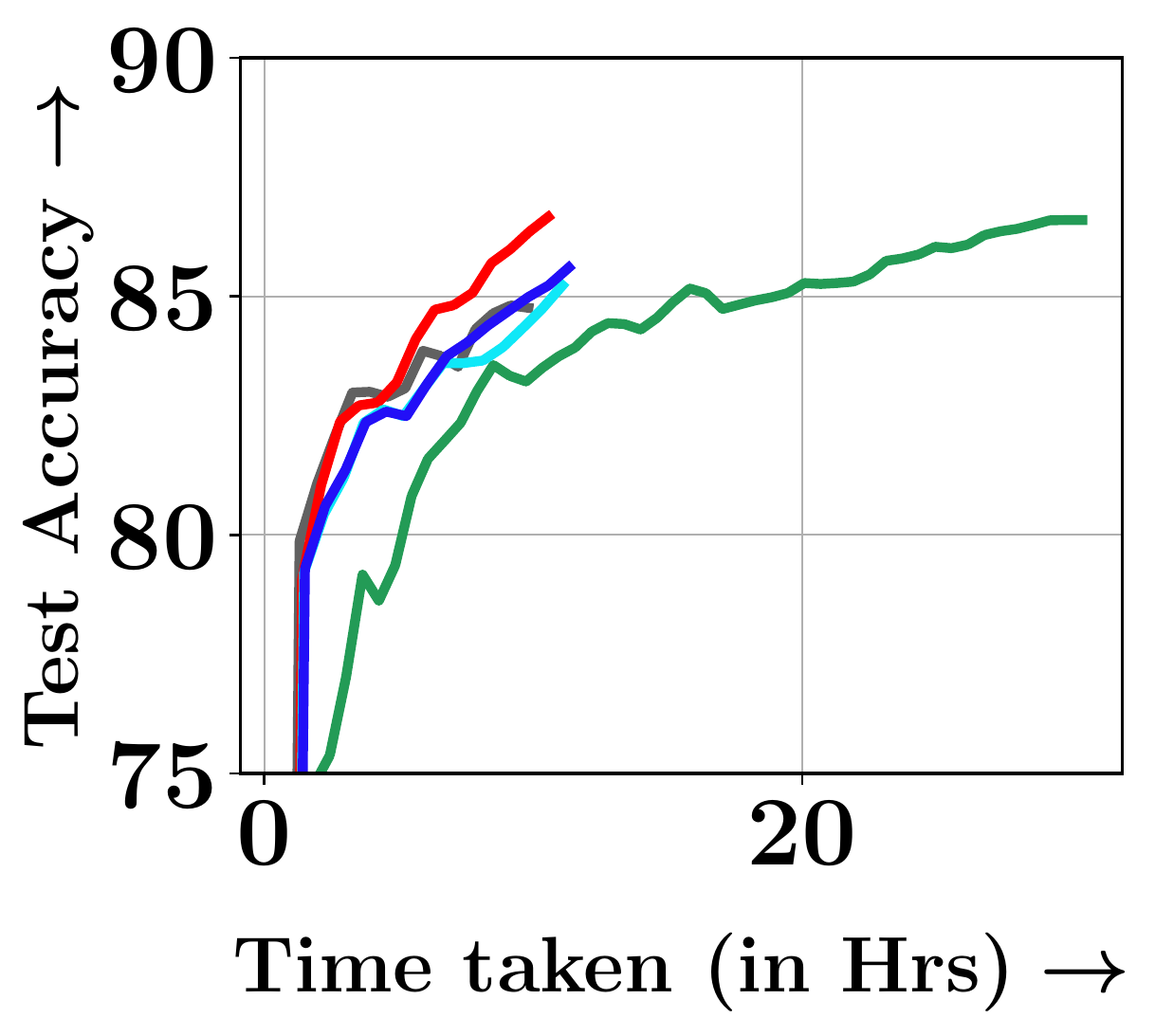}
\caption*{$\underbracket[1pt][1.0mm]{\hspace{3.2cm}}_{\substack{\vspace{-4.0mm}\\
\colorbox{white}{(h) \scriptsize MT Convergence}}}$}
\phantomcaption
\label{fig:mt-convg}
\end{subfigure}
\caption{\footnotesize{Figure 4: A comparison of \model\ with baselines (\textsc{Random, Craig}, \textsc{Full}, and \textsc{FullEarlyStop} in the traditional SSL setting. SpeedUp vs Accuracy Degradation, both compared to SSL for (a) VAT on CIFAR-10, (b) VAT on SVHN, (c) FixMatch on CIFAR-10, (d) MT on CIFAR-10, and (e) MT on SVHN. We observe that \model\ significantly outperforms existing baselines in terms of accuracy degradation and speedup tradeoff compared to SSL. Plot (f) also compares the CO2 emissions among the different approaches on FixMatch, showing again that \model\ achieves the best energy-accuracy tradeoff. Plots (g) and (h) are the convergence results comparing accuracy with the time taken. Again, we see that \model\ achieves much faster convergence than all baselines and full training.}}
\label{fig:efficient_ssl}
\end{figure*}

\subsection{Epochs vs. Iterations}
Most of the existing SSL algorithms are trained using a fixed number of iterations instead of epochs. However, for easier comprehension of the \model{} algorithm, we use the epoch notation in our work. A single epoch here meant a pass over random mini-batches of data points, such that the total number of data points encountered is equal to the size of the coreset of the unlabeled data. For example, if the unlabeled set size is 50000 and the unlabeled batch size is 50, then a single epoch over 100\%, 50\%, and 30\% subsets are equivalent to 1000, 500, and 300 iterations, respectively. 
\vspace{-1ex}
\section{Experiments}
Our experimental section aims to verify the efficiency and effectiveness of \model{} by evaluating \model{} through three semi-supervised learning scenarios a) traditional SSL scenario with clean data, b) robust SSL with OOD, and c) robust SSL with class imbalance, to demonstrate the efficiency and the robustness of \model{}. Furthermore, our work's experimental scenarios are very relevant in terms of research and real-world applications. We have implemented the \model{} algorithmic framework using PyTorch~\cite{paszke2017automatic}. We repeat the same experiment for three runs with different initialization and report the mean test accuracies in our plots. A detailed table with both mean test accuracy and the standard deviations was given in Appendix~(\ref{app:traditional_ssl}, \ref{app:robust_ssl}). For a fair comparison, we use the same random seed in each trial for all methods.  We explain implementation details, datasets, and baselines used in each scenario in the following subsections.

\textbf{Baselines in each setting.} In this section, we discuss baselines that are used in all the scenarios considered. We begin with the \textbf{traditional SSL scenario}. In this setting, we run \model{} (and all baselines) with warm-start. We incorporate \model\ with three representative SSL methods, including Mean Teacher (MT)~\cite{tarvainen2018mean}, Virtual Adversarial Training (VAT)~\cite{miyato2018virtual} and FixMatch~\cite{sohn2020fixmatch}. The baselines considered are \textsc{Random} (where we just randomly select a subset of unlabeled data points of the same size as \model), \textsc{Craig}~\cite{mirzasoleiman2020coresets,killamsetty2021gradmatch} and \textsc{Full-EarlyStop}. \textsc{Craig}~\cite{mirzasoleiman2020coresets,killamsetty2021gradmatch} was actually proposed in the supervised learning scenario. We adapt it to SSL by choosing a representative subset of unlabeled points such that the gradients are similar to the unlabeled loss gradients. We run the per-batch variant of \textsc{Craig} proposed in~\cite{killamsetty2021gradmatch}, where we select a subset of mini-batches instead of data instances for efficiency and scalability. Similarly, we use the per-batch version of \textsc{GradMatch} proposed in~\cite{killamsetty2021gradmatch} adapted to SSL setting as another baseline. For more information on the formulation of \textsc{Craig} and \textsc{GradMatch} in the SSL case, see Appendix~\ref{app:craig}, ~\ref{app:gradmatch}. Again, we emphasize that \textsc{Random}, \textsc{Craig}, and \textsc{GradMatch} are run with early stopping for the same duration as \model. In \textsc{Full-EarlyStop} baseline, we train the model on the entire unlabeled set for the time taken by \textsc{Retrieve} and report the test accuracy. In the traditional SSL scenario, we use warm variants of \model{}, \textsc{Random}, \textsc{Craig} for SSL training because warm variants are better in accuracy and efficiency compared to not performing warm start -- see Appendix~\ref{app:traditional_ssl} for a careful comparison of both.  \textbf{Robust SSL with OOD and Imbalance: } In the robust learning scenario for both OOD and imbalance, we analyze the performance of \model{} with the VAT~\cite{miyato2018virtual} algorithm. Note that for the Robust SSL scenario, we \emph{do not warm start the model by training for a few iterations on the full unlabeled set} because training on an entire unlabeled set(containing OOD or class imbalance) creates a biased model due to a distribution mismatch between labeled set and unlabeled set. We empirically compare not warm starting the model with warm starting in Appendix~\ref{app:robust_ssl}. In the robust SSL case, we compare \model\ with two robust SSL algorithms DS3L~\cite{pmlr-v119-guo20i} and L2RW~\cite{ren2018learning}. DS3L (also called Safe-SSL) is a robust learning approach using a meta-weight network proposed specifically for robust SSL. We adapt L2RW(Learning to Reweight), originally proposed for robust supervised learning, to the SSL case and use it as a baseline. Similarly, we adapt the robust coreset selection method \textsc{CRUST}~\cite{NEURIPS2020_crust} originally proposed to tackle noisy labels scenario in supervised learning to SSL setting and use it as a baseline in Robust SSL scenario.

\begin{wrapfigure}{L}{0.4\textwidth}
\centering
\includegraphics[width=3.5cm, height=3cm]{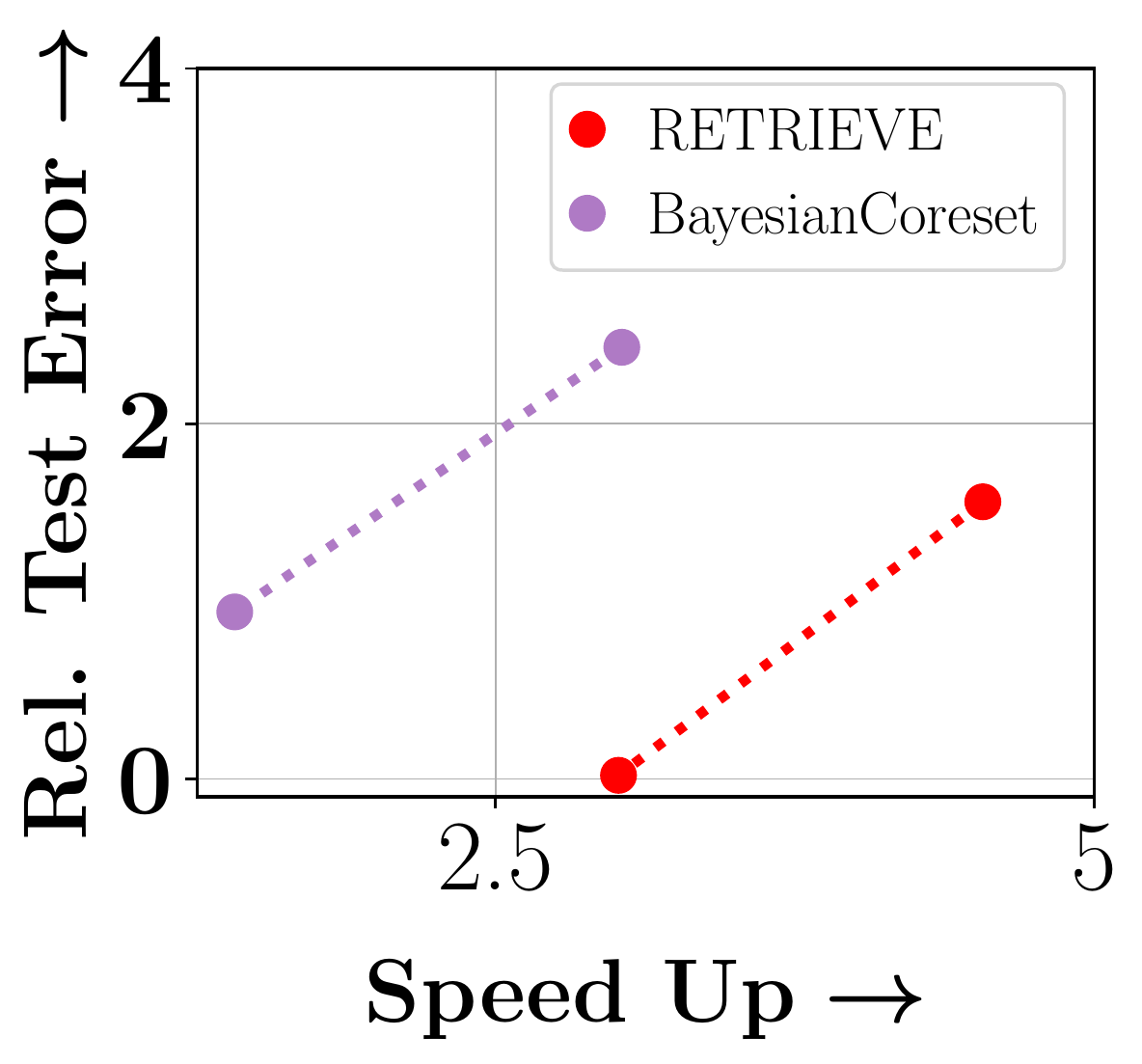}
\caption*{\scriptsize{Figure 5: Performance comparison of \model{} vs Bayesian Coreset using VAT for 20\% and 30\% CIFAR10 subsets}}
\phantomcaption
\label{fig:bcoreset}
\vspace{-3ex}
\end{wrapfigure}

\vspace{-1ex}
\textbf{Datasets, Model architecture and Experimental Setup: } We begin by providing details common across the three scenarios. We perform experiments on the following image classification datasets: CIFAR-10~\cite{krizhevsky2009learning} (60000 instances), SVHN~\cite{netzer2011reading} (99289 instances) and the following sentiment analysis datasets: IMDB~\cite{maas-etal-2011-learning}\footnote{https://ai.stanford.edu/~amaas/data/sentiment/} (10000 instances), and ELEC~\cite{Johnson2015SemisupervisedCN}\footnote{http://riejohnson.com/cnn\_data.html}(246714 instances) datasets. We use a modified version of the ELEC dataset where the duplicate sentences are removed. 
For CIFAR-10, we use a labeled set of 4000 instances with 400 instances from each class, an unlabeled set of 50000 instances, a test set of 10000 instances. For SVHN, we used a labeled set of 1000 instances with 100 instances from each class, an unlabeled set of 73257 instances, a test set of 26032 instances. For IMDB and ELEC datasets, we use the labeled and unlabeled splits following the work~\cite{MiyatoDG17}. For CIFAR10 and SVHN datasets, we use the Wide-ResNet-28-2~\cite{zagoruyko2017wide} model that is commonly used in SSL~\cite{oliver2019realistic,sohn2020fixmatch}. For MNIST, we use a variant of LeNet~\cite{lecun1989backpropagation} (see Appendix~\ref{app:experimental} for details). For IMDB and ELEC datasets, we use a model comprising of Word Embedding layer, LSTM model, and two-layer MLP model following the architecture given in the work~\cite{MiyatoDG17}. Similar to the work~\cite{MiyatoDG17}, we initialize the embedding matrix and LSTM model weights using a pretrained recurrent language model using both the labeled and unlabeled data. For Image datasets, with \model (and baselines like \textsc{Random}, \textsc{GradMatch} and \textsc{Craig}), we use the Nesterov's accelerated SGD optimizer with a learning rate of 0.03, weight decay of 5e-4, the momentum of 0.9, and a cosine annealing~\cite{LoshchilovH16a} learning rate scheduler for all the experiments. For the \textsc{Full} and \textsc{FullEarlyStop}, we use the Adam optimizer~\cite{kingma2014adam} and follow the experimental setting from the SSL papers~\cite{sohn2020fixmatch,miyato2018virtual,tarvainen2018mean}. For Text datasets, with \model (and baselines like \textsc{Random}, \textsc{GradMatch} and \textsc{Craig}), we set the hyperparameter values following the work~\cite{MiyatoDG17}. Similarly, for the robust SSL baselines, we use the settings from the corresponding papers~\cite{pmlr-v119-guo20i}. For all our experiments using image datasets, we use a batch size of 50 for labeled and unlabeled sets. Next, we discuss the specific settings for the \textbf{traditional SSL scenario}. For CIFAR10, we train the model for 500 epochs, and for SVHN, we train the model for 340 epochs on an unlabeled set. Note that we mention the epochs here because the number of iterations depends on the size of the unlabeled sets since that would determine the number of mini-batches. For a fair comparison, we train all algorithms for a fixed number of epochs. Next, we look at \textbf{robust SSL for OOD}. In this scenario, we consider the presence of OOD in the unlabeled set. We introduce OOD into CIFAR-10 following~\cite{oliver2019realistic}, by adapting it to a 6-class dataset, with 400 labels per class (from the 6 animal classes) as ID and rest of the classes as OOD (ID classes are: "bird", "cat", "deer", "dog", "frog", "horse", and OOD data are from classes: "airline", "automobile", "ship", "truck").  Similarly, we adapt MNIST~\cite{lecun-mnisthandwrittendigit-2010} to a 6-class dataset, with  classes 1-6 as ID and  classes 7-10 as OOD. We denote the OOD ratio=$\mathcal{U}_{ood}/(\mathcal{U}_{ood}+\mathcal{U}_{in})$ where $\mathcal{U}_{in}$ is ID unlabeled set, $\mathcal{U}_{ood}$ is OOD unlabeled set. For CIFAR10, we use a labeled set of 2400 instances and an unlabeled set of 20000 instances, and for MNIST, we use a labeled set of 60 instances and an unlabeled set of 30000 instances. Finally, for \textbf{robust SSL for class imbalance}, we consider imbalance both in the labeled set and unlabeled set. We introduce imbalance into the CIFAR-10 dataset by considering classes 1-5 as imbalanced classes and a class imbalance ratio. The class imbalance ratio is defined as the ratio of instances from classes 1-5 and the number of instances from classes 6-10. For CIFAR-10, we use a labeled set of 2400 and an unlabeled set of 20000 instances.
 
\begin{wrapfigure}{R}{0.6\textwidth}
\centering
\includegraphics[width = 8cm, height=0.4cm]{figs/legend_notbold_eff.pdf}
\captionsetup{labelformat=empty}
\hspace{-0.6cm}
\begin{subfigure}[b]{0.24\textwidth}
\centering
\includegraphics[width=2.8cm, height=2.3cm]{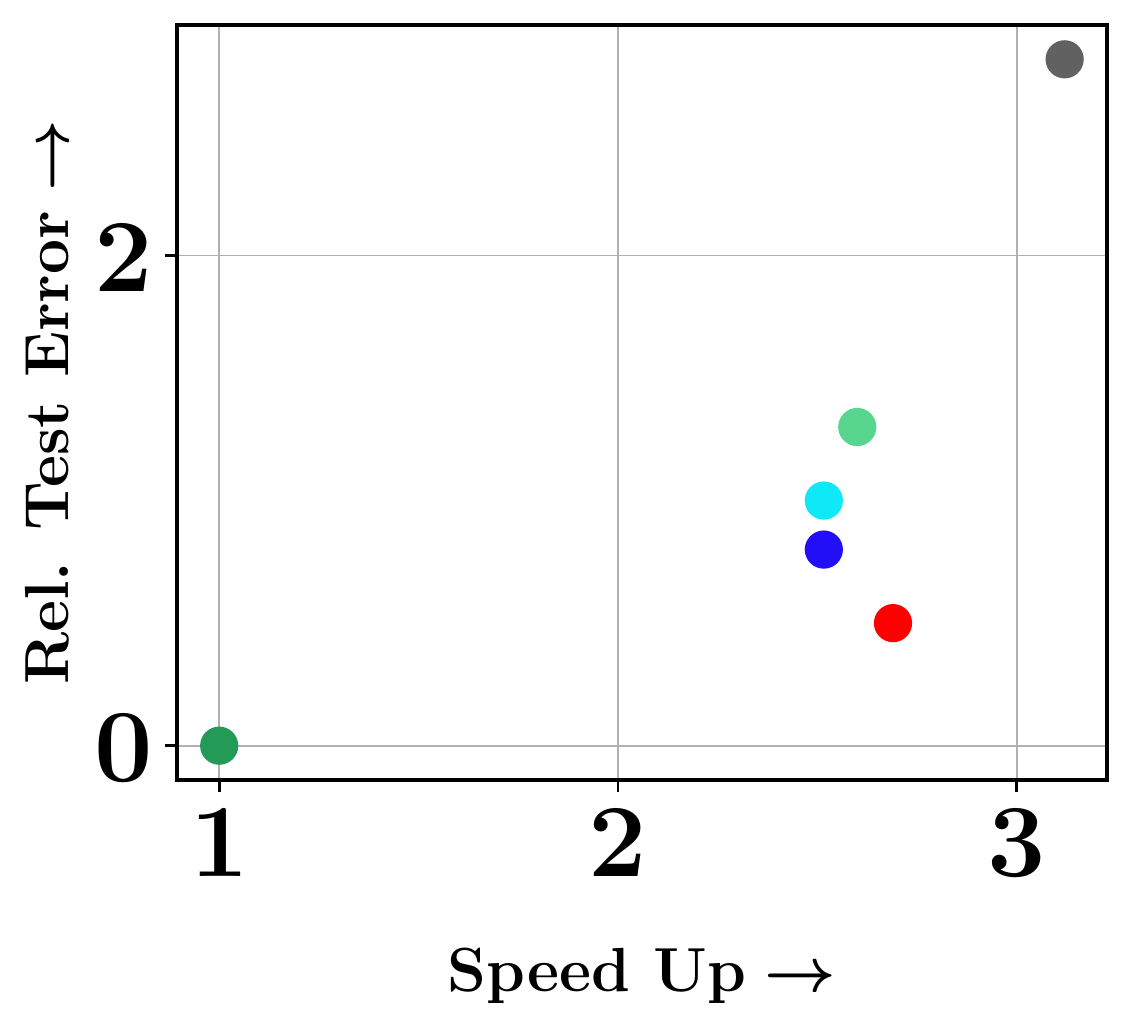}
\caption*{$\underbracket[1pt][1.0mm]{\hspace{3.2cm}}_{\substack{\vspace{-4.0mm}\\
\colorbox{white}{(a) \scriptsize 30\% IMDB-VAT}}}$}
\phantomcaption
\label{fig:imdb-vat}
\end{subfigure}
\begin{subfigure}[b]{0.24\textwidth}
\centering
\includegraphics[width=2.8cm, height=2.3cm]{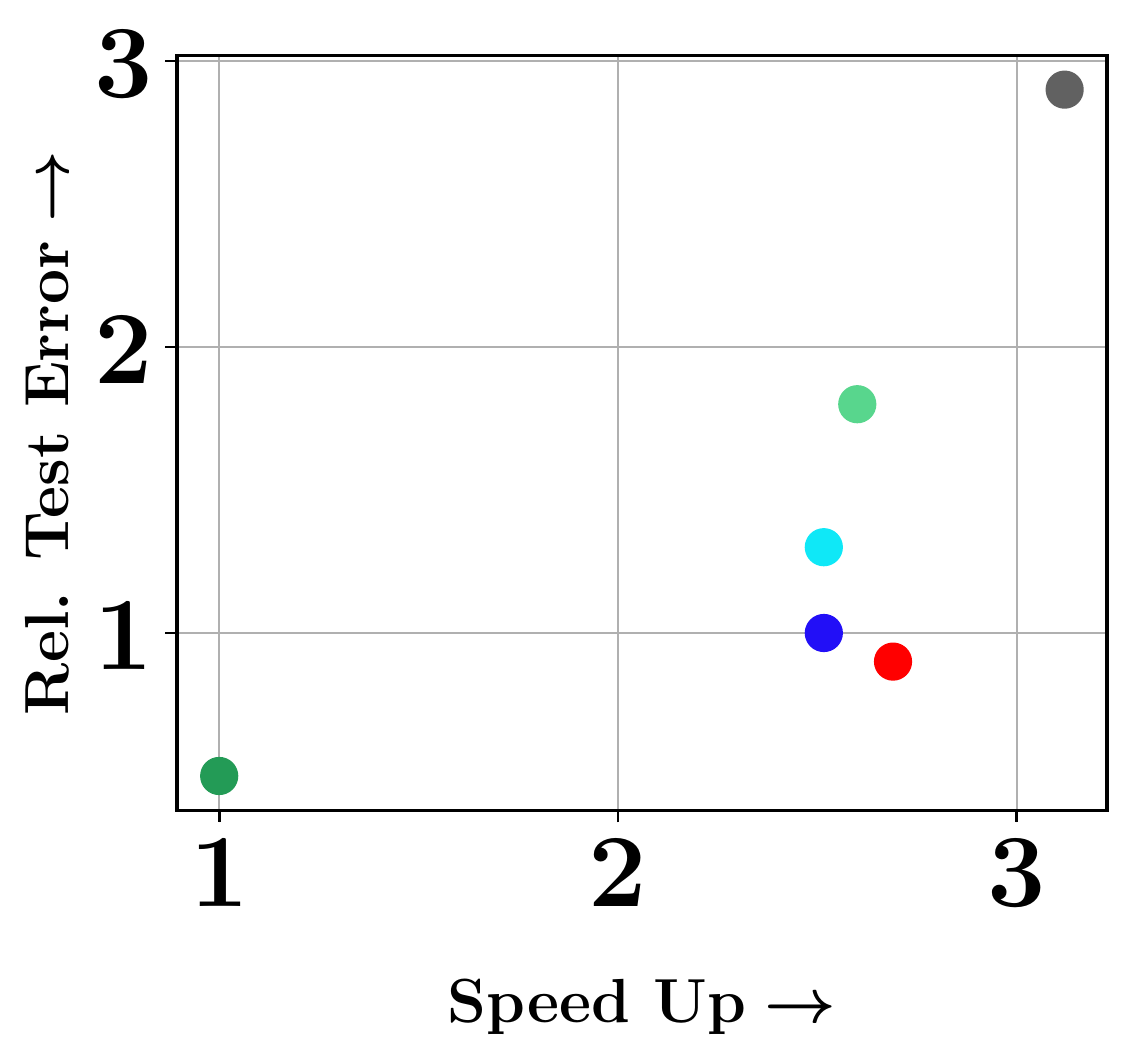}
\caption*{$\underbracket[1pt][1.0mm]{\hspace{3.2cm}}_{\substack{\vspace{-4.0mm}\\
\colorbox{white}{(b) \scriptsize 30\% ELEC-VAT}}}$}
\phantomcaption
\label{fig:elec-vat}
\end{subfigure}
\caption{\footnotesize{Figure 6: A comparison of \model\ with baselines (\textsc{Random, Craig, GradMatch}, \textsc{Full}, and \textsc{FullEarlyStop} in the traditional SSL setting for text datasets. SpeedUp vs Accuracy Degradation, both compared to SSL for (a) VAT on 30\% IMDB, (b) VAT on 30\% ELEC.}
\vspace{-3ex}} 
\label{fig:efficient_text}
\end{wrapfigure}

\textbf{Traditional SSL Results: }  The results comparing the accuracy-efficiency tradeoff between the different subset selection approaches are shown in Figure 4. We compare the performance for different subset sizes of the unlabeled data: 10\%, 20\%, and 30\% and three representative SSL algorithms VAT, Mean-Teacher, and FixMatch. For warm-start, we set kappa value to $\kappa=0.5$ (i.e., training for 50\% epochs on the entire unlabeled set and 50\% using coresets). Our experiments use a $R$ value of 20 (i.e., coreset selection every 20 epochs). Sub-figures(\ref{fig:cifar-vat}, \ref{fig:svhn-vat}, \ref{fig:cifar-mt}, \ref{fig:svhn-mt}, \ref{fig:cifar-fm}) shows the plots of relative error vs speedup, both w.r.t full training (i.e., original SSL algorithm). Sub-figure~\ref{fig:fm-co2} shows the plot of relative error vs CO2 emissions efficiency, both w.r.t full training. CO2 emissions were estimated based on the total compute time using the \href{https://mlco2.github.io/impact#compute}{Machine Learning Impact calculator} presented in \cite{lacoste2019quantifying}. From the results, it is evident that \model\ achieved the best speedup vs. accuracy tradeoff and is environmentally friendly based on CO2 emissions compared to other baselines (including \textsc{Craig} and \textsc{FullEarlyStop}). In particular, \model\ achieves speedup gains of 2.7x and 4.4x with a performance loss of 0.7\% and 0.3\% using VAT on CIFAR10 and SVHN datasets. Further, \model\ achieves speedup gains of 2.9x, 3.2x with a performance loss of 0.02\% and 0.5\% using Mean-Teacher on CIFAR10 and SVHN datasets. Additionally, \model\ achieves a speedup of 3.8x with a performance loss of 0.7\% using FixMatch on the CIFAR10 dataset. Sub-figures(\ref{fig:imdb-vat}, \ref{fig:elec-vat}) shows the plots of relative error vs speedup both w.r.t full training (i.e., original SSL algorithm) on IMDB and ELEC datasets for 30\% subset size. In particular, \model\ achieves speedup gains of 2.68x and 2.5x with a performance loss of 0.5\% and 0.6\% for 30\% subset of IMDB and ELEC datasets. Figure~\ref{fig:bcoreset} shows the results comparing the ~\textsc{BayesianCoreset} method~\cite{pmlr-v80-campbell18a}, adapted to the SSL setting with \model{} using the VAT algorithm for 20\% and 30\% CIFAR10 subsets. The results show that \model{} achieves better performance than the SSL extension of the ~\textsc{BayesianCoreset} selection method in terms of model performance and speedup. One possible explanation for it is that the ~\textsc{BayesianCoreset} approach was not developed for efficient learning but instead was developed to capture coresets that try to represent the log-likelihood of the entire dataset that MCMC methods can further use. We would also like to point out that we used the original code implementation of ~\textsc{BayesianCoreset} that is not meant for GPU usage in our experiments. Hence the speedups of the ~\textsc{BayesianCoreset} approach can be further improved with efficient code implementation. Subfigure~\ref{fig:mt-convg} shows that \textsc{Retrieve} achieves faster convergence compared to all other methods on CIFAR10 for 30\% subset with Mean-Teacher. Subfigure~\ref{fig:vat-convg} shows the extended convergence of \textsc{Retrieve} on CIFAR10 for 20\% and 30\% subsets using VAT, where the \model\ is allowed to train for larger epochs to achieve comparable accuracy with Full training at the cost of losing some efficiency. Note that the points marked by \textbf{*} in subfigure~\ref{fig:vat-convg} denote the actual training endpoint, i.e. the usual number of epochs/iterations used to obtain points in subfigures~(\ref{fig:cifar-vat}, \ref{fig:svhn-vat}, \ref{fig:cifar-mt}, \ref{fig:svhn-mt}, \ref{fig:cifar-fm}). We observe that \model\ matches the performance of VAT while being close to $2\times$ faster in running times (and correspondingly energy efficiency). We repeat this experiment with MT in the Appendix~\ref{app:traditional_ssl}. Also, more detailed results with additional convergence plots and tradeoff curves are in Appendix~\ref{app:traditional_ssl}.

\begin{figure}[t]
\centering
\includegraphics[width = 12cm, height=0.5cm]{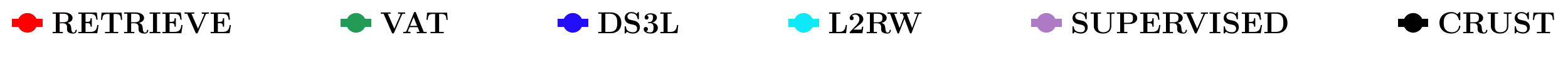}
\captionsetup{labelformat=empty}
\hspace{-0.6cm}
\begin{subfigure}[b]{0.24\textwidth}
\centering
\includegraphics[width=3.2cm, height=2.5cm]{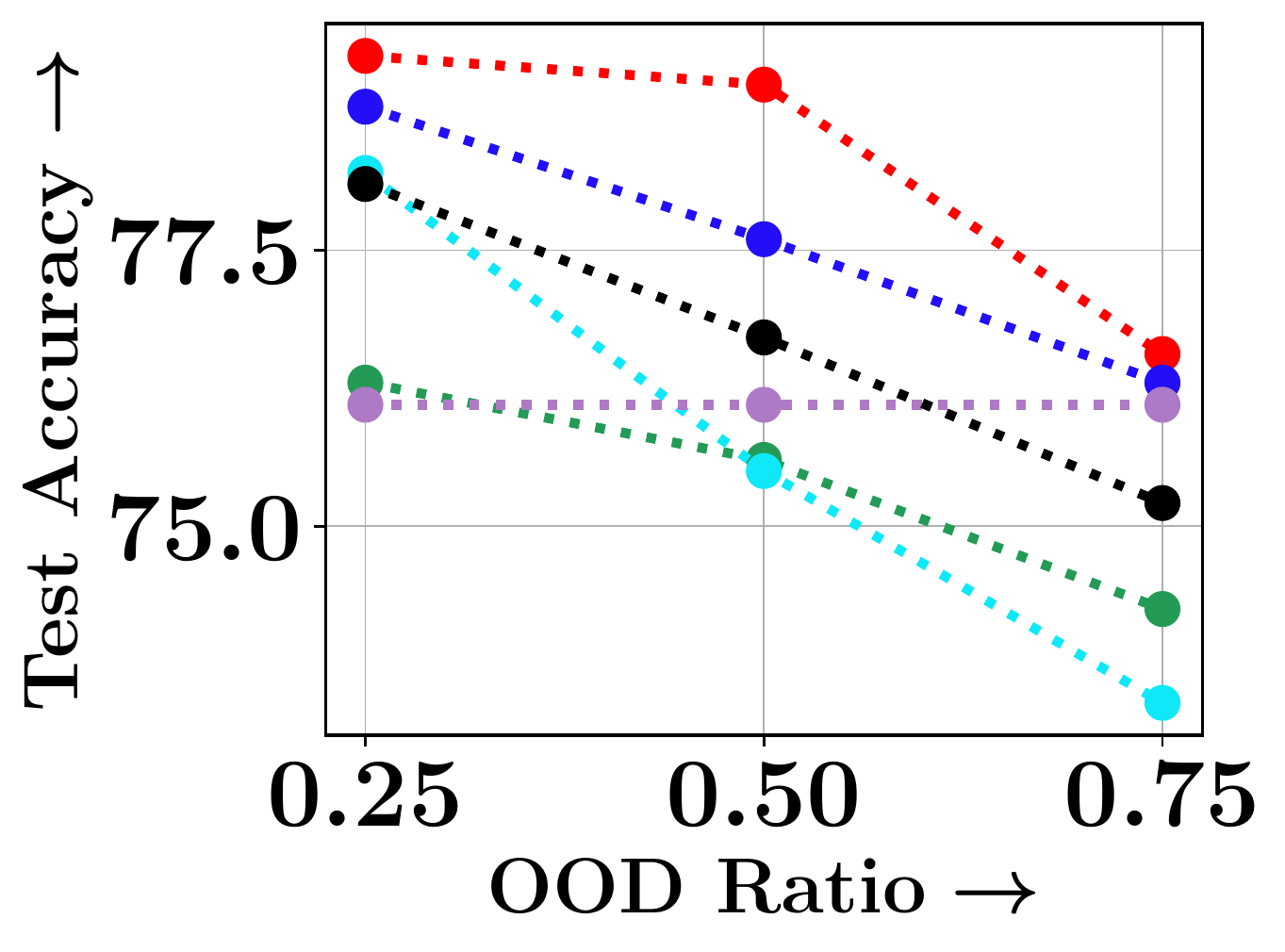}
\caption*{$\underbracket[1pt][1.0mm]{\hspace{3.2cm}}_{\substack{\vspace{-4.0mm}\\
\colorbox{white}{(a) \scriptsize VAT CIFAR10 OOD }}}$}
\phantomcaption
\label{fig:cifarood}
\end{subfigure}
\begin{subfigure}[b]{0.24\textwidth}
\centering
\includegraphics[width=3.2cm, height=2.5cm]{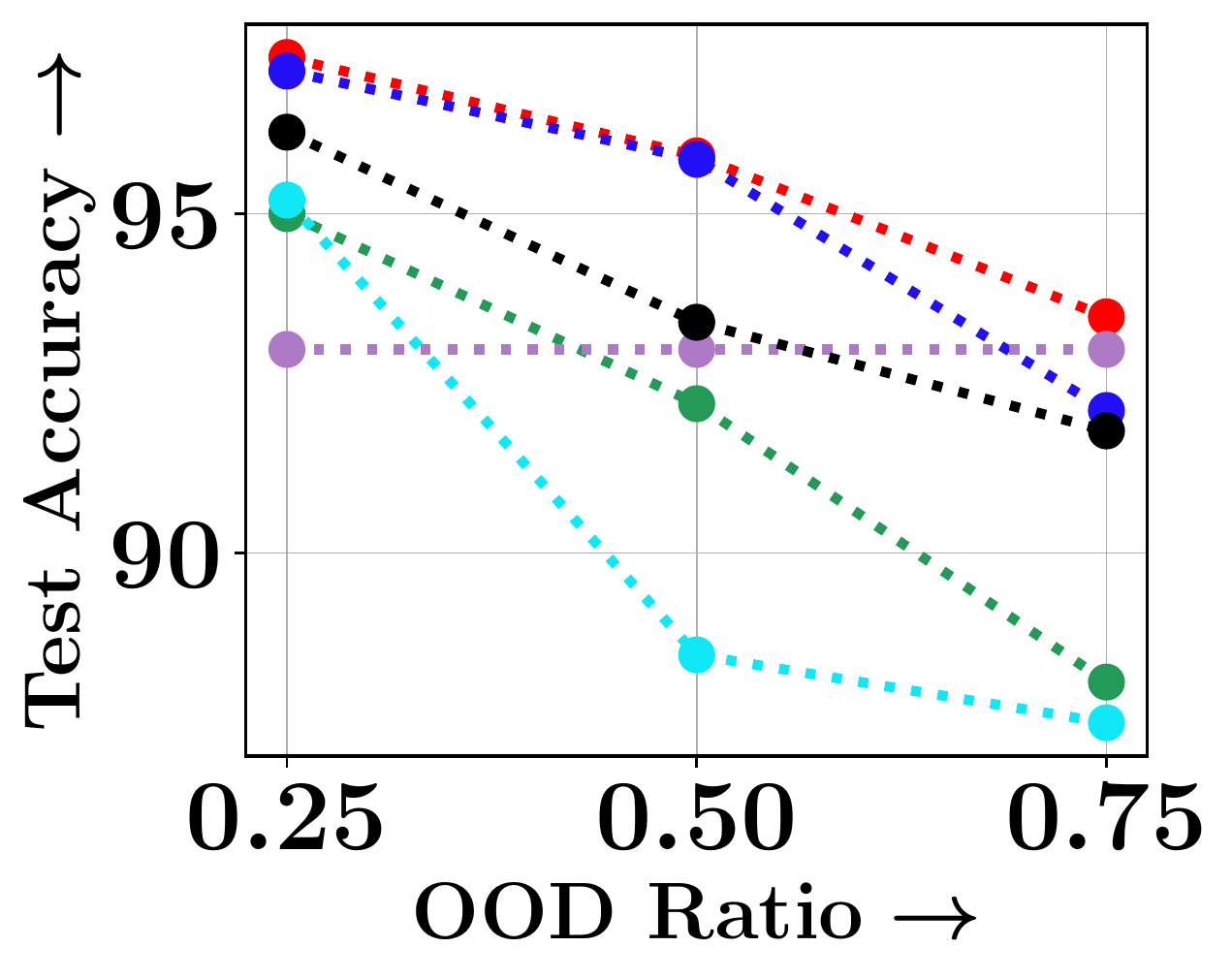}
\caption*{$\underbracket[1pt][1.0mm]{\hspace{3.2cm}}_{\substack{\vspace{-4.0mm}\\
\colorbox{white}{(b) \scriptsize VAT MNIST OOD}}}$}
\phantomcaption
\label{fig:mnistood}
\end{subfigure}
\begin{subfigure}[b]{0.24\textwidth}
\centering
\includegraphics[width=3.2cm, height=2.5cm]{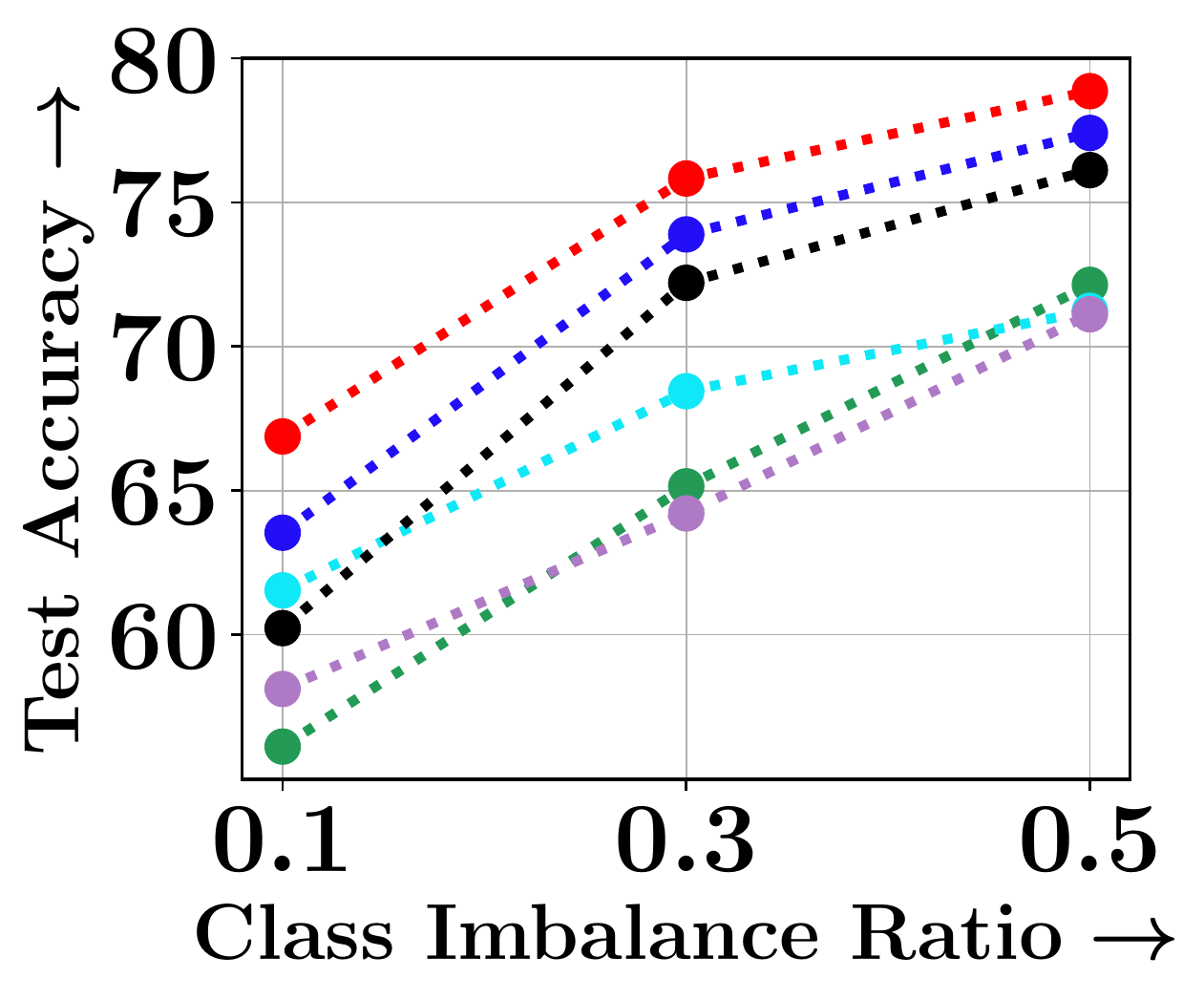}
\caption*{$\underbracket[1pt][1.0mm]{\hspace{3.2cm}}_{\substack{\vspace{-4.0mm}\\
\colorbox{white}{(c) \scriptsize VAT CIFAR10 Imb}}}$}
\phantomcaption
\label{fig:cifarimb}
\end{subfigure}
\begin{subfigure}[b]{0.24\textwidth}
\centering
\includegraphics[width=3.8cm, height=2.5cm]{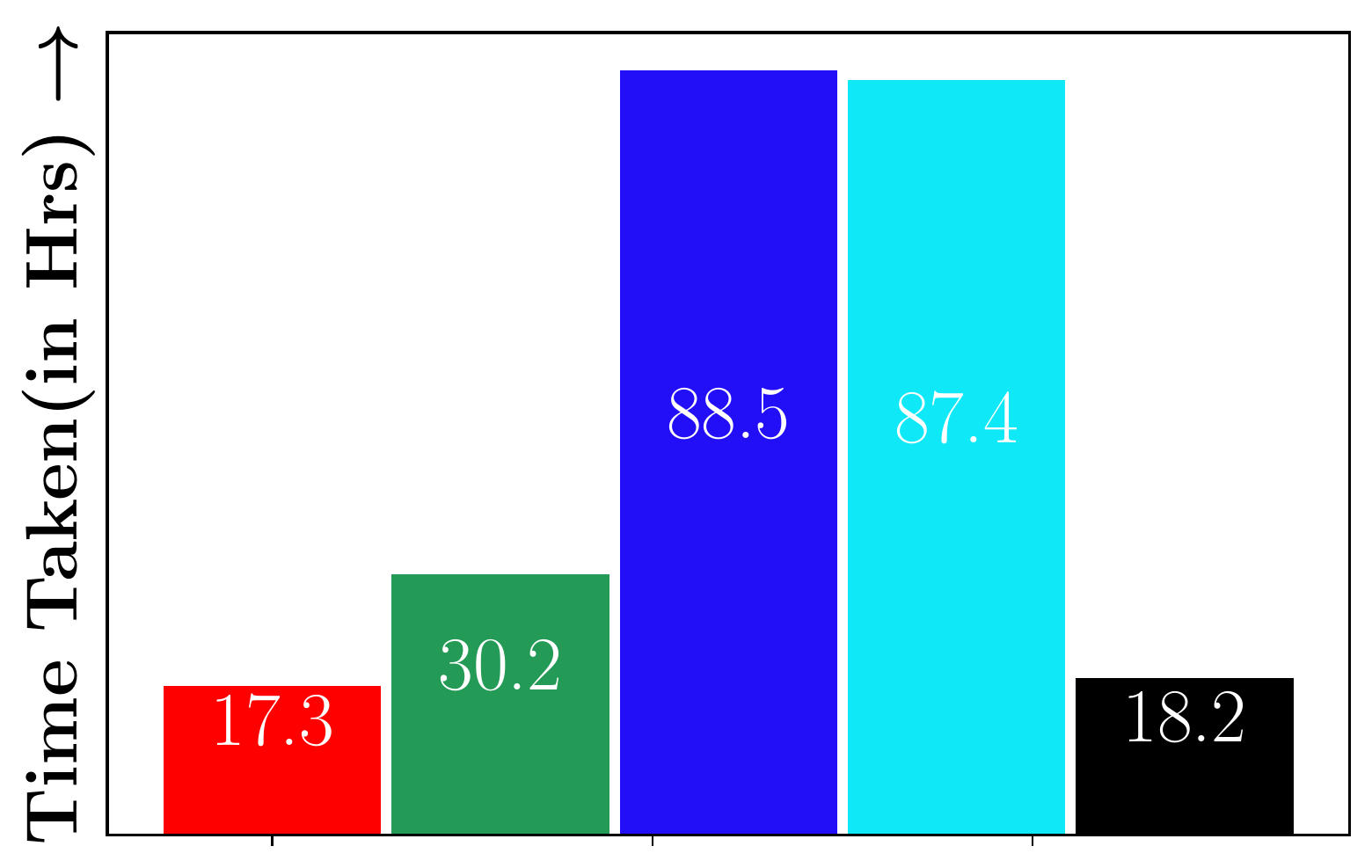}
\caption*{$\underbracket[1pt][1.0mm]{\hspace{3.8cm}}_{\substack{\vspace{-4.0mm}\\
\colorbox{white}{(d) \scriptsize VAT CIFAR10 Imb Timings}}}$}
\phantomcaption
\label{fig:cifarimb-timings}
\end{subfigure}
\caption{\footnotesize{Figure 7: Subfigures (a) to (c) compare the performance of \model\ with baselines (including DS3L and L2RW) for different OOD ratios (a,b) and imbalance ratios (c). We see that \model\ outperforms both baselines in most of the cases. Furthermore, \model\ achieves this while being close to $2\times$ faster than VAT (the SSL algorithm) and $5\times$ faster than the robust SSL algorithms (DS3L and L2RW).
\vspace{-4ex}}}
\label{fig:robust_ssl}
\end{figure}

\looseness -1
\textbf{Robust SSL Results: } We test the performance of \model{} on CIFAR10 and MNIST datasets with OOD in the unlabeled set and CIFAR10 dataset with the class imbalance in both labeled and unlabeled sets. sub-figures{~\ref{fig:cifarood}, \ref{fig:mnistood}} shows the accuracy plots of \model{} for different OOD ratios of 25\%, 50\% and 75\%. The results show that \model{} with VAT outperforms all other baselines, including DS3L~\cite{pmlr-v119-guo20i}, a state-of-the-art robust SSL baseline in the OOD scenario. Next, sub-figure~\ref{fig:cifarimb} shows the accuracy plots of \model{} for different class imbalance ratios of 10\%, 30\% and 50\% on CIFAR-10 dataset. The results show that \model{} with VAT outperforms all other baselines, including DS3L~\cite{pmlr-v119-guo20i} (also run with VAT) in the class imbalance scenario as well. In particular, \model{} outperforms other baselines by at least 1.5\% on the CIFAR-10 with imbalance. Sub-figure~\ref{fig:cifarimb-timings} shows the time taken by different algorithms on the CIFAR10 dataset with a 50\% class imbalance ratio. The results show that \textsc{CRUST} did not perform well in terms of accuracy and speedups achieved compared to \model{}. Except for MixUP, \textsc{CRUST} is similar to \textsc{CRAIG}, which did not perform well compared to \model{} in a traditional SSL setting. Furthermore, the performance gain due to MixUP for coreset selection in the SSL setting is minimal. The minimal gain can be attributed to the fact that the hypothesized labels used for MixUP in the earlier stages of training are noisy. Furthermore, as stated earlier, \textsc{CRUST} was developed to tackle noisy labels in a supervised learning setting and is not developed to deal with OOD or Class Imbalance in general. The results show that \model{} is more efficient compared to the other baselines. In particular, \model{} is 5x times faster compared to DS3L method. Other detailed results (tradeoff curves and convergence curves) are in Appendix~\ref{app:robust_ssl}.
\looseness-1

\section{Conclusion and Broader Impacts}
We introduce \model{}, a discrete-continuous bi-level optimization based coreset selection method for efficient and robust semi-supervised learning. We show connections with weak-submodularity, which enables the coreset selection in \model\ to be solved using a scalable stochastic greedy algorithm. Empirically, we show that \model\ is very effective for SSL. In particular, it achieves $3\times$ speedup on a range of SSL approaches like VAT, MT, and FixMatch with around $0.7\%$ accuracy loss and a $2\times$ speedup with no accuracy loss. In the case of robust SSL with imbalance and OOD data, \model\ outperforms existing SOTA methods while being $5\times$ faster. We believe \model\ has a significant positive societal impact by making SSL algorithms (specifically robust SSL) significantly faster and more energy-efficient, thereby reducing the CO2 emissions incurred during training.

\section{Acknowledgments and Disclosure of Funding}
We would like to thank NeurIPS area chairs and anonymous reviewers for their efforts in reviewing this paper and their constructive comments! RI and KK were funded by the National Science Foundation(NSF) under Grant Number 2106937, a startup grant from UT Dallas, and a Google and Adobe research award. FC and XZ were funded by National Science Foundation(NSF) under Grant Numbers 1815696, 1750911, and 2107449. Any opinions, findings, and conclusions or recommendations expressed in this material are those of the author(s) and do not necessarily reflect the views of the National Science Foundation, Google or Adobe.

\bibliography{main}

\appendix
\newpage
\begin{center}
    \Huge{Supplementary Material}
\end{center}
\addcontentsline{toc}{section}{Appendix} 
\part{Appendix} 
\parttoc 
\newpage
\section{Code and Licenses}\label{app:code}
\subsection{Code}
The code of \model{} for VAT, MT is available at the following link: \url{https://github.com/decile-team/cords}. The code of \model{} for FixMatch is available at the following link: \url{https://github.com/krishnatejakk/EfficientFixMatch}. We will transfer the FixMatch code to the CORDS repository to have a unified repository shortly.

\subsection{Licenses}
We release both the code repositories of \model{} with MIT license, and it is available for everybody to use freely.
For MT and VAT, we built upon the open-source Pytorch implementation\footnote{\url{https://github.com/perrying/pytorch-consistency-regularization}} which is an MIT licensed repo. For the FixMatch method, we implemented it based on an open-source Pytorch implementation\footnote{\url{https://github.com/kekmodel/FixMatch-pytorch}}.
For DS3L~\cite{pmlr-v119-guo20i}, we implemented it based on the released code \footnote{\url{https://github.com/guolz-ml/DS3L}} which has an unknown license. Nevertheless, the authors of the DS3L~\cite{pmlr-v119-guo20i} made the code available for everyone to use.
For L2RW~\cite{ren2018learning}, we used the open-source Pytorch implementation\footnote{\url{https://github.com/danieltan07/learning-to-reweight-examples}} which has an unknown license and adapted it to the SSL settings. Nevertheless, the owner of the repository made the code available for everyone to use.

As far as the datasets are considered, we use CIFAR10~\cite{Krizhevsky09learningmultiple}, SVHN~\cite{netzer2011reading} and MNIST~\cite{lecun-mnisthandwrittendigit-2010} datasets. CIFAR10 dataset is released with an MIT license. MNIST dataset is released with an Creative Commons Attribution-Share Alike 3.0 license. SVHN dataset is released with a CC0:Public Domain license. All the datasets used in this work are publicly available. Furthermore, the datasets used do not contain any personally identifiable information.

\section{Proof of Theorem~\ref{thm:submod-proof}} \label{app:submod}
\label{app-conv-res-proof}
We begin by first stating and then proving Theorem~\ref{thm:submod-proof}.
\begin{theorem-nono}
Optimization problem (\eqref{retrieve-meta-approx}) is NP hard, even if $l_s$ is a convex loss function. If the labeled set loss function $l_s$ is cross-entropy loss, then the optimization problem give in the \eqref{retrieve-meta-approx} can be converted into an instance of cardinality constrained weakly submodular maximization.
\end{theorem-nono}

We use the proof techniques similar to the ones used in Theorem-1 of \textsc{Glister}~\cite{killamsetty2021glister}. \textsc{Glister} proved the weak-submodularity only for the case when both the loss functions in the bi-level optimization problem are cross-entropy losses. In our work, we prove the weak submodularity with the SSL objective as an inner level loss. Furthermore, we prove the weak-submodularity when the unlabeled set loss is either cross-entropy loss or squared loss functions.
\subsection{Proof Sketch}
We introduce the notations used in the proof of the theorem in subsection~\ref{sec:thm_not}. In our proof, we prove that the optimization problem given in \eqref{retrieve-meta-approx} is an $\alpha$-submodular function. We give the definitions of $\alpha$-submodularity, and the approximation guarantees achieved by greedy algorithms in subsection~\ref{sec:thm_alpha_submod}. We state the lemmas of the $\alpha$-submodularity satisfied by the \model{} framework for different cases in subsection~\ref{sec:thm_ret_alpha_submod}. Finally, we give the proof of $\alpha$-submodularity of the \model{} when the unlabeled set loss is cross-entropy loss or squared loss in the subsection~\ref{sec:thm_proof}.
\subsection{Notation}\label{sec:thm_not}
Let $(x_i, y_i)$ be the $i^{th}$ data-point in the labeled set $\mathcal{D}$ where $i \in [1, n]$, and $(x_i^t)$ be the $i^{th}$ data-point in the unlabeled set $\mathcal{U}$ where $i \in [1, m]$. Consider a classification task with $C$ classes. Let the classifier model be characterized by the model parameters $\theta$. As shown in the \eqref{retrieve-setf}, the set function of \model{} is given by $f(\theta_t, \Scal) = -L_S(\Dcal, \theta_t - \alpha_t \nabla_{\theta}L_S(\Dcal, \theta_t) - \alpha_t \lambda_t \underset{j \in \Scal}{\sum} \mb_{jt} \nabla_{\theta}l_u(x_j, \theta_t))$ where $L_S$ is the cross-entropy loss.

The coreset selection optimization problem of \model{} can be written as follows:

\begin{align} 
    \mathcal{S}_{t+1} &= \underset{{\Scal \subseteq {\Ucal}, |\Scal| = k}}{\operatorname{argmax\hspace{0.7mm}}} f(\theta_t, \Scal)
\end{align}

Denote the gain of a set function as  $G_{\theta, v}(s) = f(\theta, v \cup s) - f(\theta, v)$.
In this proof, we prove that the above optimization problem is approximately submodular~\cite{das2011submodular}. 

\subsection{$\alpha$-submodularity} \label{sec:thm_alpha_submod}

Here, we discuss some prior works on submodularity. The definition of $\alpha$-submodularity is given below:

\textbf{Definition: }A function is called $\alpha$-submodular~\cite{DBLP:journals/corr/abs-1811-07863}, if the gain of adding an element $e$ to set $X$ is $1-\alpha$ times greater than or equals to the gain of adding an element $e$ to set $Y$ where $X \subseteq Y$.
i.e.,
\begin{equation}
    \underset{X,Y | X \subseteq Y}{\forall}G_{\theta, X}(e) \geq (1-\alpha) G_{\theta, Y}(e) 
\end{equation}

This definition is different from the notion of $\gamma$-weakly submodular functions~\cite{das2011submodular}. However, as stated in the Proposition 4 of~\cite{DBLP:journals/corr/abs-1811-07863}, $\alpha$-submodular functions and $\gamma$-weakly submodular functions are closely related, where the function that is $\alpha$-submodular is also $\gamma$-weakly submodular with the submodularity ratio $\gamma \geq 1- \alpha$. 

This further implies the following approximation guarantee given below:
\begin{lemma}\cite{das2011submodular,DBLP:journals/corr/abs-1811-07863} Given a $\alpha$-approximate monotone submodular function $f$, the greedy algorithm achieves a $1 - e^{-(1-\alpha)}$ approximation factor for the problem of maximizing $f$ subject to cardinality constraints. 
\end{lemma}
Next, we show that $f(\theta_t, \Scal)$ is a $\alpha$-approximate submodular function. To this extent, we assume that the norm of data points in the labeled and unlabeled sets are bounded such that $\Vert x_i\Vert \leq R$. Note that this is common assumption made in most convergence analysis results. When $l_s$ is cross entropy loss function, we prove that the set function $f(\theta_t, \Scal)$ of \model{} is $\alpha$-approximate submodular function where $\alpha = \frac{2R^2}{2R^2 + 1}$. 

\subsection{$\alpha$-submodularity of \model{}} \label{sec:thm_ret_alpha_submod}
We now show the $\alpha$ submodularity of the set function for different cases of the unlabeled loss $l_u$.
\begin{lemma}
If the labeled set loss function $l_s$ is the cross entropy loss and the unlabeled set loss function $l_u$ is the cross entropy loss, then the optimization problem given in equation~\eqref{retrieve-meta-approx} is an instance of cardinality constrained $\alpha$-approximate submodular maximization, where $\alpha = \frac{2R^2}{2R^2 + 1}$ such that $R$ is the maximum l-2 norm of the data instances in both labeled and the unlabeled sets. 
\end{lemma}

\begin{lemma}
If the labeled set loss function $l_s$ is the cross entropy loss and the unlabeled set loss function $l_u$ is the squared loss, then the optimization problem given in equation~\eqref{retrieve-meta-approx} is an instance of cardinality constrained $\alpha$-approximate submodular maximization, where $\alpha = \frac{R^2}{R^2 + 1}$ such that $R$ is the maximum l-2 norm of the data instances in both labeled and the unlabeled sets. 
\end{lemma}


\subsection{Proof} \label{sec:thm_proof}
Assuming that we start at $\theta_0$ and for the ease of notation we use $f(\Scal)$ instead of $f(\theta_0, \Scal)$. When $l_s$ is cross entropy loss, the optimization problem given in \eqref{retrieve-meta-approx} can be written as follows:
\begin{equation}
\begin{aligned}
    \Scal_{1} &= \underset{\Scal \subseteq \Ucal:|\Scal| \leq k}{\argmax}f(\Scal)\\
    &=\underset{\Scal \subseteq \Ucal:|\Scal| \leq k}{\argmax}-L_S(\Dcal, \theta_0 - \alpha_0 \nabla_{\theta}L_S(\Dcal, \theta_0) - \alpha_0 \lambda_0 \underset{j \in \Scal}{\sum} \mb_{j0} \nabla_{\theta}l_u(x_j, \theta_0))\text{\hspace{1.7cm}}
\end{aligned}
\end{equation}
Substituting $l_s$ with the cross-entropy loss function, then the set function $f(\Scal)$ can be written as follows:
\begin{align}
    f(\Scal) = \sum_{i=1}^{n}\log\left(\frac{\exp((\theta_{0}^{y_i} - \alpha_0 \nabla_{\theta}L_S(\Dcal, \theta_0^{y_i}) - \alpha_0 \lambda_0 \underset{j \in \Scal}{\sum} \mb_{j0} \nabla_{\theta}l_u(x_j, \theta_0^{y_i}))^T x_i)}{\underset{c\in [1, C]}{\sum}\exp((\theta_{0}^{c} - \alpha_0 \nabla_{\theta}L_S(\Dcal, \theta_0^{y_i}) - \alpha_0 \lambda_0 \underset{j \in \Scal}{\sum} \mb_{j0} \nabla_{\theta}l_u(x_j, \theta_0^{c}))^T x_{i})}\right)
\end{align}

Rewriting the above equation, we achieve:
\begin{equation}
\begin{aligned}
    f(\Scal) =  \sum_{i=1}^{n}\Bigg(\log\bigg(\exp((\theta_{0}^{y_i} - \alpha_0 \nabla_{\theta}L_S(\Dcal, \theta_0^{y_i}) - \alpha_0 \lambda_0 \underset{j \in \Scal}{\sum} \mb_{j0} \nabla_{\theta}l_u(x_j, \theta_0^{y_i}))^T x_i)\bigg) \\- \log\bigg(\underset{c\in [1, C]}{\sum}\exp((\theta_{0}^{c} - \alpha_0 \nabla_{\theta}L_S(\Dcal, \theta_0^{y_i}) - \alpha_0 \lambda_0 \underset{j \in \Scal}{\sum} \mb_{j0} \nabla_{\theta}l_u(x_j, \theta_0^{c}))^T x_{i})\bigg)\Bigg)
\end{aligned}
\end{equation}

\begin{align}
    f(\Scal) &=  \sum_{i=1}^{n}\Bigg((\theta_{0}^{y_i} - \alpha_0 \nabla_{\theta}L_S(\Dcal, \theta_0^{y_i}) - \alpha_0 \lambda_0 \underset{j \in \Scal}{\sum} \mb_{j0} \nabla_{\theta}l_u(x_j, \theta_0^{y_i}))^T x_i \\
    &- \log\bigg(\underset{c\in [1, C]}{\sum}\exp((\theta_{0}^{c} - \alpha_0 \nabla_{\theta}L_S(\Dcal, \theta_0^{y_i}) - \alpha_0 \lambda_0 \underset{j \in \Scal}{\sum} \mb_{j0} \nabla_{\theta}l_u(x_j, \theta_0^{c}))^T x_{i})\bigg)\Bigg)
\end{align}
Since, the term $(\theta_{0}^{y_i} - \alpha_0 \nabla_{\theta}L_S(\Dcal, \theta_0^{y_i}))^T x_{i}$ does not depend on the subset $S$, we can remove it from our optimization problem,
\begin{align}
    f(\Scal) &= \sum_{i=1}^{n} \Bigg(\sum_{j \in \Scal} - \alpha_0 \lambda_0 \mb_{j0} \nabla_{\theta}{l_u(x_j, \theta_0^{y_i})}^T x_{i} \\
    &- \log\bigg(\underset{c\in [1, C]}{\sum}\exp((\theta_{0}^{c} - \alpha_0 \nabla_{\theta}L_S(\Dcal, \theta_0^{y_i}) - \alpha_0 \lambda_0 \underset{j \in \Scal}{\sum} \mb_{j0} \nabla_{\theta}l_u(x_j, \theta_0^{c}))^T x_{i})\bigg)\Bigg)
\end{align}
Assume $g_{ijc} = \mb_{j0}\nabla_{\theta}l_u(x_j, \theta_0^{c}))^T x_{i}$,
\begin{align}
    f(\Scal) &= \sum_{i=1}^{n} \Bigg(\sum_{j \in \Scal} - \alpha_0 \lambda_0 g_{ijy_{i}} 
    - \log\bigg(\underset{c\in [1, C]}{\sum}\exp((\theta_{0}^{c} - \alpha_0 \nabla_{\theta}L_S(\Dcal, \theta_0^{y_i}))^T x_{i} - \alpha_0 \lambda_0 \underset{j \in \Scal}{\sum} g_{ijc})\bigg)\Bigg)
\end{align}

\begin{align}
    f(\Scal) &= \sum_{i=1}^{n} \Bigg(\sum_{j \in \Scal} - \alpha_0 \lambda_0 g_{ijy_{i}} 
    - \log\bigg(\underset{c\in [1, C]}{\sum}\exp((\theta_{0}^{c} - \alpha_0 \nabla_{\theta}L_S(\Dcal, \theta_0^{y_i}))^T x_{i})\exp(-\alpha_0 \lambda_0 \underset{j \in \Scal}{\sum} g_{ijc})\bigg)\Bigg)
\end{align}

Let $h_{ic} = \exp((\theta_{0}^{c} - \alpha_0 \nabla_{\theta}L_S(\Dcal, \theta_0^{y_i}))^T x_{i})$ where $h_{ic} \geq 0$ as $h_{ic}$ is an exponential function.

\begin{align}
    f(\Scal) &= \sum_{i=1}^{n} \Bigg(\sum_{j \in \Scal} - \alpha_0 \lambda_0 g_{ijy_{i}} 
    - \log\bigg(\underset{c\in [1, C]}{\sum}h_{ic}\exp(-\alpha_0 \lambda_0 \underset{j \in \Scal}{\sum} g_{ijc})\bigg)\Bigg)
\end{align}

As $g_{ijc}$ is not always greater than zero, we can make some transformations to convert the problem into a monotone submodular function. First, we transform $g_{ijc}$ to $\hat{g}_{ijc}$ such that $g_{ijc} = \hat{g}_{ijc} + g_{m} - 1$ where $g_{m} = \underset{i,j,c}{\min}g_{ijc}$. This transformation ensures that $\hat{g}_{ijc} \geq 1$. Denote $g_{nm} = \underset{i,j,c}{\min}(-g_{ijc})$, and then we define a transformation of $g_{ijc}$ to $g_{ijc}^{''}$ such that $-g_{ijc} = g^{''}_{ijc} + g_{nm}$. Note that both $\hat{g}_{ijc}$ and $g^{''}_{ijc}$ are greater than or equal to zero after the transformations. 

\begin{align}
    f(\Scal) &= \sum_{i=1}^{n} \Bigg(\sum_{j \in \Scal} \alpha_0 \lambda_0 (g^{''}_{ijy_{i}} + g_{nm}) 
    - \log\bigg(\underset{c\in [1, C]}{\sum}h_{ic}\exp(-\alpha_0 \lambda_0 \underset{j \in \Scal}{\sum}(\hat{g}_{ijc} + g_{m} - 1))\bigg)\Bigg)\\
    &= \alpha_{0} \lambda_{0}kng_{nm} + \sum_{i=1}^{n}\Bigg(\sum_{j \in \Scal} \alpha_0 \lambda_0 g^{''}_{ijy_{i}} - \log\bigg(\underset{c\in [1, C]}{\sum}h_{ic}\exp(-\alpha_0 \lambda_0 \underset{j \in \Scal}{\sum}\hat{g}_{ijc}) \exp(k(g_{m} - 1))\bigg)
\end{align}

where $k$ is the size of the subset.

Denote $H_{ic} = h_{ic} \exp(k(g_{m} - 1))$. Further as $\alpha_{0} \lambda_{0}nk(g_{nm})$ is a constant, we can remove it from the optimization problem and we can define the new optimization set function $\hat{f}(\Scal)$ as shown below:
\begin{align}
    \hat{f}(\Scal) =  \sum_{i=1}^{n} \sum_{j \in \Scal} \alpha_{0} \lambda_{0} (g^{''}_{ijy_{i}})  - \sum_{i=1}^{n} \log\bigg(\underset{c\in [1, C]}{\sum}H_{ic}\exp(-\alpha_0 \lambda_0 \underset{j \in \Scal}{\sum}\hat{g}_{ijc})\bigg)
\end{align}

In the above equation, denote the first part as $f_1(\Scal) = \sum_{i=1}^{n} \sum_{j \in S} \alpha_{0} \lambda_{0} (g^{''}_{ijy_{i}})$ which is a monotone modular function in $\Scal$. Similarly, denote the second part $f_2(\Scal) = - \sum_{i=1}^{n} \log\left(\underset{c \in [1, C]}{\sum}H_{ic} \exp(- \alpha_0 \lambda_0 \sum_{j \in \Scal}(\hat{g}_{ijc}))\right)$ is a monotone function but is not submodular.

Hence, we prove that the function $f_2(\Scal)$ is an $\alpha$-submodular function in the following proof section. 
Furthermore, since the first part is positive modular, it is easy to see that if $f_2(X)$ is $\alpha$ submodular (with $\alpha \geq 0$), then the function $f_1(X) + f_2(X)$ will also be an $\alpha$-submodular function. 

Note that, a function $h(X)$ is $\alpha$-submodular if $h(j | X) \geq (1- \alpha) h(j | Y)$ for all subsets $X \subseteq Y$. Assuming that $f_2$ is $\alpha$-submodular, then the following holds:
$$f_2(j | X) \geq (1- \alpha) f_2(j | Y)$$ for all subsets $X \subseteq Y$. 

As $f_1$ is positive modular, we have the following:
$$f_1(j | X) = f_1(j | Y) \geq (1- \alpha) f_1(j | Y)$$. 

This implies that the function $f_1(\Scal) + f_2(\Scal)$ is an $\alpha$-submodular function which further implies that it is also an $\gamma$-weakly submodular function.

\textbf{$\alpha$-submodularity proof of function $f_2$:} 

The gain of adding an element $e$ to the set $X$ is given as follows:
\begin{equation}
\begin{aligned}
    f_2(e|X) &=  - \sum_{i=1}^{n} \log\bigg(\underset{c\in [1, C]}{\sum}H_{ic}\exp(-\alpha_0 \lambda_0 \underset{j \in \Scal \cup e}{\sum}\hat{g}_{ijc})\bigg)\\
    &  + \sum_{i=1}^{n} \log\bigg(\underset{c\in [1, C]}{\sum}H_{ic}\exp(-\alpha_0 \lambda_0 \underset{j \in \Scal}{\sum}\hat{g}_{ijc})\bigg)
\end{aligned}
\end{equation}

\begin{equation}
\begin{aligned}
\label{f2-setf}
    f_2(e|X) &=  - \sum_{i=1}^{n} \log\bigg(\underset{c\in [1, C]}{\sum}H_{ic}\exp(-\alpha_0 \lambda_0 \underset{j \in \Scal}{\sum}\hat{g}_{ijc} - \alpha_0 \lambda_0 \hat{g}_{iec})\bigg)\\
    &  + \sum_{i=1}^{n} \log\bigg(\underset{c\in [1, C]}{\sum}H_{ic}\exp(-\alpha_0 \lambda_0 \underset{j \in \Scal}{\sum}\hat{g}_{ijc})\bigg)
\end{aligned}
\end{equation}

Let, $\hat{g}_{m} = \min_{ijc}\hat{g}_{ijc}$. Then, we can rewrite the above equation as following:
\begin{equation}
\begin{aligned}
    f_2(e|X) &\geq  - \sum_{i=1}^{n} \log\bigg(\underset{c\in [1, C]}{\sum}H_{ic}\exp(-\alpha_0 \lambda_0 \underset{j \in \Scal}{\sum}\hat{g}_{ijc} - \alpha_0 \lambda_0 \hat{g}_{m})\bigg)\\
    &  + \sum_{i=1}^{n} \log\bigg(\underset{c\in [1, C]}{\sum}H_{ic}\exp(-\alpha_0 \lambda_0 \underset{j \in \Scal}{\sum}\hat{g}_{ijc})\bigg)
\end{aligned}
\end{equation}

\begin{align}
    f_2(e|X) &\geq  - \sum_{i=1}^{n} \log\bigg(\underset{c\in [1, C]}{\sum}H_{ic}\exp(-\alpha_0 \lambda_0 \underset{j \in \Scal}{\sum}\hat{g}_{ijc}) \exp(- \alpha_0 \lambda_0 \hat{g}_{m})\bigg)\\
     &+ \sum_{i=1}^{n} \log\bigg(\underset{c\in [1, C]}{\sum}H_{ic}\exp(-\alpha_0 \lambda_0 \underset{j \in \Scal}{\sum}\hat{g}_{ijc})\bigg)
\end{align}

\begin{align}
    f_2(e|X) &\geq  - \sum_{i=1}^{n} \log\bigg(\underset{c\in [1, C]}{\sum}H_{ic}\exp(-\alpha_0 \lambda_0 \underset{j \in \Scal}{\sum}\hat{g}_{ijc}) \bigg) + \sum_{i=1}^{n}\alpha_0 \lambda_0 \hat{g}_{m}\\
     &+ \sum_{i=1}^{n} \log\bigg(\underset{c\in [1, C]}{\sum}H_{ic}\exp(-\alpha_0 \lambda_0 \underset{j \in \Scal}{\sum}\hat{g}_{ijc})\bigg)
\end{align}

\begin{equation}
\begin{aligned}
\label{min-bound}
    f_2(e|X) &\geq  n \alpha_0 \lambda_0 \hat{g}_{m}\\
\end{aligned}
\end{equation}
Let, $\hat{g}_{max} = \max_{ijc}\hat{g}_{ijc}$. Then, we can rewrite the \eqref{f2-setf} as following:
\begin{equation}
\begin{aligned}
    f_2(e|X) &\leq  - \sum_{i=1}^{n} \log\bigg(\underset{c\in [1, C]}{\sum}H_{ic}\exp(-\alpha_0 \lambda_0 \underset{j \in \Scal}{\sum}\hat{g}_{ijc} - \alpha_0 \lambda_0 \hat{g}_{max})\bigg)\\
    &  + \sum_{i=1}^{n} \log\bigg(\underset{c\in [1, C]}{\sum}H_{ic}\exp(-\alpha_0 \lambda_0 \underset{j \in \Scal}{\sum}\hat{g}_{ijc})\bigg)
\end{aligned}
\end{equation}

\begin{align}
    f_2(e|X) &\leq  - \sum_{i=1}^{n} \log\bigg(\underset{c\in [1, C]}{\sum}H_{ic}\exp(-\alpha_0 \lambda_0 \underset{j \in \Scal}{\sum}\hat{g}_{ijc}) \exp(- \alpha_0 \lambda_0 \hat{g}_{max})\bigg)\\
     &+ \sum_{i=1}^{n} \log\bigg(\underset{c\in [1, C]}{\sum}H_{ic}\exp(-\alpha_0 \lambda_0 \underset{j \in \Scal}{\sum}\hat{g}_{ijc})\bigg)
\end{align}

\begin{align}
    f_2(e|X) &\leq  - \sum_{i=1}^{n} \log\bigg(\underset{c\in [1, C]}{\sum}H_{ic}\exp(-\alpha_0 \lambda_0 \underset{j \in \Scal}{\sum}\hat{g}_{ijc}) \bigg) + \sum_{i=1}^{n}\alpha_0 \lambda_0 \hat{g}_{max}\\
     &+ \sum_{i=1}^{n} \log\bigg(\underset{c\in [1, C]}{\sum}H_{ic}\exp(-\alpha_0 \lambda_0 \underset{j \in \Scal}{\sum}\hat{g}_{ijc})\bigg)
\end{align}

\begin{equation}
\begin{aligned}
    \label{max-bound}
    f_2(e|X) &\leq  n \alpha_0 \lambda_0 \hat{g}_{max}\\
\end{aligned}
\end{equation}
Using the minimum bounds and the maximum bounds given in \eqref{min-bound} and \eqref{max-bound} on $f_2(e|X)$, we have:

\begin{equation}
    \underset{X, Y | X \subseteq Y}{\forall} \frac{f_2(e|X)}{f_2(e|Y)} \geq \frac{\hat{g}_{m}}{\hat{g}_{max}}
\end{equation}

Since, $g_{ijc} = \mb_{j0}\nabla_{\theta}l_u(x_j, \theta_0^{c})^T x_{i}$, and $\hat{g}_{ijc} = g_{ijc} - g_{m} + 1$, we have:

\begin{equation}
    \hat{g}_{ijc} = \mb_{j0}\nabla_{\theta}l_u(x_j, \theta_0^{c})^T x_{i} - \underset{i,j,c}{\min \hspace{0.7mm}}\mb_{j0}\nabla_{\theta}l_u(x_j, \theta_0^{c}))^T x_{i} + 1
\end{equation}

For most consistency based SSL algorithms, $l_u$ is either cross-entropy loss or mean-squared loss on the hypothesized label probability prediction.

\textbf{$\alpha$-submodularity when $l_u$ is a cross-entropy loss function: }
Let $[p_{j1}, \cdots, p_{jC}]$  be the class probabilities output by the model for instance $x_j$ in the unlabeled set after the softmax operator and $[q_{j1}, \cdots, q_{jC}]$ be the target probability. If $l_u$ is a cross-entropy loss function, we know that $\nabla_{\theta}l_u(x_j, \theta_0^c) = \sum_{k=1}^{C}q_{jk}(1_{k=c} - p_{jk})x_j$ where $1_{k=c} = 1$ if $k = c$ and $0$ otherwise.

Hence,
\begin{align}
 \underset{i, j, c}{\forall}\nabla_{\theta}\mb_{j0}l_u((x_j^t, \theta_0^c))^T x_i \leq R^2 \text{\hspace{0.2cm}where\hspace{0.2cm}} R \geq \underset{i}{\forall}\left\Vert x_i^t \right\Vert
\end{align}

Similarly,
\begin{align}
 \underset{i, j, c}{\forall}\nabla_{\theta}\mb_{j0}l_u((x_j^t, \theta_0))^T x_i \geq -R^2 \text{\hspace{0.2cm}where\hspace{0.2cm}} R \geq \underset{i}{\forall}\left\Vert x_i^t \right\Vert
\end{align}

Similarly, norm of the labeled set points are bounded from above by $R$. Therefore. $\hat{g}_{m} = 1$ and $\hat{g}_{max} =  2R^2 + 1$. This implies that:

\begin{align}
    \underset{X, Y | X \subseteq Y}{\forall} \frac{f_2(e|X)}{f_2(e|Y)} \geq \frac{1}{2R^2+1}
\end{align}

Since $1- \alpha = \frac{1}{2R^2+1}$, then $\alpha = \frac{2R^2}{2R^2+1}$.

\textbf{$\alpha$-submodularity when $l_u$ is a squared loss function: }
Let $[p_{j1}, \cdots, p_{jC}]$  be the class probabilities output by the model for instance $x_j$ in the unlabeled set after the softmax operator and $[q_{j1}, \cdots, q_{jC}]$ be the target probability. If $l_u$ is a squared loss function, we know that $\nabla_{\theta_0^c}l_u(x_j, \theta_0) = \sum_{k=1}^{C}2(q_{jk} - p_{jk})(1_{k=c} - p_{jk})p_{jk}x_j$ where $1_{k=c} = 1$ if $k = c$ and $0$ otherwise.

\begin{align}
 \underset{i,j,c}{\forall}\nabla_{\theta}\mb_{j0}l_u((x_j^t, \theta_0))^T x_i \leq R^2/2 \text{\hspace{0.2cm}where\hspace{0.2cm}} R \geq \underset{i}{\forall}\left\Vert x_i^t \right\Vert
\end{align}

Similarly,
\begin{align}
 \underset{i,j,c}{\forall}\nabla_{\theta}\mb_{j0}l_u((x_j^t, \theta_0))^T x_i \geq -R^2/2 \text{\hspace{0.2cm}where\hspace{0.2cm}} R \geq \underset{i}{\forall}\left\Vert x_i^t \right\Vert
\end{align}

Similarly, norm of the labeled set points are bounded from above by $R$. Therefore. $\hat{g}_{m} = 1$ and $\hat{g}_{max} =  R^2 + 1$. This implies that:

\begin{align}
    \underset{X, Y | X \subseteq Y}{\forall} \frac{f_2(e|X)}{f_2(e|Y)} \geq \frac{1}{R^2+1}
\end{align}

Since $1- \alpha = \frac{1}{2R^2+1}$, then $\alpha = \frac{R^2}{R^2+1}$.

From both the cases, this implies that $f_2$ is $\alpha$ submodular having $\alpha = \frac{2R^2}{2R^2+1}$ when $l_u$ is cross-entropy loss and $\alpha = \frac{R^2}{R^2+1}$ when $l_u$ is squared loss, which further implies that $f$ is $\alpha$-submodular. This further implies that, any greedy algorithm will achieve a $1 - e^{-(1- \alpha)}$ approximation factor, for the coreset selection step when $l_s$ is a cross-entropy loss and $l_u$ is squared loss or cross-entropy loss.

Finally, the proof of the NP-hardness of the mixed discrete-continuous bi-level optimization problem is shown in Lemma-1 of the work~\cite{killamsetty2021glister}.

\section{Loss formulations for different SSL algorithms}\label{app:loss_form}
\subsection{Notation}
Denote $\Dcal = \{x_i, y_i\}_{i=1}^n$  to be the labeled set with $n$ labeled data points, and $\Ucal = \{x_j\}_{j=1}^m$ to be the unlabeled set with $m$ data points. Let $\theta$ be the classifier model parameters, $l_s$ be the labeled set loss function (such as cross-entropy loss) and $l_u$ be the unlabeled set loss, e.g. consistency-regularization loss, entropy loss, etc.. Denote $L_S(\Dcal, \theta) = \underset{i \in \Dcal}{\sum}l_{s}(\theta, x_i, y_i)$ and $L_U(\Ucal, \theta, \mb) = \underset{j \in \Ucal}{\sum} \mb_j l_u(x_j, \theta)$ where $\mb \in \{0, 1\}^m$ is the binary mask vector for unlabeled set. For notational convenience, we denote $l_{si}(\theta) = l_s(x_i, y_i, \theta)$ and denote $l_{uj}(\theta) = l_u(x_j, \theta)$. We also assume that the functions $l_s$ and $l_u$ involves the scaling constants like $\frac{1}{n}, \frac{1}{m}$
required to consider other loss reductions like mean loss.

\noindent \textbf{Semi-supervised loss: }
Following the above notations, the loss function for many existing SSL algorithms can be written as $L_S(\Dcal, \theta) + \lambda L_U(\Ucal, \theta, \mb)$, where $\lambda$ is the regularization coefficient for the unlabeled set loss. For Mean Teacher~\cite{tarvainen2018mean}, VAT~\cite{miyato2018virtual}, MixMatch~\cite{berthelot2019mixmatch}, the mask vector $\mb$ is made up entirely of ones, whereas for FixMatch~\cite{sohn2020fixmatch}, $\mb$ is confidence-thresholded binary vector, indicating whether to include an unlabeled data instance or not. Usually, $L_S$ is a cross-entropy loss for classification experiments and squared loss for regression experiments. 

Detailed description of SSL loss formulation for different SSL algorithms are given below:

\subsection{Mean-Teacher} Mean Teacher~\cite{tarvainen2018mean} proposed to generate a more stable target output for data points in the unlabeled set using the output of the model using the exponential moving average of model parameter values at previous iterations.
Denote the exponential moving average of model parameters as $\hat{\theta}$. Further, denote $f(\theta, x_i)$ as the softmax of the logits of the datapoint $x_i$ obtained from the model with model parameters $\theta$.

The loss function of Mean-Teacher algorithm is as follows:
$$L_S(\Dcal, \theta) + \lambda \sum_{j=1}^{m} \frac{1}{m} \Vert f(\theta, x_j) - f(\hat{\theta}, x_j) \Vert_2^2$$

where $L_S$ is the mean cross-entropy loss for classification experiments. Further, the mask vector $\mb$ in the case of Mean-Teacher algorithm is made up entirely of ones. And the unlabeled set loss function is a squared loss function.

\subsection{VAT} Virtual adversarial training(VAT)~\cite{miyato2018virtual} tries to find the additional perturbation to the unlabeled data points such that the KL divergence loss is maximized with respect to class predictions distribution after the perturbation.

Let, $f$ be the classifier model characterized by the model parameters $\theta$. Let, $d$ be the additive perturbation to the unlabeled set. Let, $KL(p,q)$ be the KL-Divergence loss between distributions $p$ and $q$. Further, denote $f(\theta, x_i)$ as the softmax of the logits of the datapoint $x_i$ obtained from the model with model parameters $\theta$.

Then, the additional perturbation is given as follows:
    $$\hat{d} = \underset{d}{\argmax} \sum_{j=1}^{m} \frac{1}{m} KL(f(\theta, x_j), f(\theta, x_j+d))$$

The loss function of VAT algorithm is as follows:
$$L_S(\Dcal, \theta) + \lambda \sum_{j=1}^{m} \frac{1}{m} KL(f(\theta, x_j), f(\theta, x_j+\hat{d}))$$

where $L_S$ is the mean cross-entropy loss for classification experiments. Further, the mask vector $\mb$ in the case of VAT algorithm is made up entirely of ones. And the unlabeled set loss function is a KL divergence loss function.

\subsection{MixMatch} MixMatch~\cite{berthelot2019mixmatch} performs augmentations on unlabeled instances and gets a pseudo-label prediction after sharpening the average predictions with different augmentations like shifts, cropping, image flipping, weak and strong augmentation to design the regularization function. Finally, the augmented labeled set and unlabeled sets are concatenated and shuffled to form a new dataset which is used in mix-up~\cite{zhang2018mixup}. 

Let, $f$ be the classifier model characterized by the model parameters $\theta$. Let, ${(\hat{x}_i, \hat{p}_i)}_{i \in [1,n]}$ be the labeled set after mix-up and ${(\hat{x}_j, \hat{p}_j)}_{j \in [1,m]}$ be the unlabeled set after mix-up with predicted labels. Further, denote $f(\theta, x_i)$ as the softmax of the logits of the datapoint $x_i$ obtained from the model with model parameters $\theta$.

The loss function of Mix-Match algorithm is as follows:
$$\frac{1}{n}\sum_{i=1}^{n}CE(f(\theta, \hat{x}_i), \hat{p}_i) + \lambda \sum_{j=1}^{m} \frac{1}{m} \Vert f(\theta, \hat{x}_j) - \hat{p}_j \Vert_2^2$$

where $CE(p, q)$ is the cross-entropy loss between distributions $p$ and $q$. Further, the mask vector $\mb$ in the case of MixMatch algorithm is made up entirely of ones. And the unlabeled set loss function is a l2 squared loss function.

\subsection{FixMatch}FixMatch~\cite{sohn2020fixmatch} uses the cross-entropy loss between class predictions of weak augmented and strong augmented data points as the regularization function. Further, FixMatch uses confidence-based thresholding to consider only unlabeled instances with confident model predictions.

Let, $f$ be the classifier model characterized by the model parameters $\theta$. Let, $\hat{x}_i$ be the weakly augmented version of data point $x_i$ and $\hat{x}^s_i$ be the strong augmented version of data point $x_i$. Further, denote $f(\theta, x_i)$ as the softmax of the logits of the datapoint $x_i$ obtained from the model with model parameters $\theta$.

Then the loss function of FixMatch algorithm is as follows:
$$\frac{1}{n}\sum_{i=1}^{n}CE(f(\theta, \hat{x}_i), y_i) + \lambda \sum_{j=1}^{m} 1_{\max(f(\theta, \hat{x}_j)) \geq \tau}\frac{1}{m} CE(f(\theta, \hat{x}_j), f(\theta, \hat{x}^s_j))$$

where $CE(p, q)$ is the cross-entropy loss between distributions $p$ and $q$. Further, the mask vector $\mb$ in the case of FixMatch algorithm is a binary vector based on confidence thresholding i.e., $\mb_{j} = 1_{\max(f(\theta, \hat{x}_j))}$. And the unlabeled set loss function is also a cross-entropy loss between the weakly and strongly augmented versions.

\section{\textsc{Craig} Algorithm for SSL}\label{app:craig}
In this section, we discuss the formulation of \textsc{Craig}~\cite{mirzasoleiman2020coresets} for coreset selection in the semi-supervised learning scenario. \textsc{Craig} tries to select a coreset of the unlabeled set $\Ucal$ such that the unlabeled loss gradient on the entire unlabeled set is equal to the weighted sum of the unlabeled loss of the individual data points in the selected coreset.

The optimization problem of \textsc{Craig} in the semi-supervised learning scenario can be written as follows:

\begin{equation}
    \begin{aligned}
        \Scal^* = \underset{\Scal \subseteq \Ucal:|\Scal| \leq k, \{\gamma_j\}_{j \in [1, |\Scal|]}:\forall_{j} \gamma_j \geq 0}{\argmin} \left \Vert \underset{i \in \Ucal}{\sum} \mb_i \nabla_{\theta}l_u(x_i, \theta) - \underset{j \in \Scal}{\sum} \mb_j \gamma_j \nabla_{\theta} l_u(x_j, \theta)\right \Vert
    \end{aligned}
\end{equation}

Let the objective function of \textsc{Craig} be denoted as $f(\Scal, \theta) = \left \Vert \underset{i \in \Ucal}{\sum} \mb_i \nabla_{\theta}l_u(x_i, \theta) - \underset{j \in \Scal}{\sum} \mb_j \gamma_j \nabla_{\theta} l_u(x_j, \theta)\right \Vert$.

The above objective function can be upper bounded by converting it into a k-medoids objective function as shown in \textsc{Craig}\cite{mirzasoleiman2020coresets}:

\begin{equation}
    \begin{aligned}
        f(\Scal, \theta) &= \left \Vert \underset{i \in \Ucal}{\sum} \mb_i \nabla_{\theta}l_u(x_i, \theta) - \underset{j \in \Scal}{\sum} \mb_j \gamma_j \nabla_{\theta} l_u(x_j, \theta)\right \Vert \\
                         &\leq \underset{i \in \Ucal}{\sum} \underset{j \in \Scal}{\min} \left \Vert \mb_i \nabla_{\theta}l_u(x_i, \theta) - \mb_j \nabla_{\theta}l_u(x_j, \theta) \right \Vert
    \end{aligned}
\end{equation}

Then the coreset selection problem of \textsc{Craig} in the semi-supervised learning scenario can be written as follows:
\begin{equation}
    \begin{aligned}
        \Scal^* = \underset{\Scal \subseteq \Ucal:|\Scal| \leq k}{\argmin}\underset{i \in \Ucal}{\sum} \underset{j \in \Scal}{\min} \left \Vert \mb_i \nabla_{\theta}l_u(x_i, \theta) - \mb_j \nabla_{\theta}l_u(x_j, \theta) \right \Vert
    \end{aligned}
\end{equation}

Then the weights for each data instance in the selected coreset is calculated as follows:
\begin{equation}
    \begin{aligned}
        \gamma_j = \underset{i \in \Ucal}{\sum} 1_{j = \underset{k \in \Scal}{\argmin}\left \Vert \mb_i \nabla_{\theta}l_u(x_i, \theta) - \mb_k \nabla_{\theta}l_u(x_k, \theta) \right \Vert}
    \end{aligned}
\end{equation}

where $1_x = 1$ if $x = True$ and $1_x = 0$ otherwise.

However, in our experiments, we used a per-batch version of the \textsc{Craig} problem discussed above since it is shown to be more effective in work~\cite{killamsetty2021gradmatch}. In the per-batch version, we assume that the unlabeled set is divided into a set of mini-batches denoted by $B^u = \{b^u_1, b^u_2, \cdots , b^u_{\floor{m/B}}\}$ where $b^u_1 = \{x_i:x_i \in \Ucal\}_{i=1}^{B}$ is a mini-batch of unlabeled set of size $B$. Further, we select $\floor{k/B}$ mini-batches in the per-batch version of \textsc{Craig} instead of $k$ data points.

The per-batch version of \textsc{Craig} can be given as follows:
\begin{equation}
    \begin{aligned}
    \label{craigpb-obj}
        \Scal^* = \underset{\Scal \subseteq B^u:|\Scal| \leq \floor{k/B}}{\argmin}\underset{i \in B^u}{\sum} \underset{j \in \Scal}{\min} \left \Vert \sum_{k \in b^u_{i}}\mb_k \nabla_{\theta}l_u(x_k, \theta) - \sum_{l \in b^u_{j}}\mb_l \nabla_{\theta}l_u(x_l, \theta) \right \Vert
    \end{aligned}
\end{equation}

Then the weights for all the data instances in a selected mini-batch $b^u_j$ is calculated as follows:

\begin{equation}
    \begin{aligned}
    \label{craigpb-wts}
        \gamma_j = \underset{i \in B^u}{\sum} 1_{j = \underset{k \in \Scal}{\argmin}\left \Vert \sum_{p \in b^u_{i}}\mb_p \nabla_{\theta}l_u(x_p, \theta) - \sum_{l \in b^u_{k}}\mb_l \nabla_{\theta}l_u(x_l, \theta) \right \Vert}
    \end{aligned}
\end{equation}

As discussed earlier, in our experiments, we use the per-batch versions of \textsc{Craig} and the optimization problem is given in \eqref{craigpb-obj}. Furthermore, the weights for each data instance are calculated as shown in the \eqref{craigpb-wts}.

\section{\textsc{GradMatch} Algorithm for SSL}\label{app:gradmatch}
In this section, we discuss the formulation of \textsc{GradMatch}~\cite{killamsetty2021gradmatch} for coreset selection in the semi-supervised learning scenario. \textsc{GradMatch} tries to select a coreset of the unlabeled set $\Ucal$ such that the unlabeled loss gradient on the entire unlabeled set is equal to the weighted sum of the unlabeled loss of the individual data points in the selected coreset.

The optimization problem of \textsc{GradMatch} in the semi-supervised learning scenario can be written as follows:

\begin{equation}
    \begin{aligned}
        \Scal^* = \underset{\Scal \subseteq \Ucal:|\Scal| \leq k, \{\gamma_j\}_{j \in [1, |\Scal|]}:\forall_{j} \gamma_j \geq 0}{\argmin} \left \Vert \underset{i \in \Ucal}{\sum} \mb_i \nabla_{\theta}l_u(x_i, \theta) - \underset{j \in \Scal}{\sum} \mb_j \gamma_j \nabla_{\theta} l_u(x_j, \theta)\right \Vert
    \end{aligned}
\end{equation}

Let the objective function of \textsc{GradMatch} be denoted as $f(\Scal, \theta) = \left \Vert \underset{i \in \Ucal}{\sum} \mb_i \nabla_{\theta}l_u(x_i, \theta) - \underset{j \in \Scal}{\sum} \mb_j \gamma_j \nabla_{\theta} l_u(x_j, \theta)\right \Vert$.

The above objective function can be solved using the Orthogonal Matching Pursuit(OMP) algorithm as shown in \textsc{GradMatch}\cite{killamsetty2021gradmatch}.

However, in our experiments, we used a per-batch version of the \textsc{GradMatch} problem discussed above since it is shown to be more effective in work~\cite{killamsetty2021gradmatch}. In the per-batch version, we assume that the unlabeled set is divided into a set of mini-batches denoted by $B^u = \{b^u_1, b^u_2, \cdots , b^u_{\floor{m/B}}\}$ where $b^u_1 = \{x_i:x_i \in \Ucal\}_{i=1}^{B}$ is a mini-batch of unlabeled set of size $B$. Further, we select $\floor{k/B}$ mini-batches in the per-batch version of \textsc{GradMatch} instead of $k$ data points.

The per-batch version of \textsc{GradMatch} can be given as follows:
\begin{equation}
    \begin{aligned}
    \label{gradmatchpb-obj}
        \Scal^* = \underset{\Scal \subseteq B^u:|\Scal| \leq \floor{k/B}}{\argmin} {\min} \left \Vert \underset{i \in B^u} \sum \sum_{k \in b^u_{i}}\mb_k \nabla_{\theta}l_u(x_k, \theta) - \underset{j \in \Scal} \sum \sum_{l \in b^u_{j}}\mb_l \nabla_{\theta}l_u(x_l, \theta) \right \Vert
    \end{aligned}
\end{equation}

Then the weights and the mini-batches are selected using the Orthogonal Matching Pursuit (OMP) algorithm. As discussed earlier, in our experiments, we use the per-batch versions of \textsc{GradMatch} and the optimization problem is given in \eqref{gradmatchpb-obj}.

\section{More Details on Experimental Setup, Datasets, and Baselines}\label{app:experimental}
\subsection{Datasets}

\subsubsection{Traditional SSL scenario}

\begin{table}[!tbh]
  \centering
  \begin{adjustbox}{width=1\textwidth}
      \begin{tabular}{|c|c|c|c|c|c|c|} 
     \hline 
     \textbf{Name} & \textbf{No. of classes} & \textbf{No. samples for} & \textbf{No. samples for} & \textbf{No. samples for} & \textbf{No. of features} & \textbf{License} \\  
     ~ & ~ & \textbf{training} & \textbf{validation} & \textbf{testing} & ~ & ~ \\ [0.5ex] 
     \hline
     CIFAR10 & 10  & 50,000 & - & 10,000 & 32x32x3 & MIT\\ 
     \hline
     SVHN & 10 & 73,257 & - & 26,032 & 32x32x3 & CC0:Public Domain\\
     \hline
     \end{tabular}
 \end{adjustbox}
 \caption{Description of the datasets}
  \label{tab:tssl_datasets}
\end{table}

\begin{table}[!tbh]
  \centering
  \begin{adjustbox}{width=1\textwidth}
      \begin{tabular}{|c|c|c|c|c|c|} 
     \hline 
     \textbf{Name} & \textbf{Labeled set size} & \textbf{Unlabeled set size} & \textbf{Test set size} & \textbf{Labeled set batch size} & \textbf{Unlabeled set batch size} \\  
     \hline
     CIFAR10 & 4000  & 50,000 & 10,000 & 50 & 50\\ 
     \hline
     SVHN & 1000 & 73,257 & 26,032 & 50 & 50\\
     \hline
     \end{tabular}
 \end{adjustbox}
 \caption{Dataset Splits used in the traditional SSL scenario}
  \label{tab:tssl_split}
\end{table}

We used various standard datasets, viz., CIFAR10, SVHN, to demonstrate the effectiveness and stability of \model{} in the traditional SSL scenario. The descriptions of the datasets used along with the licenses are given in the \tabref{tab:tssl_datasets}. Furthermore, the labeled, unlabeled, and test data splits for each dataset considered along with the labeled and the unlabeled set batch sizes are given in \tabref{tab:tssl_split}. Both CIFAR10 and SVHN datasets are publicly available. Furthermore, the datasets used do not contain any personally identifiable information.

\subsubsection{Robust SSL scenario}
We used CIFAR10, MNIST, to demonstrate the effectiveness and stability of \model{} in the robust SSL scenario. The descriptions of the datasets used along with the licenses are given in the \tabref{tab:tssl_datasets}. Both CIFAR10 and MNIST datasets are publicly available. Furthermore, the datasets used do not contain any personally identifiable information.

\begin{table}[!tbh]
  \centering
  \begin{adjustbox}{width=1\textwidth}
      \begin{tabular}{|c|c|c|c|c|c|c|c|} 
     \hline 
     \textbf{Name} & \textbf{No. of classes} & \textbf{No. of classes} & \textbf{No. samples for} & \textbf{No. samples for} & \textbf{No. samples for} & \textbf{No. samples for} & \textbf{No. of features}  \\  
     ~ & \textbf{for ID} & \textbf{for OOD} & \textbf{labeled} & \textbf{unlabeled} & \textbf{validation} & \textbf{testing} & ~  \\ [0.5ex] 
     \hline
     CIFAR10 & 6  & 4 & 2,400 & 20,000 & 5,000 & 10,000 & 32x32x3 \\ 
     \hline
     MNIST & 6 & 4 & 60 & 30,000 & 10,000 & 10,000 & 28x28x1 \\
     \hline
     \end{tabular}
 \end{adjustbox}
 \caption{Description of the datasets for robust SSL OOD scenario}
  \label{tab:ood_datasets}
\end{table}

\begin{table}[!tbh]
  \centering
  \begin{adjustbox}{width=1\textwidth}
      \begin{tabular}{|c|c|c|c|c|c|c|c|} 
     \hline 
     \textbf{Name} & \textbf{Imbalanced classes} & \textbf{balanced classes} & \textbf{No. samples for} & \textbf{No. samples for} & \textbf{No. samples for} & \textbf{No. samples for} & \textbf{No. of features}  \\  
     ~ & ~  & ~  & \textbf{labeled} & \textbf{unlabeled} & \textbf{validation} & \textbf{testing} & ~  \\ [0.5ex] 
     \hline
     CIFAR10 & 1-5  & 6-10 & 2,400 & 20,000 & 5,000 & 10,000 & 32x32x3 \\ 
     \hline
     \end{tabular}
 \end{adjustbox}
 \caption{Description of the datasets for robust SSL imbalanced scenario}
  \label{tab:imb_datasets}
\end{table}

\begin{sidewaystable}[!tbhp]
\centering
\scalebox{0.9}{
\begin{tabular}{c c c|c c c|c c c} \hline \hline
\multicolumn{9}{c}{VAT traditional SSL Results}\\ \hline
\multicolumn{3}{c|}{} & \multicolumn{3}{c|}{Top-1 Test accuracy(\%)} & \multicolumn{3}{c}{Model Training time(in hrs)} \\ 
\multicolumn{1}{c}{} & \multicolumn{1}{c}{} & \multicolumn{1}{c|}{Budget(\%)} &  \multicolumn{1}{c}{10\%} & \multicolumn{1}{c}{20\%} & \multicolumn{1}{c|}{30\%} & \multicolumn{1}{c}{10\%} & \multicolumn{1}{c}{20\%} & \multicolumn{1}{c}{30\%} \\ \hline
\multicolumn{1}{c}{Dataset} & \multicolumn{1}{c}{Model}  &\multicolumn{1}{c|}{Selection Strategy} & \multicolumn{3}{c|}{} & \multicolumn{3}{c}{} \\ \hline
CIFAR10 &Wide-ResNet-28-2 &\textsc{Full} (skyline for test accuracy)  &87.8&  87.8&  87.8 &30.41 &30.41  &30.41 \\ 
 & &\textsc{Random} (skyline for training time) &81.95 &84.98 &85.6  &3.08 &6.69 &9.98\\ \cline{3-9}
 & &\textsc{Craig}  &83.2 &85.3 &86.8  &3.54 &7.19 &{\color{red}10.14} \\ 
 & &\textsc{Retrieve} &{\color{red}84.0} &{\color{red}85.9} &{\color{red}87.02}  &{\color{red}3.50} &{\color{red}7.06} &10.27  \\\hline \hline
 SVHN &Wide-ResNet-28-2 &\textsc{Full} (skyline for test accuracy)  &93.62 &93.62 &93.62 &19.17 &19.17 &19.17\\ 
 & &\textsc{Random} (skyline for training time) &87.86 &90.12 &91.24 &1.98 &3.94 &5.65 \\ \cline{3-9}
 & &\textsc{Craig}  &88.86 &91.25 &91.94 &{\color{red}2.12} &{\color{red}4.08} &{\color{red}5.98} \\
 & &\textsc{Retrieve} &{\color{red}89.3} &{\color{red}93.2} &{\color{red}93.3}  &2.18 &4.2 &6.1 \\\hline \hline
\end{tabular}}
    \caption{Traditional SSL Results for CIFAR10 and SVHN datasets using VAT algorithm}
    \label{tab:vat_tssl_results}
    \centering
    \scalebox{0.9}{
    \begin{tabular}{c c c|c c c|c c c} \hline \hline
\multicolumn{9}{c}{Mean-Teacher traditional SSL Results}\\ \hline
\multicolumn{3}{c|}{} & \multicolumn{3}{c|}{Top-1 Test accuracy(\%)} & \multicolumn{3}{c}{Model Training time(in hrs)} \\ 
\multicolumn{1}{c}{} & \multicolumn{1}{c}{} & \multicolumn{1}{c|}{Budget(\%)} &  \multicolumn{1}{c}{10\%} & \multicolumn{1}{c}{20\%} & \multicolumn{1}{c|}{30\%} & \multicolumn{1}{c}{10\%} & \multicolumn{1}{c}{20\%} & \multicolumn{1}{c}{30\%} \\ \hline
\multicolumn{1}{c}{Dataset} & \multicolumn{1}{c}{Model}  &\multicolumn{1}{c|}{Selection Strategy} & \multicolumn{3}{c|}{} & \multicolumn{3}{c}{} \\ \hline
CIFAR10 &Wide-ResNet-28-2 &\textsc{Full} (skyline for test accuracy)  &86.61 &86.61  &86.61 &30.41 &30.41 &30.41 \\ 
 & &\textsc{Random} (skyline for training time) &83.3 &84.41  &84.5  &3.26 &6.54 &9.82\\ \cline{3-9}
 & &\textsc{Craig}  &83.57 &84.68 &85.48  &{\color{red}3.33} &{\color{red}6.7} &11.08 \\ 
 & &\textsc{Retrieve} &{\color{red}83.66} &{\color{red}85.05} &{\color{red}86.59} &3.58 &6.92 &{\color{red}10.55}  \\\hline \hline
 SVHN &Wide-ResNet-28-2 &\textsc{Full} (skyline for test accuracy)  &94.35 &94.35 &94.35 &13.75 &13.75 &13.75\\ 
 & &\textsc{Random} (skyline for training time) &91.48 &92.79 &93.15 &1.52 &2.66 &4.11 \\ \cline{3-9}
 & &\textsc{Craig}  &{\color{red}92.11} &91.81 &91.98 &1.7 &{\color{red}2.88} &4.49 \\
 & &\textsc{Retrieve} &91.84 &{\color{red}93.13} &{\color{red}93.76} &{\color{red}1.57} &2.94 &{\color{red}4.30} \\\hline \hline
\end{tabular}}
    \caption{FixMatch traditional SSL Results for CIFAR10 and SVHN datasets using Mean-Teacher algorithm}
    \label{tab:mt_tssl_results}
    \centering
    \scalebox{0.9}{
    \begin{tabular}{c c c|c c c|c c c} \hline \hline
\multicolumn{9}{c}{Traditional SSL Results}\\ \hline
\multicolumn{3}{c|}{} & \multicolumn{3}{c|}{Top-1 Test accuracy(\%)} & \multicolumn{3}{c}{Model Training time(in hrs)} \\ 
\multicolumn{1}{c}{} & \multicolumn{1}{c}{} & \multicolumn{1}{c|}{Budget(\%)} &  \multicolumn{1}{c}{10\%} & \multicolumn{1}{c}{20\%} & \multicolumn{1}{c|}{30\%} & \multicolumn{1}{c}{10\%} & \multicolumn{1}{c}{20\%} & \multicolumn{1}{c}{30\%} \\ \hline
\multicolumn{1}{c}{Dataset} & \multicolumn{1}{c}{Model}  &\multicolumn{1}{c|}{Selection Strategy} & \multicolumn{3}{c|}{} & \multicolumn{3}{c}{} \\ \hline
CIFAR10 &Wide-ResNet-28-2 &\textsc{Full} (skyline for test accuracy)  &95.52 &95.52  &95.52  &100  &100  &100 \\ 
 & &\textsc{Random} (skyline for training time) &93.8 &94.31 &94.4 &12.76 &25.6 &39.1\\ \cline{3-9}
 & &\textsc{Craig}  &93.9 &94.52 &94.3 &13.18 &26.4 &40 \\ 
 & &\textsc{Retrieve} &{\color{red}94.6} &{\color{red}94.8} &{\color{red}94.83} &{\color{red}13.14} &{\color{red}26.2} &{\color{red}39.8}  \\\hline \hline
\end{tabular}}
    \caption{Traditional SSL Results for CIFAR10 dataset using FixMatch algorithm}
    \label{tab:fm_tssl_results}
\end{sidewaystable}

\begin{table}[!ht]
    \centering
    \scalebox{0.8}{
    \begin{tabular}{c c c|c c c} \hline \hline
\multicolumn{6}{c}{VAT Standard Deviation Results}\\ \hline
\multicolumn{3}{c|}{} & \multicolumn{3}{c}{Standard deviation of the Model(for 3 runs)} \\ 
\multicolumn{1}{c}{} & \multicolumn{1}{c}{} & \multicolumn{1}{c|}{Budget(\%)} & \multicolumn{1}{c}{10\%} & \multicolumn{1}{c}{20\%} & \multicolumn{1}{c}{30\%} \\ \hline
\multicolumn{1}{c}{Dataset} & \multicolumn{1}{c}{Model}  &\multicolumn{1}{c|}{Selection Strategy} & \multicolumn{3}{c}{} \\ \hline
CIFAR10 &Wide-ResNet-28-2 &\textsc{Full}  &0.124	&0.124	&0.124 \\ 
 & &\textsc{Random}  &0.526 &0.538 &0.512 \\ \cline{3-6}
 & &\textsc{Craig}  &0.368	&0.285	&0.195\\
 & &\textsc{Retrieve}  &0.198 &0.148 &0.105\\
 \hline \hline
SVHN &Wide-ResNet-28-2 &\textsc{Full}  &0.114	&0.114	&0.114 \\ 
 & &\textsc{Random}  &0.372 &0.358 &0.348 \\ \cline{3-6}
 & &\textsc{Craig}  &0.284	&0.241	&0.207\\
 & &\textsc{Retrieve}  &0.187 &0.154 &0.112\\
 \hline \hline
\end{tabular}}
    \caption{Standard deviation results using VAT in traditional SSL scenario for CIFAR10, SVHN datasets for three runs}
    \label{tab:tssl_vat_std}
\end{table}

\begin{table}[!ht]
    \centering
    \scalebox{0.8}{
    \begin{tabular}{c c c|c c c} \hline \hline
\multicolumn{6}{c}{Mean-Teacher Standard Deviation Results}\\ \hline
\multicolumn{3}{c|}{} & \multicolumn{3}{c}{Standard deviation of the Model(for 3 runs)} \\ 
\multicolumn{1}{c}{} & \multicolumn{1}{c}{} & \multicolumn{1}{c|}{Budget(\%)} & \multicolumn{1}{c}{10\%} & \multicolumn{1}{c}{20\%} & \multicolumn{1}{c}{30\%} \\ \hline
\multicolumn{1}{c}{Dataset} & \multicolumn{1}{c}{Model}  &\multicolumn{1}{c|}{Selection Strategy} & \multicolumn{3}{c}{} \\ \hline
CIFAR10 &Wide-ResNet-28-2 &\textsc{Full}  &0.105	&0.105	&0.105 \\ 
 & &\textsc{Random}  &0.578 &0.524 &0.564 \\ \cline{3-6}
 & &\textsc{Craig}  &0.482	&0.386	&0.324\\
 & &\textsc{Retrieve}  &0.196 &0.162 &0.121\\
 \hline \hline
SVHN &Wide-ResNet-28-2 &\textsc{Full}  &0.11	&0.11	&0.11 \\ 
 & &\textsc{Random}  &0.374 &0.329 &0.354 \\ \cline{3-6}
 & &\textsc{Craig}  &0.268	&0.284	&0.245\\
 & &\textsc{Retrieve}  &0.146 &0.094 &0.078\\
 \hline \hline
\end{tabular}}
    \caption{Standard deviation results using Mean-Teacher in traditional SSL scenario for CIFAR10, SVHN datasets for three runs}
    \label{tab:tssl_mt_std}
\end{table}

\begin{table}[!ht]
    \centering
    \scalebox{0.8}{
    \begin{tabular}{c c c|c c c} \hline \hline
\multicolumn{6}{c}{FixMatch Standard Deviation Results}\\ \hline
\multicolumn{3}{c|}{} & \multicolumn{3}{c}{Standard deviation of the Model(for 3 runs)} \\ 
\multicolumn{1}{c}{} & \multicolumn{1}{c}{} & \multicolumn{1}{c|}{Budget(\%)} & \multicolumn{1}{c}{10\%} & \multicolumn{1}{c}{20\%} & \multicolumn{1}{c}{30\%} \\ \hline
\multicolumn{1}{c}{Dataset} & \multicolumn{1}{c}{Model}  &\multicolumn{1}{c|}{Selection Strategy} & \multicolumn{3}{c}{} \\ \hline
CIFAR10 &Wide-ResNet-28-2 &\textsc{Full}  &0.12	&0.12	&0.12 \\ 
 & &\textsc{Random}  &0.523 &0.618 &0.584 \\ \cline{3-6}
 & &\textsc{Craig}  &0.386	&0.342	&0.305\\
 & &\textsc{Retrieve}  &0.174 &0.142 &0.105\\
 \hline \hline
\end{tabular}}
    \caption{Standard deviation results using FixMatch in traditional SSL scenario for CIFAR10, SVHN datasets for three runs}
    \label{tab:tssl_fm_std}
\end{table}

\subsection{Traditional SSL baselines}
In this setting, we run \model{} (and all baselines) with warm-start. We incorporate \model\ with three representative SSL methods, including Mean Teacher (MT)~\cite{tarvainen2018mean}, Virtual Adversarial Training (VAT)~\cite{miyato2018virtual} and FixMatch~\cite{sohn2020fixmatch}. The baselines considered are \textsc{Random} (where we just randomly select a subset of unlabeled data points of the same size as \model), \textsc{Craig}~\cite{mirzasoleiman2020coresets,killamsetty2021gradmatch} and \textsc{Full-EarlyStop}. \textsc{Craig}~\cite{mirzasoleiman2020coresets,killamsetty2021gradmatch} was actually proposed in the supervised learning scenario. We adapt it to SSL by choosing a representative subset of unlabeled points such that the gradients are similar to the unlabeled loss gradients. For more information on the formulation of \textsc{Craig} in the SSL case, see Appendix~\ref{app:craig}. We run the per-batch variant of \textsc{Craig} proposed in~\cite{killamsetty2021gradmatch}, where we select a subset of mini-batches instead of data instances for efficiency and scalability. Again, we emphasize that \textsc{Random} and \textsc{Craig} are run with early stopping for the same duration as \model. In \textsc{Full-EarlyStop} baseline, we train the model on the entire unlabeled set for the time taken by \textsc{Retrieve} and report the test accuracy.

\subsection{Robust SSL baselines}
In this setting, we run \model{} (and all baselines) without warm-start. We incorporate \model\ and other baselines with VAT method. DS3L considers a shallow neural network (also called meta-network) to predict the weights of unlabeled examples and estimate the parameters of the neural network based on a clean labeled set (which could also be the original labeled set) via bi-level optimization. For L2RW method, it directly considers the sample weights are hyperparameter and optimize the hyperparameter via bi-level optimization. 

\subsection{Experimental Setup}
In our experiments, we implement our approaches \model{} for three representative SSL methods, including Mean Teacher
(MT), Virtual Adversarial Training (VAT) and fixmatch. For MT and VAT, we built upon the open-source Pytorch implementation\footnote{\url{https://github.com/perrying/pytorch-consistency-regularization}}. For fixmatch method, we implemented it based on a open-source Pytorch implementation\footnote{\url{https://github.com/kekmodel/FixMatch-pytorch}}.
For DS3L~\cite{pmlr-v119-guo20i}, we implemented it based on the released code \footnote{\url{https://github.com/guolz-ml/DS3L}}.
For L2RW~\cite{ren2018learning}, we used the open-source Pytorch implementation\footnote{\url{https://github.com/danieltan07/learning-to-reweight-examples}}and adapted it to the SSL settings.

We use a WideResNet-28-2~\cite{zagoruyko2017wide} model and a Nesterov's accelerated SGD optimizer with a learning rate of 0.03, weight decay of 5e-4, the momentum of 0.9, and a cosine annealing~\cite{LoshchilovH16a} learning rate scheduler for all the experiments except with MNIST OOD. For the MNIST OOD experiment, we used a two-layer CNN model consisting of two conv2d layers of dimensions 1x16x3 and 16x32x3, two MaxPool2d layers with size=3, stride=2, padding=1, and a RELU activation function. Finally, for MNIST OOD experiments, the optimizer and learning rate schedulers are the same as given above, while the learning rate used is 0.003.

\begin{figure}[!tbhp]
\centering
\begin{subfigure}[b]{0.48 \textwidth}
\centering
\includegraphics[width=4.2cm, height=3.5cm]{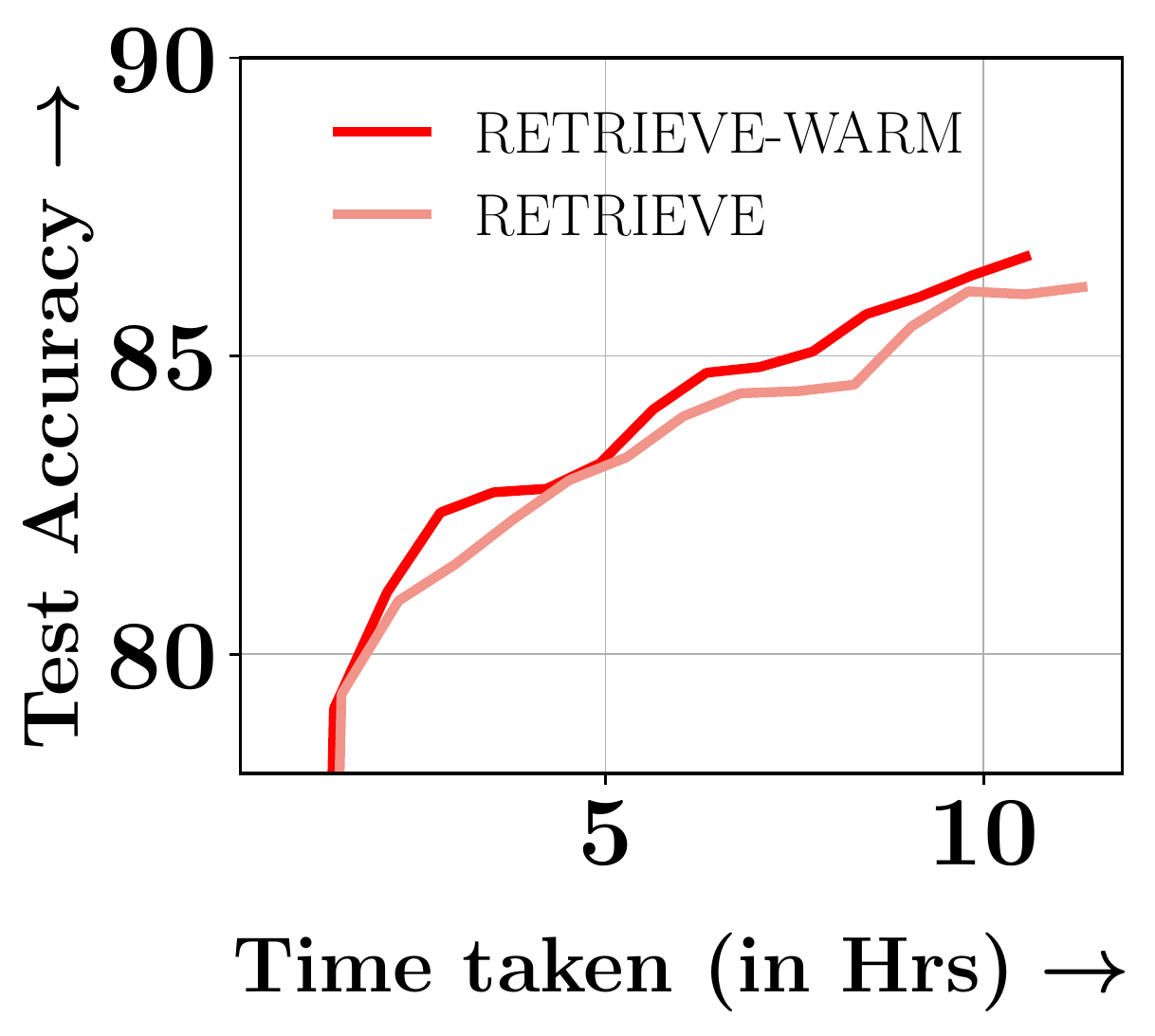}
\caption*{$\underbracket[1pt][1.0mm]{\hspace{5.2cm}}_{\substack{\vspace{-4.0mm}\\
\colorbox{white}{\scriptsize (a) MT \textsc{Retrieve} vs \textsc{Retrieve-Warm}}}}$}
\phantomcaption
\label{fig:mt-retvsretw}
\end{subfigure}\quad
\begin{subfigure}[b]{0.48 \textwidth}
\centering
\includegraphics[width=4.2cm, height=3.5cm]{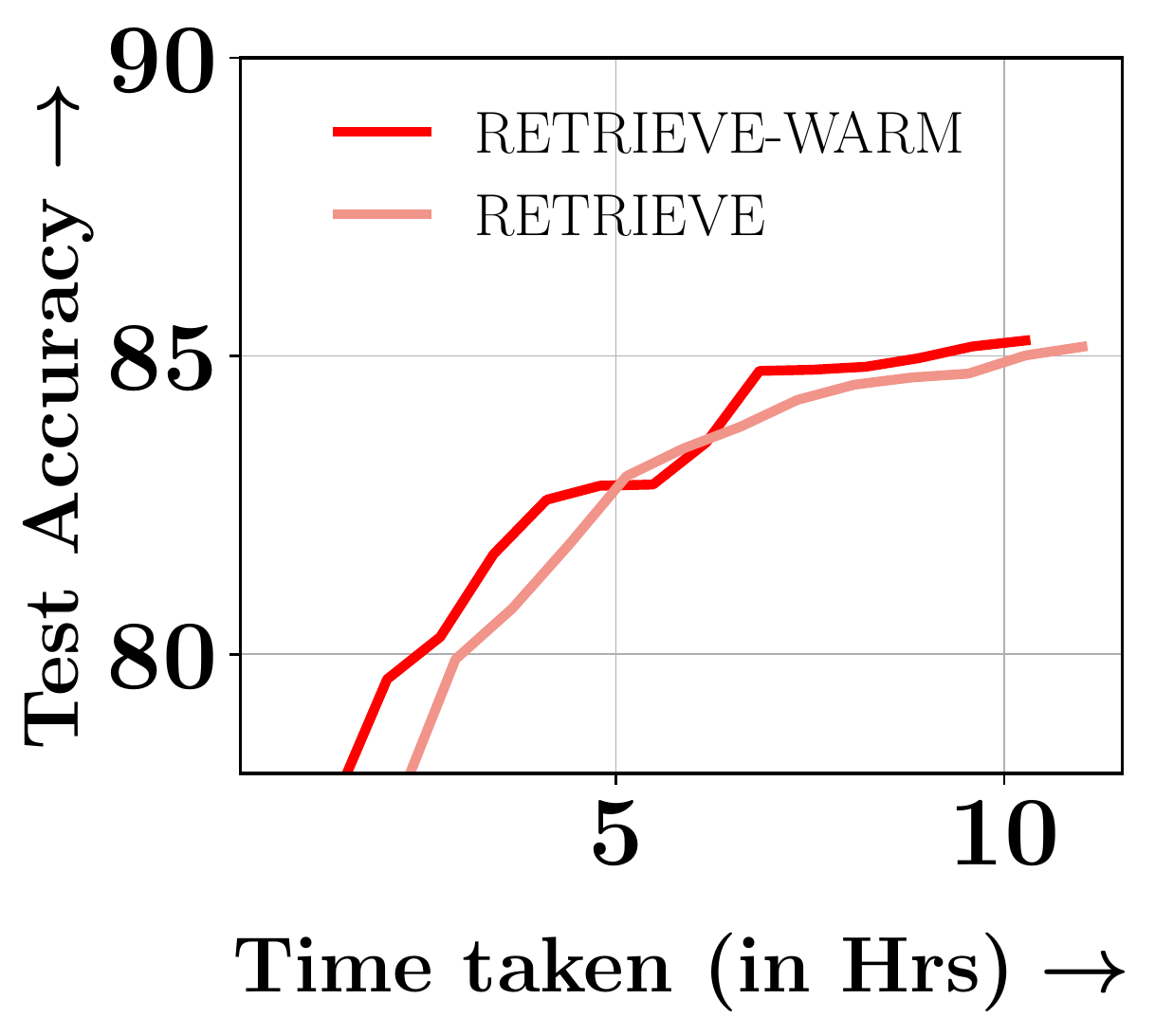} 
\caption*{$\underbracket[1pt][1.0mm]{\hspace{5.2cm}}_{\substack{\vspace{-4.0mm}\\
\colorbox{white}{\scriptsize (b) VAT \textsc{Retrieve} vs \textsc{Retrieve-Warm}}}}$}
\phantomcaption
\label{fig:vat-retvsretw}
\end{subfigure}\quad
\caption{\small{Subfigures (a), (b) show comparison of \model{} vs \textsc{Retrieve-Warm} with MT, VAT on CIFAR-10 dataset with 30\% subset fraction: We show that the \textsc{Retrieve-Warm} is more effective and efficient compared to \model{} in traditional SSL setting.}
\vspace{-3ex}}
\label{fig:retvsretw}
\end{figure}

\begin{figure}[!tbhp]
\centering
\includegraphics[width = 12cm, height=0.5cm]{figs/legend_notbold_eff.pdf}
\captionsetup{labelformat=empty}
\hspace{-0.6cm}
\begin{subfigure}[b]{0.48\textwidth}
\centering
\includegraphics[width=4.2cm, height=3.5cm]{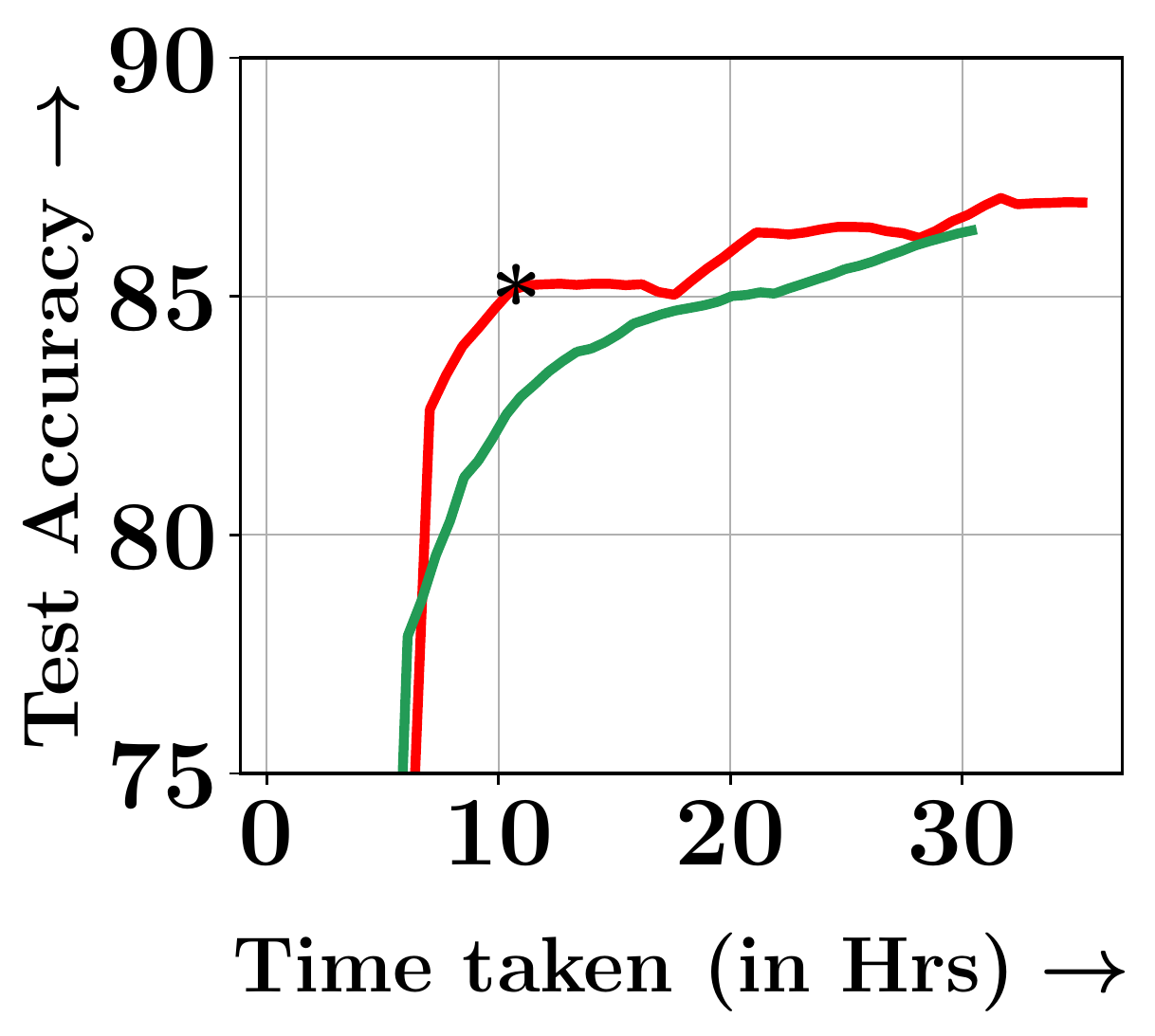}
\caption*{$\underbracket[1pt][1.0mm]{\hspace{5.2cm}}_{\substack{\vspace{-4.0mm}\\
\colorbox{white}{(a) \scriptsize CIFAR10-MT Extended Convergence}}}$}
\phantomcaption
\label{fig:mt_ext_convergence}
\end{subfigure}
\begin{subfigure}[b]{0.48\textwidth}
\centering
\includegraphics[width=4.2cm, height=3.5cm]{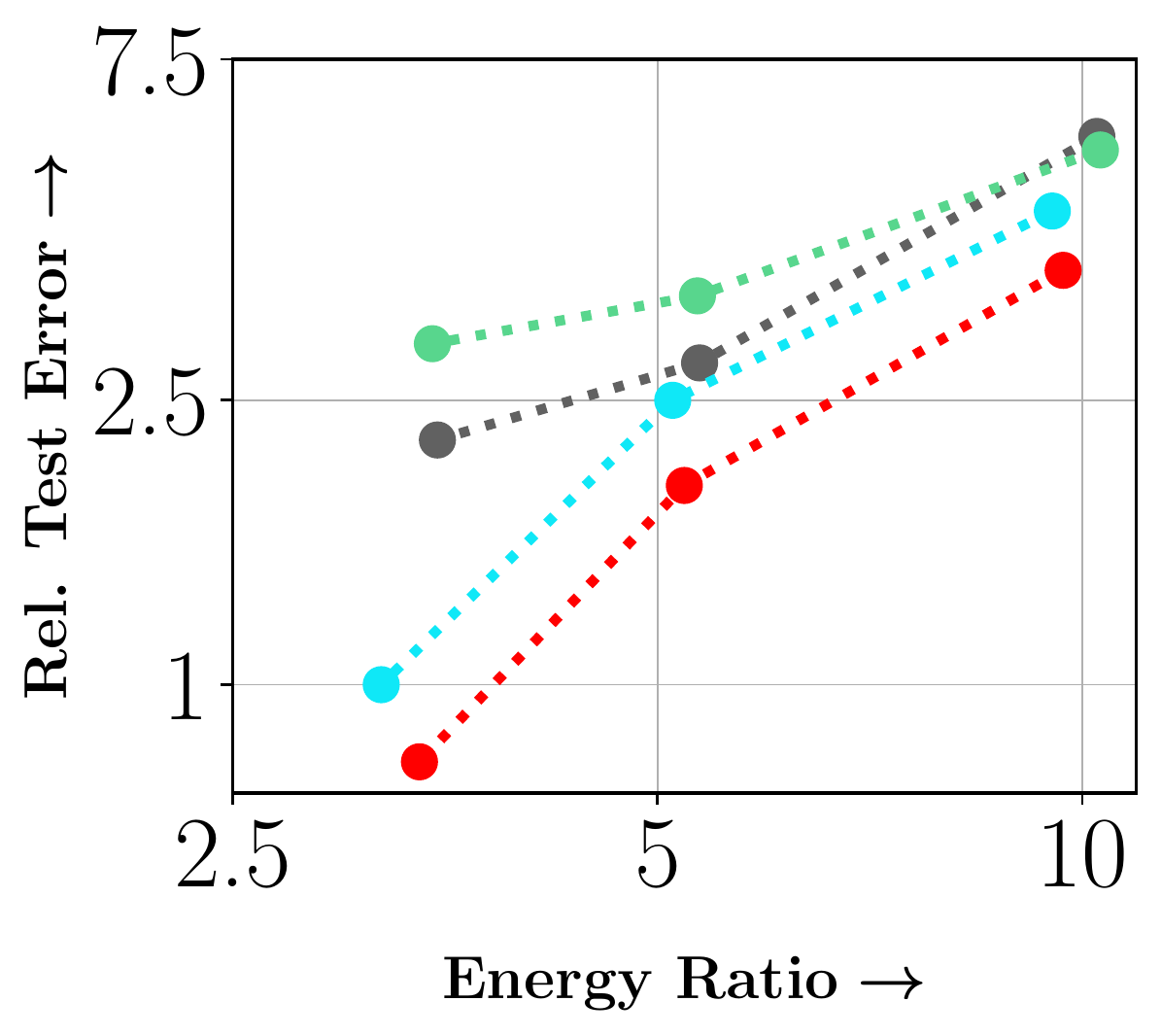}
\caption*{$\underbracket[1pt][1.0mm]{\hspace{5.2cm}}_{\substack{\vspace{-4.0mm}\\
\colorbox{white}{(b) \scriptsize CIFAR10-VAT Energy}}}$}
\phantomcaption
\label{fig:vat_energy}
\end{subfigure}\quad
\caption{Figure 7: \small{Subfigure (a) shows MT extended convergence plot on CIFAR-10 dataset with 30\% subset fraction: We show that the \model{} achieves similar performance to original MT algorithm while being 1.8X faster and better performance than the original MT algorithm while being 1.5X faster. Subfigure (b) shows the energy efficiency plot of VAT algorithm on CIFAR10 dataset with a subset fractions of 10\%, 20\%, 30\%: We show that the \model{} is 3.1X energy efficient compared to the original VAT algorithm with an accuracy degradation of 0.78\%.}}
\label{fig:add_energy_plots}
\end{figure}

\begin{figure}[!tbhp]
\centering
\includegraphics[width = 12cm, height=0.5cm]{figs/legend_notbold_eff.pdf}
\captionsetup{labelformat=empty}
\hspace{-0.6cm}
\begin{subfigure}[b]{0.32\textwidth}
\centering
\includegraphics[width=4.2cm, height=3.5cm]{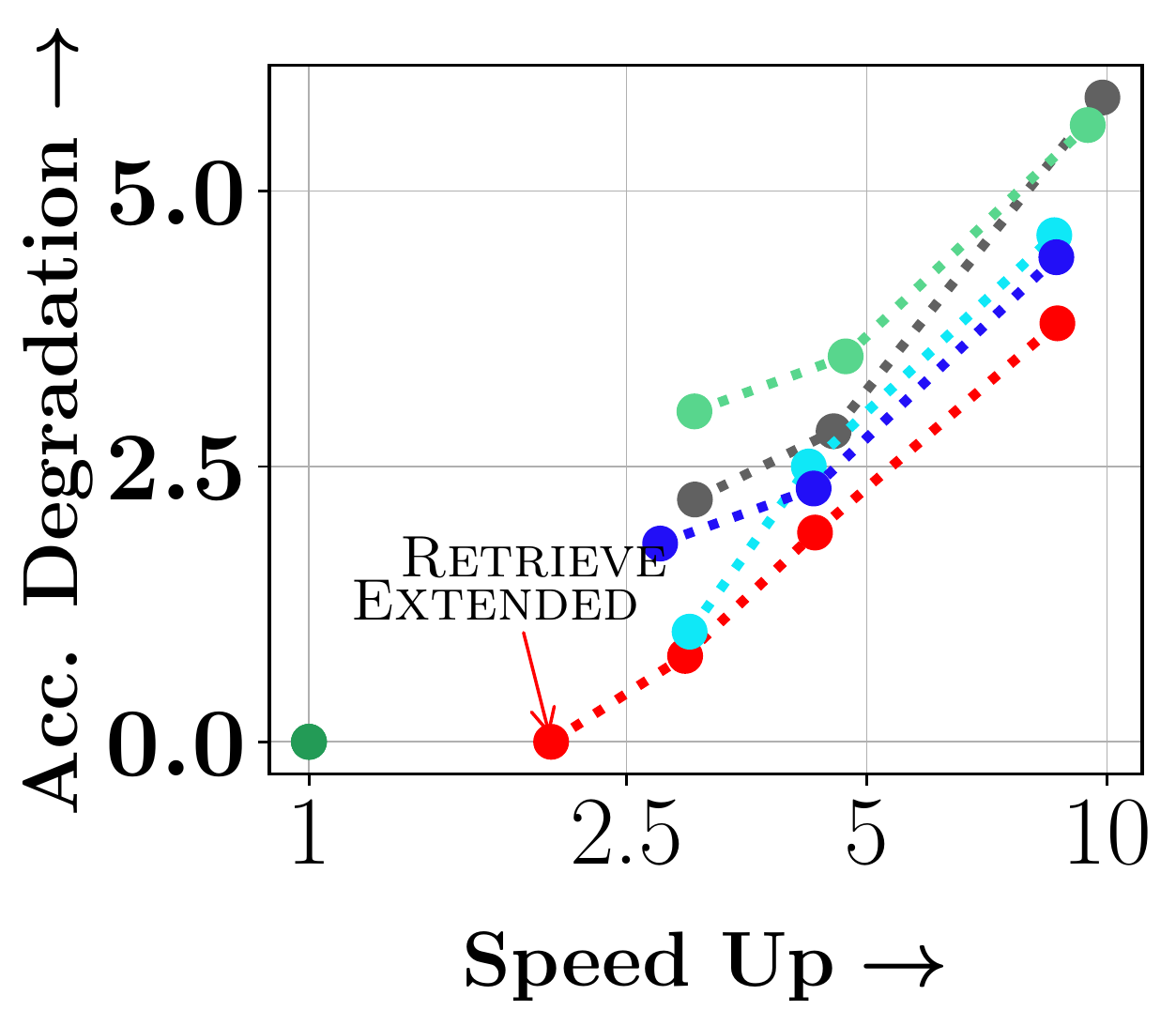}
\caption*{$\underbracket[1pt][1.0mm]{\hspace{4.2cm}}_{\substack{\vspace{-4.0mm}\\
\colorbox{white}{(a) \scriptsize CIFAR10-VAT Extended}}}$}
\phantomcaption
\label{fig:vat_cifar10_extended}
\end{subfigure}
\begin{subfigure}[b]{0.32\textwidth}
\centering
\includegraphics[width=4.2cm, height=3.5cm]{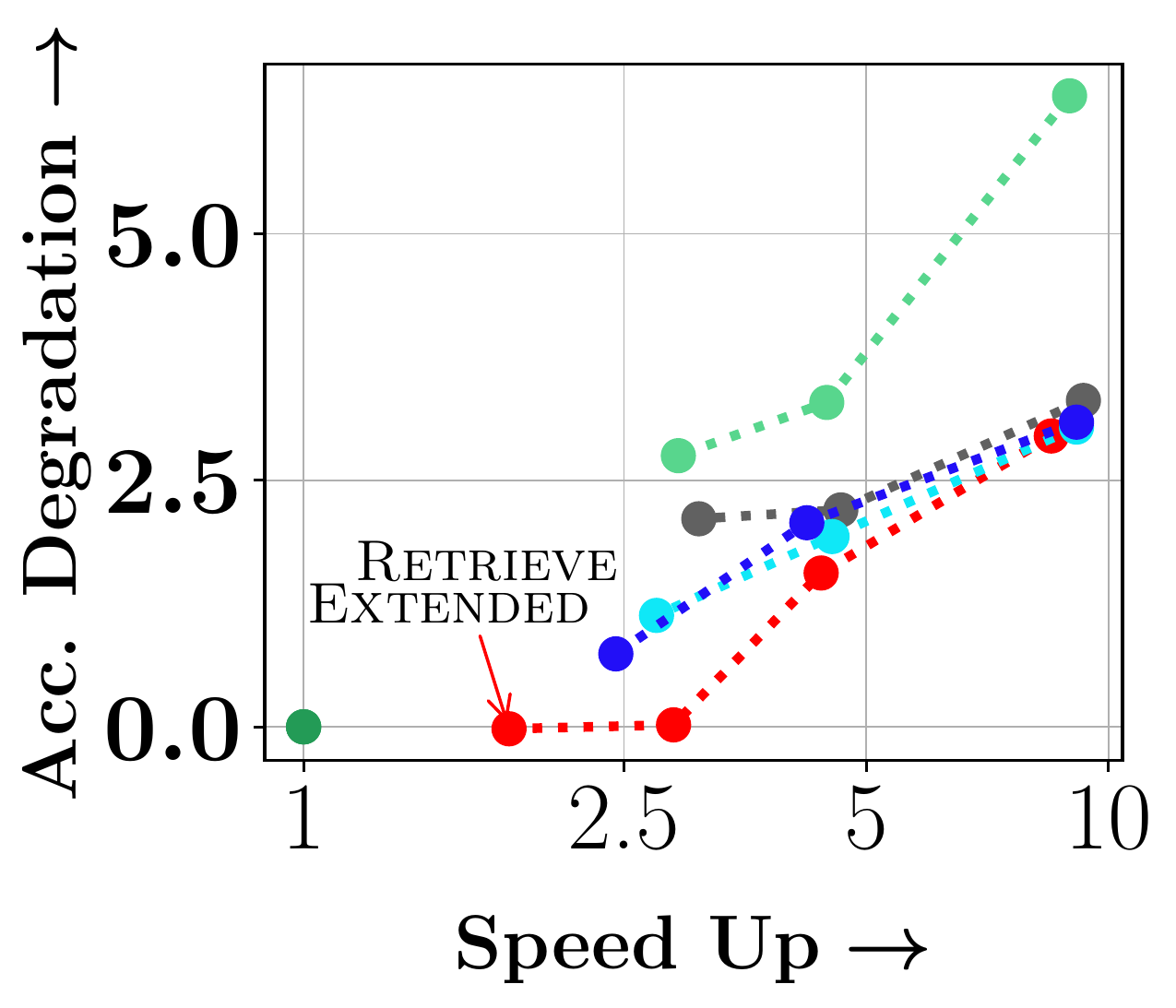}
\caption*{$\underbracket[1pt][1.0mm]{\hspace{4.2cm}}_{\substack{\vspace{-4.0mm}\\
\colorbox{white}{(b) \scriptsize CIFAR10-MT Extended}}}$}
\phantomcaption
\label{fig:mt_cifar10_extended}
\end{subfigure}\quad
\begin{subfigure}[b]{0.32\textwidth}
\centering
\includegraphics[width=4.2cm, height=3.5cm]{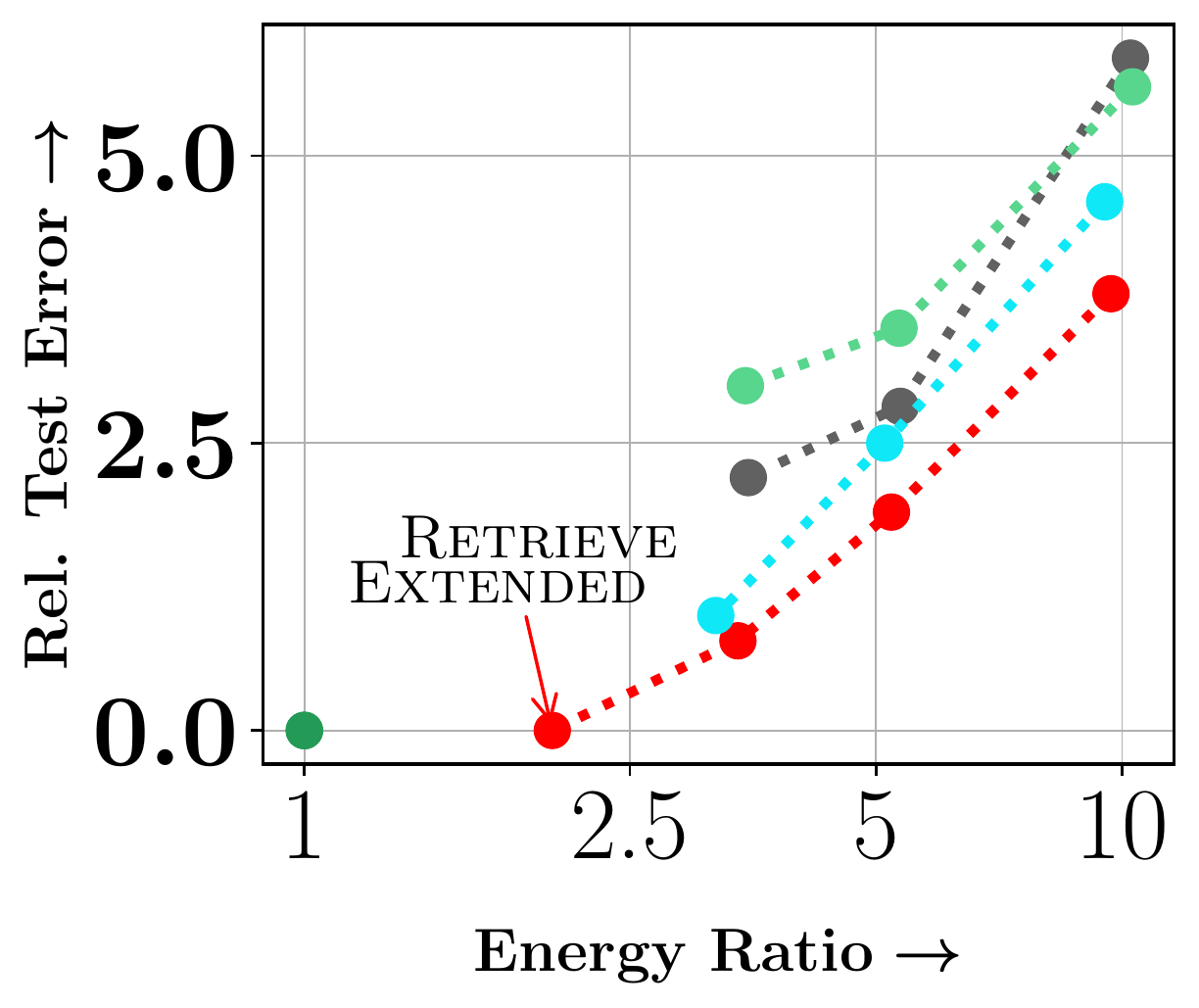}
\caption*{$\underbracket[1pt][1.0mm]{\hspace{4.2cm}}_{\substack{\vspace{-4.0mm}\\
\colorbox{white}{(b) \scriptsize CIFAR10-VAT Extended Energy}}}$}
\phantomcaption
\label{fig:vat_extended_energy}
\end{subfigure}\quad

\caption{Figure 8: \small{Subfigure (a) shows scatter plot of VAT for CIFAR10 dataset with subset fractions of 10\%, 20\%, 30\% along with \textsc{Retrieve-Extended} by training the model for more iterations: We show that the \model{} achieves similar performance to original VAT algorithm while being 2X faster. Subfigure (b) shows scatter plot of MT for CIFAR10 dataset with subset fractions of 10\%, 20\%, 30\% along with \textsc{Retrieve-Extended} by training the model for more iterations: We show that the \model{} achieves similar performance to original MT algorithm while being 1.8X faster. Subfigure (c) shows energy efficiency scatter plot of VAT for CIFAR10 dataset with subset fractions of 10\%, 20\%, 30\% along with \textsc{Retrieve-Extended} by training the model for more iterations: We show that the \model{} achieves similar performance to original VAT algorithm while being 2X more energy efficient.}}
\label{fig:add_tssl_plots}
\end{figure}
\section{Additional Experiments in Traditional SSL}\label{app:traditional_ssl}
\subsection{\textsc{Retrieve-Warm} vs \model{}}
We show that \textsc{Retrieve-Warm} is more efficient and effective compared to \model{} in the traditional SSL setting in \figref{fig:retvsretw}. Hence, in our experiments, we consider the warm variant of \model{} in traditional SSL scenario.

\subsection{Test-Accuracies, Training times and Standard deviations}
\tabref{tab:vat_tssl_results}, \tabref{tab:mt_tssl_results}, \tabref{tab:fm_tssl_results} shows the top-1 test accuracies and training times taken by \model{} and the other baselines considered in traditional SSL scenario for VAT, Mean-Teacher on CIFAR10, SVHN datasets and for FixMatch algorithm on CIFAR dataset for different fractions of 10\%, 20\% and 30\% respectively. Furthermore, \tabref{tab:tssl_vat_std}, \tabref{tab:tssl_mt_std}, \tabref{tab:tssl_fm_std} gives the standard deviation numbers of \model{} and other baselines in traditional SSL scenarios for VAT, Mean-Teacher on CIFAR10, SVHN datasets and for FixMatch algorithm on CIFAR dataset for different fractions of 10\%, 20\% and 30\% respectively. 

\subsection{MT extended convergence plot}
Subfigure~\ref{fig:mt_ext_convergence} shows the extended convergence plot of \model{} using Mean-Teacher algorithm on CIFAR10 dataset for 30\% subset fraction. From the plot, it is evident that \model{} achieves similar performance to original MT algorithm while being 1.8X faster and better performance than the original MT algorithm while being 1.5X faster.

\subsection{Energy savings}
Subfigure~\ref{fig:vat_energy} shows the energy efficiency plot of \model{} using VAT algorithm on CIFAR10 dataset for 10\%, 20\%, 30\% subset fractions. For calculating the energy consumed by the GPU/CPU cores, we use pyJoules\footnote{\scriptsize{\url{https://pypi.org/project/pyJoules/}}.}. From the plot, it is evident that \model{} is 3.1X energy efficient compared to the original VAT algorithm with an accuracy degradation of 0.78\%.

\begin{table}[!tbhp]
\centering
\scalebox{0.7}{
\begin{tabular}{c c c|c|c} \hline \hline
\multicolumn{5}{c}{\model{} vs \textsc{Retrieve-Warm} in Robust SSL}\\ \hline
\multicolumn{3}{c|}{} & \multicolumn{1}{c|}{Top-1 Test accuracy(\%)} & \multicolumn{1}{c}{Model Training time(in hrs)} \\ 
\multicolumn{1}{c}{} & \multicolumn{1}{c}{} & \multicolumn{1}{c|}{OOD ratio(\%)} &  \multicolumn{1}{c|}{50\%}  & \multicolumn{1}{c}{50\%} \\ \hline
\multicolumn{1}{c}{Dataset} & \multicolumn{1}{c}{Model}  &\multicolumn{1}{c|}{Selection Strategy} & \multicolumn{1}{c|}{} & \multicolumn{1}{c}{} \\ \hline
CIFAR10 OOD &Wide-ResNet-28-2 &\textsc{Retrieve-Warm}  &78.8 &18.2 \\ 
 & &\textsc{Retrieve} &{\color{red}79} &{\color{red}17.03} \\\hline \hline
MNIST OOD &Two layer CNN model &\textsc{Retrieve-Warm}  &95.3 &1.43 \\ 
 & &\textsc{Retrieve} &{\color{red}95.85} &{\color{red}1.37} \\\hline \hline
 \multicolumn{1}{c}{} & \multicolumn{1}{c}{} & \multicolumn{1}{c|}{Class imbalance ratio(\%)} &  \multicolumn{1}{c|}{50\%} & \multicolumn{1}{c}{50\%} \\ \hline
 CIFAR10 Imbalance &Wide-ResNet-28-2 &\textsc{Retrieve-Warm}  &76.13 &18.6 \\ 
 & &\textsc{Retrieve} &{\color{red}78.86} &{\color{red}17.3} \\\hline \hline
\end{tabular}}
    \caption{\model{} vs \textsc{Retrieve-Warm} in Robust SSL scenario for CIFAR10 OOD, MNIST OOD and CIFAR10 Imbalance datasets using VAT algorithm}
    \label{tab:vat_rssl_retvsretw}
\end{table}

\begin{sidewaystable}[!tbhp]
\centering
\scalebox{0.9}{
\begin{tabular}{c c c|c c c|c c c} \hline \hline
\multicolumn{9}{c}{VAT Robust SSL Results}\\ \hline
\multicolumn{3}{c|}{} & \multicolumn{3}{c|}{Top-1 Test accuracy(\%)} & \multicolumn{3}{c}{Model Training time(in hrs)} \\ 
\multicolumn{1}{c}{} & \multicolumn{1}{c}{} & \multicolumn{1}{c|}{OOD ratio(\%)} &  \multicolumn{1}{c}{25\%} & \multicolumn{1}{c}{50\%} & \multicolumn{1}{c|}{75\%} & \multicolumn{1}{c}{25\%} & \multicolumn{1}{c}{50\%} & \multicolumn{1}{c}{75\%} \\ \hline
\multicolumn{1}{c}{Dataset} & \multicolumn{1}{c}{Model}  &\multicolumn{1}{c|}{Selection Strategy} & \multicolumn{3}{c|}{} & \multicolumn{3}{c}{} \\ \hline
CIFAR10 OOD &Wide-ResNet-28-2 &\textsc{VAT}  &76.3&  75.6&  74.25 &30.34 &30.34  &30.34 \\ 
& &\textsc{SUPERVISED}  &76.1 &76.1 &76.1  &{\color{red}0.22} &{\color{red}0.22} &{\color{red}0.22} \\ 
 & &\textsc{L2RW}  &78.2 &75.5 &73.4  &86.51 &86.58 &86.62\\ 
 & &\textsc{DS3L}  &78.8 &77.6 & 76.3 &88.94 &88.91 &88.92\\ 
 & &\textsc{Retrieve} &{\color{red}79.26} &{\color{red}79} &{\color{red}76.56}  &17.36 &17.03 &17.12  \\\hline \hline
MNIST OOD &Two layer CNN model &\textsc{VAT}  &95&  92.2& 88.1 &2.46 &2.46 &2.46 \\ 
& &\textsc{SUPERVISED}  &93 &93 &93  &{\color{red}0.01} &{\color{red}0.01} &{\color{red}0.01} \\ 
 & &\textsc{L2RW}  &95.2 &88.5 &87.5  &7.34 &7.29 &7.23 \\ 
 & &\textsc{DS3L}  &97.1 &95.8 &92.1  &7.22 &7.18 &7.12\\ 
 & &\textsc{Retrieve} &{\color{red}97.3} &{\color{red}95.85} &{\color{red}93.48}  &1.365 &1.37 &1.36  \\\hline \hline
\multicolumn{1}{c}{} & \multicolumn{1}{c}{} & \multicolumn{1}{c|}{Class imbalance ratio(\%)} &  \multicolumn{1}{c}{10\%} & \multicolumn{1}{c}{30\%} & \multicolumn{1}{c|}{50\%} & \multicolumn{1}{c}{10\%} & \multicolumn{1}{c}{30\%} & \multicolumn{1}{c}{50\%} \\ \hline
 CIFAR10 Imbalance &Wide-ResNet-28-2 &\textsc{VAT}  &56.12&  65.15& 72.14 &30.24 &30.26  &30.2 \\ 
& &\textsc{SUPERVISED}  &58.12 &64.21 &71.12  &{\color{red}0.22} &{\color{red}0.22} &{\color{red}0.22} \\ 
 & &\textsc{L2RW}  &61.54 &68.45 &71.24  &87.35 &87.19 &87.41 \\ 
 & &\textsc{DS3L}  &63.54 &73.89 &77.41  &88.16 &88.04 &88.5\\ 
 & &\textsc{Retrieve} &{\color{red}66.88} &{\color{red}75.83} &{\color{red}78.86}  &17.27 &17.31 &17.3  \\\hline \hline
\end{tabular}}
    \caption{Robust SSL Results for CIFAR10 OOD, MNIST OOD and CIFAR10 Imbalance datasets using VAT algorithm}
    \label{tab:vat_rssl_results}
\end{sidewaystable}

\begin{table}[!ht]
    \centering
    \scalebox{0.8}{
    \begin{tabular}{c c c|c c c} \hline \hline
\multicolumn{6}{c}{VAT Standard Deviation Results}\\ \hline
\multicolumn{3}{c|}{} & \multicolumn{3}{c}{Standard deviation of the Model(for 3 runs)} \\ 
\multicolumn{1}{c}{} & \multicolumn{1}{c}{} & \multicolumn{1}{c|}{OOD ratio(\%)} & \multicolumn{1}{c}{25\%} & \multicolumn{1}{c}{50\%} & \multicolumn{1}{c}{75\%} \\ \hline
\multicolumn{1}{c}{Dataset} & \multicolumn{1}{c}{Model}  &\multicolumn{1}{c|}{Selection Strategy} & \multicolumn{3}{c}{} \\ \hline
CIFAR10 OOD &Wide-ResNet-28-2 &\textsc{VAT}  &0.13	&0.18	&0.24 \\ 
 & &\textsc{SUPERVISED}  &0.021 &0.021 &0.021 \\ \cline{3-6}
 & &\textsc{L2RW}  &0.31 & 0.39	&0.295\\
 & &\textsc{DS3L}  &0.38	&0.41	&0.34\\
 & &\textsc{Retrieve}  &0.26 &0.21 &0.27\\
 \hline \hline
MNIST OOD &Two layer CNN model &\textsc{VAT}  &0.014	&0.018 &0.021 \\ 
 & &\textsc{SUPERVISED}  &0.01 &0.01 &0.01 \\ \cline{3-6}
 & &\textsc{L2RW}  &0.04	&0.03	&0.04\\
 & &\textsc{DS3L}  &0.061	&0.041	&0.056\\
 & &\textsc{Retrieve}  &0.034 &0.039 &0.036\\
 \hline \hline
\multicolumn{1}{c}{} & \multicolumn{1}{c}{} & \multicolumn{1}{c|}{Class imbalance ratio(\%)} & \multicolumn{1}{c}{10\%} & \multicolumn{1}{c}{30\%} & \multicolumn{1}{c}{50\%} \\ \hline
CIFAR10 Imbalance &Wide-ResNet-28-2 &\textsc{VAT}  &0.295	&0.242	&0.185 \\ 
 & &\textsc{SUPERVISED}  &0.16 &0.13 &0.11 \\ \cline{3-6}
 & &\textsc{L2RW}  &0.37	&0.32	&0.26\\
 & &\textsc{DS3L}  &0.34	&0.36	&0.21\\
 & &\textsc{Retrieve}  &0.32 &0.28 &0.205\\
 \hline \hline
\end{tabular}}
    \caption{Standard deviation results using VAT in Robust SSL scenario for CIFAR10 OOD, MNIST OOD and CIFAR10 Imbalance datasets for three runs.}
    \label{tab:rssl_vat_std}
\end{table}

\section{Additional Experiments for Robust SSL}\label{app:robust_ssl}
\subsection{\textsc{Retrieve-Warm} vs \model{}}
We show that \textsc{Retrieve} is more efficient and effective compared to \textsc{Retrieve-Warm} in the robust SSL setting from the results given in \tabref{tab:vat_rssl_retvsretw}. For specific numbers, \model{} achieves 79\% accuracy in 17.03 hrs while \textsc{Retrieve-Warm} achieves 78.8\% accuracy in 18.2 hrs for CIFAR10 OOD dataset with an OOD ratio of 50\%. Further, \model{} achieves 95.85\% accuracy in 1.37 hrs while \textsc{Retrieve-Warm} achieves 95.3\% accuracy in 1.43 hrs for MNIST OOD dataset with an OOD ratio of 50\%. Finally, \model{} achieves 78.86\% accuracy in 17.3 hrs while \textsc{Retrieve-Warm} achieves 76.13\% accuracy in 18.6 hrs for CIFAR10 Imbalance dataset with a class imbalance ratio of 50\%. Hence, in our experiments, we consider \model{} without warm variant in robust SSL scenario.

\subsection{Test-Accuracies, Training times and Standard deviations: }
\tabref{tab:vat_rssl_results} shows the top-1 test accuracies and training times taken by \model{} and the other baselines considered in robust SSL scenario for VAT on CIFAR10 OOD, MNIST OOD, and CIFAR10 Imbalance datasets. The results show that \model{} with VAT outperforms all other baselines, including DS3L~\cite{pmlr-v119-guo20i} (also run with VAT) in the class imbalance scenario as well. In particular, \model{} outperforms other baselines by around 1.5\% on the CIFAR-10 with imbalance. Furthermore, \tabref{tab:rssl_vat_std} gives the standard deviation numbers of \model{} and other baselines in robust SSL scenario for VAT on CIFAR10 OOD, MNIST OOD, and CIFAR10 Imbalance datasets.

\section{Broader Impacts and Limitations}\label{app:imp_lim}

\textbf{Limitations: } One of the main limitations of \model{} is that even though it reduces the training time, energy costs, and CO2 emissions of SSL algorithms, it does not reduce the memory requirement. Furthermore, the memory requirement is a little higher because it requires additional memory to store the gradients required for the coreset selection, which makes running the \model{} algorithm in devices with low memory capacity significantly harder without proper memory handling.

\textbf{Societal Impacts: } We believe \model\ has a significant positive societal impact by making SSL algorithms (and specifically robust SSL) significantly faster and energy-efficient, thereby reducing the CO2 emissions and energy consumption incurred during training. This is particularly important because state-of-the-art SSL approaches like FixMatch are very computationally expensive. Furthermore, SSL approaches often have a large number of hyper-parameters and the performance can be heavily dependent on the right tuning of these hyper-parameters~\cite{sohn2020fixmatch,oliver2019realistic}. We believe that \model\ can enable much faster and energy efficient tunings of hyper-parameters in SSL approaches thereby enabling orders of magnitude speedup and CO2 emissions being reduced. \model\ takes one step towards \textbf{Green-AI} by enabling using smaller subsets for training these models.


\end{document}